\documentclass[journal]{IEEEtran}
\usepackage[T1]{fontenc}

\usepackage{graphicx}
\usepackage{letltxmacro}

\LetLtxMacro{\origincludegraphics}{\includegraphics}
\renewcommand{\includegraphics}[2][]{%
  \IfFileExists{#2}{\origincludegraphics[#1]{#2}}{\fbox{Missing graphic: \texttt{#2}}}%
}

\usepackage{amsmath,amssymb}
\usepackage[table]{xcolor} 
\usepackage{booktabs}
\usepackage{array}
\usepackage{multirow}
\usepackage{makecell}
\usepackage{siunitx} 
\usepackage{subcaption} 
\usepackage[hidelinks]{hyperref}

\ifCLASSINFOpdf
\else
\fi

\begin{document}
%
\title{\LARGE \bf
ECGLight: Compute-Light Framework For Paper ECG Digitization and Myocardial Infarction Screening
}

\author{Shreyasvi Natraj$^{1,2*}$, Cyrus Achtari$^{1,2}$, Felice Gragnano$^{3}$, Andrea Milzi$^{4}$,\\
Marco Valgimigli$^{4}$, and Diego Paez-Granados$^{1,2*}$
\thanks{$^{1}$Spinal Cord and Artificial Intelligence (SCAI) Lab, ETH Z\"urich, Z\"urich, Switzerland}%
\thanks{$^{2}$Swiss Paraplegic Research, Nottwil, Luzern, Switzerland}%
\thanks{$^{3}$Department of Translational Medical Sciences, University of Campania ``Luigi Vanvitelli'', Naples, Italy}%
\thanks{$^{4}$Department of Biomedical Sciences, University of Italian Switzerland, Cardiocentro Ticino Institute, Lugano, Switzerland}%
\thanks{$^{*}$Corresponding author email: {\tt\small snatraj@ethz.ch}}
}

\maketitle

\begin{abstract}
Electrocardiography (ECG) is one of the most widely used tests for diagnosing cardiovascular disease. Yet several remote clinics still utilize paper ECG printouts for their analysis due to limited connectivity and computational capacity. As a result, vast numbers of physical ECGs obtained in remote areas still remain incapable of being accessed by contemporary artificial-intelligence (AI)–based decision support as they require high computational resources or strong high-speed internet connectivity. This causes several cases where conditions like acute coronary occlusion (ACS) is overlooked and reperfusion therapy delayed. Although prior work has tackled digitization and diagnosis separately, and utilized advanced AI models for them, there still remains a lack of a compute-light, on-device framework that reconstructs paper ECGs at high fidelity, while accurately supporting multiple clinically relevant endpoints. We address this need with an end-to-end \textbf{lightweight on-device digitization-to-diagnosis} pipeline that converts a smartphone photo or scan of a paper ECG into a calibrated 12-lead signal and screens for \textit{Myocardial Infarction (MI)} pathologies, with \texttt{SHapley Additive exPlanations (SHAP)} to support interpretability. Trained and evaluated on 21,799 ECGs from the \texttt{PTB-XL} dataset and further validated on hospital-acquired \texttt{ECG-Matrix} dataset, the complete system runs in \textbf{<30 s per ECG} on CPU-only resources, achieving \textbf{95.51\%} accuracy (F1 = 0.9519) for \textit{MI detection} on \texttt{PTB-XL} and \textbf{88.89\%} accuracy (F1 = 0.8862) for \textit{OMI detection} on \texttt{ECG-Matrix}. This work showcases that legacy paper records can be reliably democratized in any part of the world, providing a scalable decision support when digital ECG export, connectivity, or high-end compute are unavailable  

\end{abstract}

\begin{IEEEkeywords}
Electrocardiography, \texttt{YOLO}, offline inference, CPU-only deployment, edge AI, time-series classification, myocardial infarction detection, acute coronary syndrome, ECG-Matrix, PTB-XL, PhysioNet Challenge
\end{IEEEkeywords}

\section{INTRODUCTION}

Cardiovascular diseases (CVDs), particularly Acute Coronary Syndromes (ACS), remain the leading cause of mortality worldwide, accounting for approximately 33\% of all deaths, or nearly 20 million fatalities annually\cite{mitsis_pathophysiology_2021,bhatt_diagnosis_2022}. Within the broad spectrum of CVDs, ACS encompassing unstable angina as well as ST-elevation and non–ST-elevation myocardial infarction, represent some of the most prevalent and life-threatening clinical manifestations, affecting roughly one in five individuals globally. Given the immense clinical and socioeconomic burden associated with CVDs and ACS, there is a pressing need for advanced diagnostic modalities that enable early detection, continuous monitoring, and timely intervention.

In this context, the electrocardiogram (ECG) remains a cornerstone of cardiovascular and ACS diagnostics due to its capacity to non-invasively record the heart’s bioelectrical activity over time \cite{rosiek_risk_2016}. ECG analysis yields critical information about the electrophysiological behavior of the myocardium and supports the detection of a wide range of cardiac pathologies, including arrhythmias, myocardial ischemia or infarction, ACS-related ischemic changes, and structural or conduction abnormalities \cite{mirvis_electrocardiography_2001}. Building on this physiological foundation, electrocardiograms provide a standardized, time-resolved waveform representation of the cardiac cycle, recorded at the body surface. By measuring voltage differences across multiple leads, ECGs furnish a comprehensive spatiotemporal depiction of cardiac activity. This representation facilitates both qualitative visual interpretation by clinicians and quantitative feature extraction for downstream computational modeling, thereby positioning the ECG as a pivotal tool for the development of advanced diagnostic approaches in ACS and broader CVD care. 

In the context of acute coronary occlusion, these same waveform features become critical diagnostic markers\cite{mirvis_electrocardiography_2001,goldberger_clinical_2017}. Myocardial infarction (MI) and, more specifically, Occlusion Myocardial Infarction (OMI) are identified by characteristic changes in the QRS complex, ST segment, and T wave that reflect evolving patterns of transmural and subendocardial ischemia, injury, and necrosis\cite{mirvis_electrocardiography_2001,goldberger_clinical_2017,meyers_omi_2021}. ST-segment elevation or depression, J-point deviation, alterations in T-wave morphology (hyperacute, inverted, or biphasic T waves), the emergence of pathological Q waves, and dynamic changes across serial tracings enable clinicians to localize the infarct-related artery, estimate the extent and acuity of myocardial injury, and distinguish OMI from non-occlusive ischemic syndromes\cite{goldberger_clinical_2017,meyers_omi_2021}. Consequently, high-resolution analysis of these ECG components—both by expert readers and automated algorithms—underpins contemporary strategies for early OMI detection, triage for emergent reperfusion therapy, and risk stratification in patients presenting with suspected myocardial infarction\cite{meyers_omi_2021}.


\begin{figure*}[t]
\centering
\includegraphics[width=\textwidth]{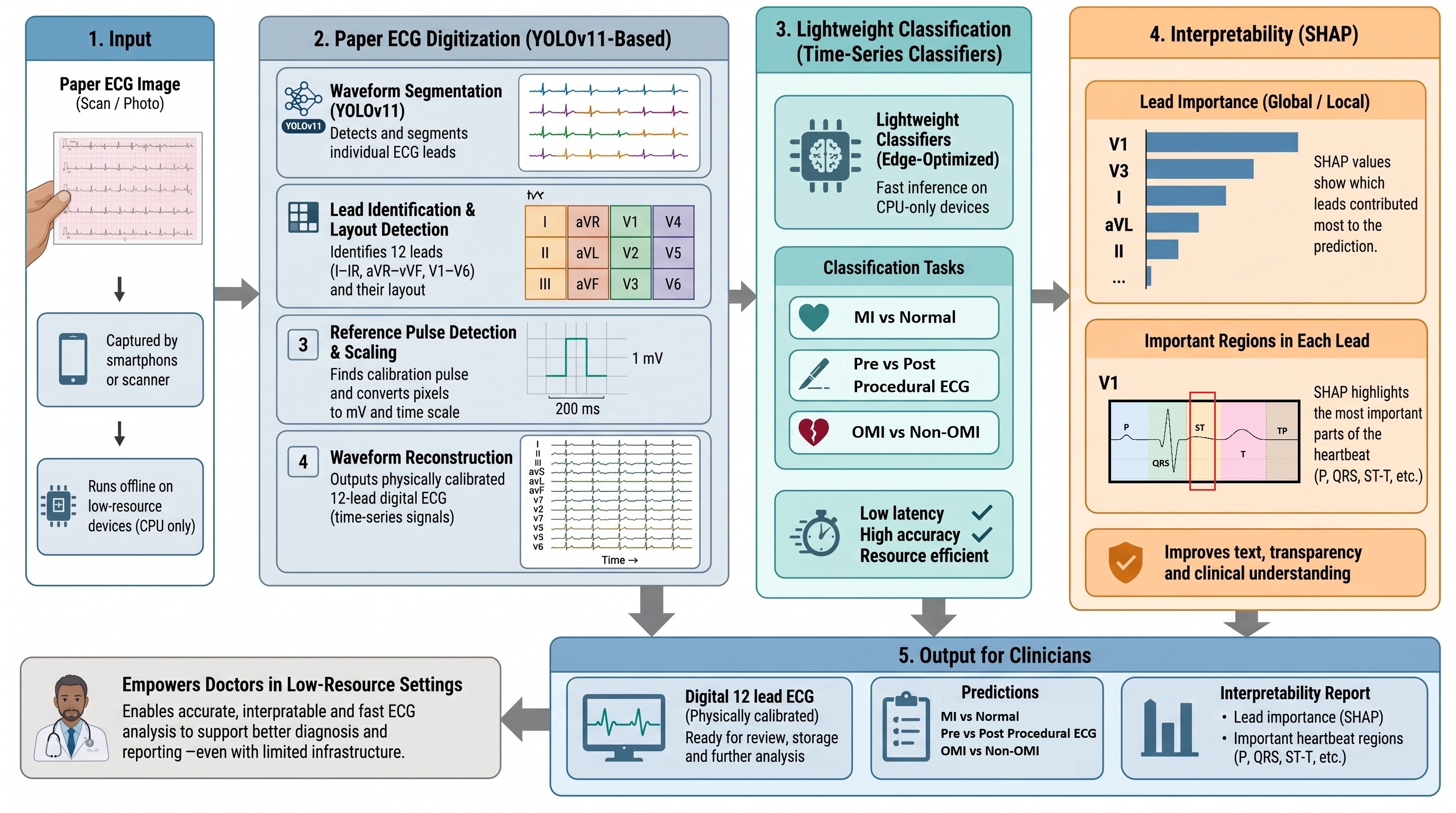}
\caption{\textbf{End-to-End ECG Workflow:} A photographed or scanned paper ECG is preprocessed (shadow/background suppression and denoising) and then analyzed with a \texttt{YOLOv11}-based module to segment the inked waveforms, detect lead labels, identify reference pulses, and infer the page layout. Reference-pulse--based calibration maps pixels to seconds and millivolts, enabling reconstruction and temporal alignment of the full 12-lead time series. The digitized signals are then post-processed (baseline trimming, per-lead normalization, and R-peak detection) and passed to downstream classifiers for automated interpretation. On a CPU-only Intel Core i9-13900H laptop, the end-to-end pipeline runs in roughly 25--30~s per ECG.}
\label{fig:pipeline}
\end{figure*}


Despite advancements in digital ECG systems, an estimated 300 million ECG tests conducted annually are still stored as paper printouts or scanned Portable Document Format (PDF) files within Electronic Health Records (EHRs), rather than as raw digital signals \cite{chang_resting_2022, sassi_pdfecg_2017}. These paper-based ECGs (see \href{fig:ecg_ex}{Figure \ref{fig:ecg_ex}}) typically include a header with patient and recording details, grid lines (scaled at 10 mm/mV and 0.04 s/mm), lead labels, reference pulses (1 mV, 0.2 s), lead signals, and often a rhythm strip (typically an extended lead II for rhythm analysis).

Paper ECGs are commonly arranged in one of 16 standard layouts (12$\times$1, 6$\times$2, 4$\times$3, or 3$\times$4), with leads grouped by type (I, II, III in one row for 4$\times$3) or in \texttt{Cabrera} format, which reorders limb leads as aVL, I, -aVR, II, aVF, III to group inferior (II, aVF, III) and lateral (aVL, I, -aVR) views for enhanced physiological interpretation \cite{lindow_why_2019}. The \texttt{Cabrera} format improves visualization of axis and regional changes, aiding clinical diagnosis. However, the analog nature of paper ECGs, coupled with paper’s susceptibility to degradation, poses significant challenges for computational analysis, archival storage, and large-scale research \cite{randazzo_development_2022}. Manual transcription is labor-intensive, error-prone, and unscalable, particularly for retrospective studies or real-time clinical applications \cite{Jones2021}. Automated ECG digitization—converting paper-based recordings into digital time-series data—is thus critical for preserving millions of records, enabling longitudinal studies, investigating rare diseases, and developing high-quality datasets for artificial intelligence (AI) applications \cite{lence_automatic_2023}. In low-resource settings, digitization can bridge gaps in access to modern diagnostic tools, enhancing care quality where digital infrastructure is limited \cite{owoyemi_artificial_2020}.

\section{RELATED WORKS}
Traditional ECG digitization methods rely on image preprocessing techniques, such as denoising and de-skewing via \texttt{Hough} or \texttt{Radon} transforms, followed by thresholding to separate signals, grid lines, and backgrounds \cite{badilini_ecgscan_2005, santamonica_ecgminer_2024, fortune_digitizing_2022}. For example, \texttt{ECGScan} requires user input for grid detection, layout selection, and anchor points, using active contours for signal extraction \cite{badilini_ecgscan_2005}. \texttt{ECGMiner}, an open-source \texttt{MATLAB} tool, employs a custom cost function for signal vectorization but still requires manual layout specification \cite{santamonica_ecgminer_2024}. Fortune et al. utilized the \texttt{Viterbi} algorithm for signal tracing, achieving accurate waveform reproduction but struggling with non-standard layouts and noise \cite{fortune_digitizing_2022}. These rule-based approaches are sensitive to variations in format, scanning artifacts, paper aging, and image quality, limiting their robustness and scalability.

Recent advancements leverage deep learning to address these limitations, offering superior robustness to noise and automation capabilities \cite{mishra_ecg_2021, li_deep_2020, demolder_high_2025}. Mishra et al. applied deep learning for adaptive thresholding, improving signal detection in noisy images \cite{mishra_ecg_2021}. Li et al. used 128$\times$128 pixel patches for precise segmentation, enhancing accuracy in challenging imaging conditions \cite{li_deep_2020}. Demolder et al. introduced the \texttt{Dotter algorithm} for distortion correction based on grid intersection points, claiming compatibility with multiple layouts, though methodological details are limited \cite{demolder_high_2025}. Lead name detection has been advanced using \texttt{Optical Character Recognition (OCR)} \cite{ganesh_combining_2021} or deep learning models to identify lead boundaries \cite{wu_fully-automated_2022}. Reference pulse detection for scale calibration has been explored, often with assumptions about pulse location \cite{wang_paper-recorded_2024, lence_ecgtizer_2024}. A systematic review by Lence et al. identified key challenges in current pipelines, including limited code availability, dataset-specific approaches, reliance on manual input, and poor robustness to image quality variations \cite{lence_automatic_2023}.

Beyond digitization itself, clinical utility ultimately depends on whether reconstructed multi-lead time-series data can support downstream diagnostic tasks at scale. Most modern ECG classifiers are developed and validated on digitally acquired waveforms, using either feature-based pipelines (wavelet or morphological descriptors) or deep neural networks (Convolutional Neural Networks (CNNs), residual networks, and Transformer variants) trained on large repositories such as \texttt{PTB-XL} \cite{ptb_xl, strodthoff_ecg_2021}. In the context of acute coronary syndromes, common classification objectives include \textit{MI} vs. \textit{Normal} \cite{acharya_automated_2017}, often implemented with convolutional or residual architectures trained on large-scale ECG corpora \cite{strodthoff_ecg_2021}. Some studies extend this further by using paired or serial ECG analyses, where Pre- vs. Post-intervention (or Pre- vs. Post-procedural) comparisons capture dynamic ST-segment and T-wave evolution following invasive management with percutaneous coronary intervention (PCI) \cite{gragnano_ecg_2019}. Building on these developments, the \textit{Occlusion MI (OMI)} paradigm was proposed to better identify angiographically occlusive events that may not meet classic ST-elevation criteria, motivating dedicated \textit{OMI} vs. \textit{non-OMI} classification models as a complementary endpoint \cite{meyers_omi_2021}.

Within this broader landscape, the \texttt{2024 George B. Moody PhysioNet Challenge} highlighted the potential of deep learning for ECG digitization, with top entries employing segmentation-based approaches on synthetic \texttt{PTB-XL} images generated using the \texttt{ECG-Image-Kit library} \cite{a_reyna_digitization_2024, kodthalu2024ecg, krones_combining_2024}. These efforts underscore the need for robust, automated pipelines capable of handling diverse ECG formats and noise conditions. However, most existing work still treats digitization and classification as separate problems. Comparatively few end-to-end systems jointly address digitization together with multiple clinically relevant classification endpoints (\textit{MI/Normal}, \textit{pre-/post-intervention}, and \textit{OMI/non-OMI}), while also satisfying the strict latency and compute constraints required for on-device deployment \cite{howard_mobilenets_2017, banbury_tinyml_2021}.

\section{CONTRIBUTIONS}
In response to this need, our study investigates an end-to-end framework for automated ECG digitization and downstream MI-oriented classification that leverages lightweight segmentation via \texttt{YOLOv11}\cite{jocher_ultralytics_2023} with a patch-based inference strategy (see \href{fig:pipeline}{Figure~\ref{fig:pipeline}}). The system performs waveform segmentation, lead-name detection (I, II, III, aVR, aVL, aVF, V1--V6) for layout identification, and reference-pulse detection for physically grounded amplitude scaling. The digitization pipeline is developed using \texttt{PTB-XL}\cite{ptb_xl} synthetic paper renderings and applied \emph{without re-training} to \texttt{ECG-Matrix}\cite{gragnano_ecg_2019, valgimigli_matrix_2018} real hospital paper ECG images to assess cross-source generalization. Beyond digitization, the reconstructed 12-lead time-series signals are used for multiple clinically relevant endpoints, including \textit{Normal} vs. \textit{MI}, \textit{Pre-procedural} vs. \textit{Post procedural} MI, and \textit{OMI} vs. \textit{Non-OMI} classification. Finally, we prioritize compact time-series models suitable for CPU-only inference and use \texttt{SHAP} to provide physiologically grounded lead- and time-resolved attributions that support auditing of model behavior.

\section{METHODS}


\subsection*{\texttt{YOLO}-Based Patched Segmentation Of Noisy ECG Images}
To digitize noisy paper ECGs, we used \texttt{YOLO}~\cite{jocher_ultralytics_2023} for pixel-accurate instance segmentation of ECG waveforms. Anchor-free heads and feature-pyramid aggregation help preserve quasi-linear, repetitive traces under grid lines, shadows, and scanning artifacts. The \mbox{62.1 M}-parameter model with FP16 acceleration achieves sub-300~ms \textit{CPU-only} inference on \texttt{4\,$\times$\,A78 @ 2.4~GHz}~\cite{jocher_ultralytics_2023}.

Two preprocessing regimes were used: \textit{Full-Image} (upsample full sheets to 2000$\times$2000 for continuity and inter-lead context) and \textit{Patched} (partially overlapping tiles, upsampled to 1280$\times$1280, for finer local contour learning).

\subsubsection*{Data Generation}
We used \texttt{PTB-XL}~\cite{ptb_xl} (\textbf{21,799} 12-lead recordings from \textbf{18,869} patients; 10\,s at \textbf{500~Hz}). Since \texttt{PTB-XL} provides WFDB signals (not scans), we rendered each record into a realistic paper ECG with \texttt{ECG-Image-Kit}~\cite{kodthalu2024ecg, ecgimagekit2024} to obtain paired image--signal supervision. The renderer supports \texttt{12$\times$1}, \texttt{6$\times$2}, \texttt{4$\times$3}, \texttt{3$\times$4}, and extended \texttt{Cabrera} layouts; scanning variability was simulated with wrinkles/creases, rotation, temperature shifts, additive noise, and handwritten overlays (\href{fig:augs}{Figure~\ref{fig:augs}}).

Training data consisted of these rendered sheet images (\href{tab:dataset}{Table~\ref{tab:dataset}}). \textit{Full-Image} used native sheets upsampled to 2000$\times$2000. \textit{Patched} used multiscale overlapping grids (4$\times$5, 5$\times$6, 6$\times$8): from \textbf{11,000} rendered sheets we generated \textbf{$\sim$323,000} patches to increase effective sample count and improve local boundary learning (\href{fig:patchsz}{Figure~\ref{fig:patchsz}}).

\subsubsection*{Model Training and Benchmarking}
We fine-tuned \texttt{YOLOv11x-seg}~\cite{kolesnikov_koldim2001yolo-patch-based-inference_2025} (\mbox{$\sim$62.1 M} parameters) for 100 epochs with the default setup: \texttt{AdamW}~\cite{loshchilov2019decoupled}, cosine LR~\cite{loshchilov2017sgdr}, FP16~\cite{micikevicius2018mixedprecision}, batch size 16, and evaluated both \textit{Full-Image} and \textit{Patched} regimes. For architectural generalization, we repeated training with \texttt{YOLOv12x-seg}~\cite{jocher_ultralytics_2023} under identical preprocessing/training, comparing behavior to \texttt{YOLOv11x} in the \textit{Patched} inference regime.

\subsubsection*{Post-processing and Mask Refinement}
Small patch-size changes caused local inconsistencies in mask density/edges (\href{fig:patchsz}{Figure~\ref{fig:patchsz}}). We therefore ran multiple patch grids and fused them via pixel-wise union for coverage, then applied morphological erosion~\cite{gonzalez_digital_2009} and contour filtering to remove islands and smooth boundaries. Patch masks were mapped back to sheet coordinates and merged with \texttt{NMS} plus spatial-consistency weighting into a single global mask using \texttt{YOLO-Patch-Based-Inference}~\cite{kolesnikov_koldim2001yolo-patch-based-inference_2025}.

\subsubsection*{Segmentation Evaluation}
We used a fixed 80/20 train--validation split and identical fine-tuning settings for \textit{Full-Image} and \textit{Patched} inference with \texttt{YOLOv11x-seg}; validation predictions were confidence-filtered and consolidated with \texttt{NMS}. Metrics included box precision/recall and mAP@50, plus mask overlap via \texttt{IoU} (\href{eq:iou}{Equation~\ref{eq:iou}}) and Dice (\href{eq:dice}{Equation~\ref{eq:dice}}). \textit{Full-Image} compared predictions on the upsampled sheet to full-resolution ground truth; \textit{Patched} reprojected and merged tile masks (multi-grid fusion + morphological/contour refinement) to form one global mask before scoring, with qualitative checks for seam continuity and boundary preservation (\href{tab:results}{Table~\ref{tab:results}}).

\textbf{Intersection over Union (IoU):} A measure of overlap between predicted ($A$) and ground truth ($B$) areas.
\begin{equation}
\label{eq:iou}
\text{IoU} = \frac{|A \cap B|}{|A \cup B|}
\end{equation}

\textbf{Dice Coefficient:} Similarity between predicted segmentation ($A$) and ground truth ($B$), ranging from 0 (no overlap) to 1 (perfect overlap).
\begin{equation}
\label{eq:dice}
\text{Dice} = \frac{2 \times |A \cap B|}{|A| + |B|}
\end{equation}

The same evaluation was replicated with \texttt{YOLOv12x-seg}~\cite{jocher_ultralytics_2023}. Across \textit{Full-Image} and \textit{Patched} settings (\href{tab:results}{Table~\ref{tab:results}A}), \texttt{YOLOv11x-seg} achieved higher precision, recall, and spatial coherence, suggesting its anchor-free heads and feature-pyramid/contextual fusion better match quasi-linear ECG structure than the attention-centric priors in \texttt{YOLOv12x}.

\subsection*{\texttt{YOLOv11x}-Based Lead Name Detection \& Layout Identification}

\subsubsection*{Dataset Preparation}
For lead-name localization and downstream layout inference, we trained a \texttt{YOLOv11x} detector on \textbf{20,000} synthetic samples by overlaying labeled lead names (\texttt{I}, \texttt{II}, \texttt{III}, \texttt{aVR}, \texttt{aVL}, \texttt{aVF}, \texttt{V1--V6}) onto \texttt{PTB-XL}-rendered ECG backgrounds, varying font, size, orientation, and placement (\href{tab:dataset}{Table~\ref{tab:dataset}}). \texttt{Albumentations}~\cite{buslaev_albumentations_2020} augmentations (affine transforms, rotation, brightness/contrast shifts, Gaussian noise, dropout) emulated scanning/print degradation; random distractor characters reduced false positives in metadata/text regions (\href{fig:objdet}{Figure~\ref{fig:objdet}}).

\subsubsection*{Model Training}
The detector was fine-tuned from a pre-trained \texttt{YOLOv11x} backbone (56.9~M parameters) for 100 epochs (\texttt{AdamW}, cosine scheduler, FP16, batch size 16) with an 80/20 train--validation split; \texttt{YOLOv11x} was chosen for its anchor-free heads and strong multi-scale feature aggregation.

\subsubsection*{Layout Identification}
Lead name detections were combined with the segmentation masks to infer the ECG sheet layout via four steps.

(1) \textbf{Row center estimation:} the binary mask was horizontally summed to form a 1D projection; peaks correspond to row centers (\href{fig:peak}{Figure~\ref{fig:peak}A}). Peak detection used adaptive height and spacing parameters (rather than fixed thresholds) to remain stable across scales/resolutions and to tolerate distortions such as baseline wander~\cite{venkatachalam_signals_2011}.

(2) \textbf{ROI selection:} the signal ROI was bounded by the first/last row centers with an added margin equal to the mean inter-row distance (\href{fig:peak}{Figure~\ref{fig:peak}C}), retaining only lead detections within the ROI and excluding headers/calibration regions.

(3) \textbf{Layout classification:} sixteen layout types (including \texttt{12$\times$1}, \texttt{6$\times$2}, \texttt{4$\times$3}, \texttt{3$\times$4}, plus single-rhythm-strip variants) were defined from common \texttt{PTB-XL} configurations and split into conventional vs. \texttt{Cabrera}. Layouts missing any of the twelve standard leads, containing unaligned rows, or having multiple rhythm strips were excluded. The detected row count provided the initial hypothesis; for four-row cases (\texttt{4$\times$3} vs. \texttt{3$\times$4 + rhythm strip}), the alignment of precordial leads (\texttt{V1--V3} vs. \texttt{V4--V6}) resolved ambiguity (\href{fig:layout}{Figure~\ref{fig:layout}}).

(4) \textbf{Conventional vs. Cabrera:} Cabrera ordering \texttt{aVL, I, -aVR, II, aVF, III} replaces \texttt{aVR} with \texttt{-aVR}~\cite{lindow_why_2019}. For linear layouts (e.g., \texttt{12$\times$1}, \texttt{6$\times$2}), doubled spacing between alternating limb leads indicated \texttt{Cabrera} (\href{fig:cabrera}{Figures~\ref{fig:cabrera}A--B}). For grid layouts (\texttt{4$\times$3}, \texttt{3$\times$4}), \texttt{Cabrera} was detected via \texttt{aVF} misalignment relative to \texttt{aVL}/\texttt{aVR} (\href{fig:cabrera}{Figures~\ref{fig:cabrera}C--D}).

\subsubsection*{Lead Name \& Layout Detection Evaluation}  
Performance of the combined lead name detection and layout identification pipeline was assessed on a synthetic test set of 1,600 ECG images, evenly distributed across the 16 defined layout types. The \texttt{YOLOv11x} lead detection model achieved high localization accuracy (precision = $0.997$, recall = $0.991$, mAP@50 = $0.995$), while the layout inference process reached an overall accuracy of $93.5\%$ (\href{tab:results}{Table~\ref{tab:results}B-C}). The best results were observed for structured configurations such as \texttt{3$\times$4} ($0.995$) and \texttt{6$\times$2} ($0.983$), indicating that the model effectively captures the structured visual regularities of printed ECG layouts. These results validate the \texttt{YOLOv11}-based approach as a robust method for automated lead identification and topological reconstruction of ECG sheets, providing a critical foundation for downstream signal extraction and digitization.

\subsection*{Reference Pulse Detection \& Amplitude Scaling}

After segmentation and lead identification, ECG waveforms were quantitatively reconstructed by converting pixel coordinates into physical units using printed reference pulses. These rectangular calibration markers (present on most clinical ECGs) define the vertical and horizontal scales: $1~\mathrm{mV}$ in amplitude and $200~\mathrm{ms}$ in duration. Compared with grid-based calibration, pulses remain visually distinct in degraded scans, enabling robust automated scaling across heterogeneous layouts.

\subsubsection*{Dataset and Model Training}
A dedicated \texttt{YOLOv11x} detector was trained to localize reference pulses. Training used the same synthetic \texttt{PTB-XL} ECG images as the lead-name experiments, augmented with manufacturer-style variations in pulse placement and rectangular templates (width/height/thickness) (\href{tab:dataset}{Table~\ref{tab:dataset}}). Realism was increased with \texttt{Albumentations}~\cite{buslaev_albumentations_2020} (affine transforms, Gaussian noise, contrast changes, motion blur, and occlusions) to emulate fading, scanning noise, and compression artifacts. The model was fine-tuned from a pre-trained \texttt{YOLOv11x} backbone (56.9~M parameters) for 100~epochs using \texttt{AdamW}, a cosine learning-rate schedule, and \texttt{FP16} mixed precision.

\subsubsection*{Amplitude Scaling and Signal Quantification}
Detected pulse regions were processed to estimate pixel-based height and width, yielding vertical and horizontal scale factors. Pulse boundaries were extracted via \texttt{Otsu multi-thresholding}~\cite{otsu_threshold_1979} and \texttt{morphological opening}~\cite{gonzalez_digital_2009} (kernel $1\times25$) to suppress grid/background artifacts while preserving vertical edges. A \texttt{Probabilistic Hough Line Transform}~\cite{kiryati_probabilistic_1991} and \texttt{Line Segment Detector (LSD)}~\cite{grompone_von_gioi_lsd_2010} then measured pulse height $h_{\text{pixels}}$ and width $w_{\text{pixels}}$ (\href{fig:ref}{Figure~\ref{fig:ref}}). An adaptive envelope-based peak detection method~\cite{Scholkmann2012} refined the vertical displacement to reduce errors from clipping or faint printing. Final scaling factors were computed as $S_V = 1~\mathrm{mV}/h_{\text{pixels}}$ and $S_H = 200~\mathrm{ms}/w_{\text{pixels}}$, and applied uniformly to all segmented leads.

\subsection*{Digitization Quality Evaluation and Diagnostic-Specific Preparation}
\subsubsection*{Dataset and End-to-End Protocol}
Digitization fidelity was assessed by comparing reconstructed signals against ground-truth \texttt{PTB-XL} waveforms on a held-out test set of \textbf{1,600} ECG images from \textbf{1,574} patients spanning \texttt{NORM}, \texttt{MI}, \texttt{STTC}, \texttt{CD}, \texttt{HYP}, and \texttt{OTHER} (\href{tab:digitized_ptbxl}{Table~\ref{tab:digitized_ptbxl}}). The cohort had mean age $62.26 \pm 31.17$ years and a balanced sex distribution (51.6\% male, 48.4\% female). To emulate realistic acquisition artifacts, images were augmented with Gaussian noise, motion blur, rotation, and varying grid visibility. Each image was processed end-to-end with patched segmentation, lead-name detection/layout identification, and reference pulse--based calibration to recover physically scaled time-series (mV, ms).

\subsubsection*{Evaluation Metrics}
We report Pearson correlation ($r$), RMSE, and SNR to quantify morphology agreement, amplitude accuracy, and residual noise.

The \textbf{Pearson correlation ($r$)} measures linear similarity between reconstructed ($y$) and reference ($\hat{y}$) signals~\cite{benesty_pearson_2009}:
\begin{equation}
r = \frac{\sum (x_i - \bar{x})(y_i - \bar{y})}{\sqrt{\sum (x_i - \bar{x})^2} \sqrt{\sum (y_i - \bar{y})^2}}
\label{eq:pearson}
\end{equation}

The \textbf{RMSE} measures average amplitude error~\cite{smith_scientist_1997}:
\begin{equation}
\text{RMSE} = \sqrt{\frac{1}{n} \sum_{i=1}^{n} (y_i - \hat{y}_i)^2}
\label{eq:rmse}
\end{equation}

The \textbf{SNR} measures signal energy relative to reconstruction error (dB):
\begin{equation}
\text{SNR} = 10 \log_{10} \left( \frac{\sum y_i^2}{\sum (y_i - \hat{y}_i)^2} \right)
\label{eq:snr}
\end{equation}

\subsubsection*{Quantitative Performance}
Across the multi-pathology, degradation-augmented test set, our \texttt{YOLOv11x}-based digitizer compared favorably to challenge-leading baselines such as \texttt{SignalSavants} and \texttt{BAPORLab}, and showed strong stability under deterioration (\href{tab:combined_results}{Table~\ref{tab:combined_results}}).

\subsubsection*{Diagnostic-Specific Processing and Heartbeat Feature Preparation}
For diagnostic-focused experiments, we digitized two \texttt{PTB-XL} subsets: \textit{Normal} (1,307 ECGs) and \textit{MI} (1,182 ECGs) (\href{fig:pipeline}{Figure~\ref{fig:pipeline}}). Images were standardized (resolution normalization, contrast enhancement) and decomposed into overlapping patches; each patch was segmented by \texttt{YOLOv11x} into (i) ECG traces, (ii) lead regions, and (iii) calibration pulses. Patch masks were stitched with overlap-aware merging and lightly cleaned morphologically to suppress isolated noise and reconnect fragmented traces.

Leads were vectorized by skeletonizing each lead region and sampling the centerline to form an ordered $(x,y)$ sequence. A monotonic time constraint (left-to-right sorting with local continuity fixes) enforced a single-valued waveform per lead. Physical calibration used the detected reference pulse for amplitude scaling (mV/pixel) and the known sweep rate encoded in the template plus measured lead width for time scaling, avoiding reliance on grid spacing and remaining robust when grid lines were faint, occluded, or missing. Signals were resampled to the canonical \texttt{PTB-XL} sampling rate (500~Hz) and mapped back to the standard 12-lead ordering.

For \texttt{MLP}+\texttt{SHAP} analysis, record-level signals were converted to heartbeat windows using \texttt{R}-peak detection~\cite{9707247}, extracting fixed segments of \textbf{150~ms} pre-\texttt{R} and \textbf{300~ms} post-\texttt{R} (\href{tab:train_test_split}{Table~\ref{tab:train_test_split}}). Each lead was normalized independently (per-sample, per-lead) to reduce gain variability so that \texttt{SHAP} reflects morphology rather than absolute scale (\href{fig:mlp_workflow}{Figure~\ref{fig:mlp_workflow}}).

\subsection*{Benchmarking Digitized ECG-Based Time-Series Classification Models}

\subsubsection*{Digitized Inputs and Dataset Construction}
The digitization models were trained on \texttt{PTB-XL} synthetic paper-ECG renderings and were then used to generate digitized time-series for \emph{both} datasets: \textbf{(a)} for \texttt{PTB-XL}, we digitized the \texttt{PTB-XL} paper renderings and used the recovered signals for \texttt{PTB-XL} classification; \textbf{(b)} for \texttt{ECG-Matrix}, we applied the same \texttt{PTB-XL}-trained digitization pipeline \emph{without re-training} to real hospital paper ECG images, and used the resulting digitized signals for \texttt{ECG-Matrix} classification. All classification models were trained for 200 epochs with batch size 64, using a 75/25 training-testing split; for kernel-based ensemble classifiers, we used 20{,}000 estimators. This design explicitly tests whether a digitizer trained on \texttt{PTB-XL} generalizes to an external, real-world paper ECG distribution; strong downstream accuracy on \texttt{ECG-Matrix} therefore provides evidence that the recovered signals preserve diagnostically useful morphology beyond the training domain.

\subsubsection*{Clinical Endpoints and Label Definition}
We evaluated three clinically distinct binary endpoints (see \href{tab:train_test_split}{Table~\ref{tab:train_test_split}}).

\textbf{\textit{MI vs.~Normal} (\texttt{PTB-XL}):} We adopted the dataset-provided diagnostic superclass labels and restricted this task to recordings labeled \texttt{NORM} (normal) or \texttt{MI} (myocardial infarction), excluding recordings assigned to other superclasses (\texttt{STTC}, \texttt{CD}, \texttt{HYP}, \texttt{OTHER}).\cite{ptb_xl}

\textbf{\textit{Pre-procedural vs.~Post procedural} (\texttt{ECG-Matrix}):} We used paired ECGs from the electrocardiography sub-study of the \texttt{MATRIX} trial (an acute coronary syndrome cohort including unstable angina and myocardial infarction undergoing invasive management).\cite{gragnano_ecg_2019} For each patient, the ECG recorded immediately prior to the index catheterization/percutaneous coronary intervention (PCI) was labeled \textit{pre-procedural}, and the matched ECG recorded after PCI was labeled \textit{post procedural}.

\textbf{\textit{OMI vs.~non-OMI} (\texttt{ECG-Matrix}):} To obtain an occlusion-centric endpoint, we defined occlusive myocardial infarction (OMI) based on coronary angiography. To reduce confounding and better isolate ischemic ECG morphology, we applied hierarchical clinical inclusion/exclusion criteria during label generation: we excluded patients with baseline conduction delays (complete or incomplete bundle branch block), excluded non-emergent staged (planned) catheterizations, and excluded index catheterizations without active PCI treatment (no balloon angioplasty and no stent implantation). In the remaining treated index-catheterization cohort, \textit{OMI} was defined by angiographic evidence of a \textit{total coronary occlusion} (100\% stenosis) in at least one culprit/index lesion, whereas \textit{non-OMI} comprised included patients with no total occlusion in any index lesion.

\subsubsection*{Input Representation and Splitting Strategy}
All experiments use patient-wise splits to avoid subject leakage (\href{tab:train_test_split}{Table~\ref{tab:train_test_split}}). We considered two input representations:
\textbf{(i) full-sequence} recordings ($T\times 12$) and \textbf{(ii) segmented-heartbeat} windows ($280\times 12$; 0.56\,s at 500~Hz) extracted around detected R-peaks after light bandpass filtering. Beat windows were rejected when RR intervals were implausible, and inherited subject identifiers so that splits remain patient-wise.

Due to its larger cohort size, \texttt{PTB-XL} supports benchmarking under \emph{both} representations: the full-sequence setting comprised \textbf{1,866} train / \textbf{623} test records (input \textbf{$1,633\times 12$}), while the segmented-heartbeat setting comprised \textbf{7,045} train / \textbf{2,349} test beats (input \textbf{$280\times 12$}). In contrast, \texttt{ECG-Matrix} contains fewer records, so benchmarking on \texttt{ECG-Matrix} was performed \emph{only} using segmented-heartbeat windows to increase the effective number of training samples: \textbf{Pre-procedural vs.~Post procedural} comprised \textbf{1,032} train / \textbf{344} test beats, and \textbf{OMI vs.~non-OMI} comprised \textbf{1,997} train / \textbf{666} test beats (all with input \textbf{$280\times 12$}).

\subsubsection*{Preprocessing}
All models operated on the digitized time-series directly (no hand-crafted ECG features). Prior to model fitting, signals were resampled/standardized to a uniform 500~Hz representation and z-score normalized per lead to reduce inter-recording amplitude offsets. To mitigate digitization-specific artifacts, we applied light denoising and baseline-drift suppression (low-frequency trend removal) to attenuate residual grid remnants, scanner noise, and slow baseline wander without distorting ST segments.

\subsubsection*{Models and Training Protocol}
The benchmark suite spans compact deep architectures (\texttt{MLP}, \texttt{CNN}, \texttt{GRU}, \texttt{MCDCNN}, \texttt{LSTM-FCN}, \texttt{ResNet}, \texttt{InceptionTime}) and kernel-based time-series models (\texttt{Rocket}, \texttt{Arsenal}) \cite{Rumelhart1986,LeCun1998CNN,Cho2014GRU,Cui2016MCDCNN,Karim2018LSTMFCN,Wang2017TSCResNet,Fawaz2020InceptionTime,Dempster2020,Middlehurst2021}. When class imbalance was present, stratified sampling and/or loss reweighting were used. Deep models were trained with batch size 64 for up to 100 epochs (with early stopping as described above), using standard optimizers and learning-rate scheduling; model selection was performed on a held-out validation split derived from the training set. \texttt{Rocket} computes randomized multi-scale convolutional features followed by a linear classifier, providing a strong accuracy--latency baseline under CPU-only inference.

\subsubsection*{Evaluation Metrics and Latency}
We report standard classification metrics (accuracy, precision, sensitivity/recall, specificity, F1) alongside threshold-free summaries (AUROC/AUPRC when applicable) and uncertainty estimates via repeated runs or bootstrap confidence intervals. To quantify deployability, we additionally report \textbf{per-sample CPU inference latency} (see \href{tab:performance_comparison}{Table~\ref{tab:performance_comparison}}), measured in a batch-of-one setting with a fixed thread configuration after warm-up; we report model-only latency to enable fair comparison across architectures.

\subsection*{Interpretability Analysis Using MLP \& SHAP}
We applied a consistent \texttt{MLP}+\texttt{SHAP} interpretability workflow to all three binary endpoints: \textbf{(i)} \texttt{PTB-XL} \textit{Normal} (\textbf{$N=1{,}307$}) vs.\ \textit{MI} (\textbf{$N=1{,}182$}), \textbf{(ii)} \texttt{ECG-Matrix} \textit{pre-procedural} vs.\ \textit{post procedural} (\textbf{$N=1{,}376$} beats), and \textbf{(iii)} \texttt{ECG-Matrix} \textit{OMI} vs.\ \textit{non-OMI} (\textbf{$N=2{,}663$} beats) (see \href{tab:train_test_split}{Table~\ref{tab:train_test_split}}). Inputs were the digitized, physically calibrated 12-lead time-series produced by the proposed pipeline (mV and seconds). For localized attributions, we derived heartbeat windows via R-peak detection, retaining \textbf{150~ms} pre-R and \textbf{300~ms} post-R context; beats retained patient identifiers to ensure patient-wise, leakage-free splits and attribution analyses.

\subsubsection*{Model Definition}
For each endpoint, we trained a separate \texttt{MLP} on the 12-lead R-aligned beat windows. The \texttt{MLP} was selected for simplicity and \texttt{SHAP} compatibility. Each input feature corresponds to a specific \textbf{(lead, time)} sample, enabling aggregation of \texttt{SHAP} values \textbf{per lead} (lead importance) or \textbf{over time within a lead} (heartbeat-phase importance) (see \href{fig:mlp_workflow}{Figure~\ref{fig:mlp_workflow}}).

\subsubsection*{Training Protocol}
We used patient-wise splitting with \textbf{25\%} of patients held out for testing (fixed seed). The \texttt{MLP} used one hidden layer (100 neurons), \texttt{ReLU}, \texttt{Adam}, and was trained for \textbf{200 epochs} with \textbf{batch size 64}, using early stopping to limit overfitting.

\subsubsection*{Heartbeat Segmentation and Normalization}
R-peaks were detected per record to segment beats (150~ms pre-R, 300~ms post-R). Beat windows with implausible surrounding RR-intervals were rejected. Signals were z-score normalized per lead using training-split statistics only, reducing gain variability and keeping \texttt{SHAP} attributions morphology-driven.

\subsubsection*{SHAP Attribution (Lead Importance and R-aligned Heartbeat-Phase Importance)}
\texttt{SHAP} values were computed using a background of 50 training beats and evaluated on 100 randomly selected test beats. Because inputs are a fixed flattened \textbf{(lead, time)} ordering, attributions were mapped back to lead and R-relative latency.

\textbf{Lead importance.} We summed absolute \texttt{SHAP} values across time within each lead and averaged across test beats to rank leads (per endpoint).

\textbf{Within-beat importance.} We aggregated absolute \texttt{SHAP} values at each time index (per lead and pooled) to form time-resolved attribution profiles. Peaks were interpreted relative to clinically meaningful phases around the R-peak (QRS, ST segment, T-wave), indicating whether predictions were driven by depolarization (QRS) or repolarization/ischemia (ST/T) (see \href{fig:mlp_shap}{Figure~\ref{fig:mlp_shap}}).

\subsubsection*{Cross-validating Lead Importance via Per-Lead Classification Performance}
To verify that \texttt{SHAP}-derived lead rankings reflect predictive signal, we performed a per-lead performance analysis (see \href{tab:per_lead_performance}{Supplementary Table~\ref{tab:per_lead_performance}}). We trained otherwise identical models using \textbf{single-lead} inputs (one model per lead) with the same splitting and preprocessing, and compared AUROC/F1 against the \texttt{SHAP} lead ranking.

Overall, across endpoints, attributions concentrated on physiologically relevant morphology within the R-aligned window (QRS, ST, T-wave) rather than baseline artifacts (see \href{fig:mlp_shap}{Figure~\ref{fig:mlp_shap}}).

\section{RESULTS}
\subsection*{\texttt{YOLO}-Based Patched Segmentation For ECG Signal Segmentation}\label{sec:patched-seg}
We evaluated \texttt{YOLOv11} instance segmentation\cite{jocher_ultralytics_2023} under two preprocessing regimes: \textit{Full-Image} (whole ECG page, preserving global inter-lead context) and \textit{Patched} (partially overlapping tiles, increasing sample diversity and emphasizing local waveform detail). Experiments used \texttt{PTB-XL}~\cite{ptb_xl} (12-lead, 10~s, 500~Hz). \textit{Full-Image} used \textit{Subset A} (\textbf{21,799} recordings; \textbf{18,869} patients), while \textit{Patched} used \textit{Subset B} (\textbf{11,000} recordings) expanded into \textbf{323,000} overlapping patches (see \href{tab:dataset}{Table~\ref{tab:dataset}}).

Although \textit{Full-Image} achieved very high box metrics (precision \textbf{0.995}, recall \textbf{0.991}, mAP@50 \textbf{0.994}; Table~\ref{tab:results}A), it missed fine trace structure on high-resolution/noisy sheets. \textit{Patched} slightly reduced box detection (precision \textbf{0.953}, recall \textbf{0.890}, mAP@50 \textbf{0.934}) but substantially improved mask fidelity after stitching (\textbf{IoU=0.647}, \textbf{Dice=0.782} vs. \textbf{0.221}/\textbf{0.353} for full-image), indicating better boundary preservation in low-contrast scans. Under the official \textit{deteriorated} PhysioNet test condition (Supplementary Table~\ref{tab:combined_results}), our end-to-end pipeline achieved \textbf{4.54~dB} SNR with strong waveform agreement (\textbf{Pearson r=0.806}; \textbf{RMSE=0.043~mV}), outperforming reported baselines (e.g., \texttt{SignalSavants} \textbf{3.479~dB}, \texttt{Ahus AI Lab} \textbf{2.777~dB}, \texttt{USST\_Med} \textbf{-0.058~dB}).

We also tested \texttt{YOLOv12x}\cite{jocher_ultralytics_2023} for patched segmentation, but \texttt{YOLOv11x} remained superior in precision/recall and spatial coherence (Table~\ref{tab:results}A). Overall, patched \texttt{YOLOv11x} provides accurate segmentation at practical CPU speed (\,$\sim$\textbf{0.3~s/image}).

\begin{table*}[!htbp]
\centering
\footnotesize 
\begin{subtable}[t]{\textwidth}
\centering
\begin{tabular}{lcccccc}
\toprule
\textbf{Model Variant} & \textbf{Precision} & \textbf{Recall} & \textbf{mAP@50} & \textbf{mAP@50–95} & \textbf{IoU (Mask)} & \textbf{Dice (Mask)} \\
\midrule
\texttt{YOLOv11x} (Unpatched, Box)  & 0.995 & 0.991 & 0.994 & 0.947 & --    & --    \\
\texttt{YOLOv11x} (Unpatched, Mask) & 0.994 & 0.990 & 0.994 & 0.811 & 0.221 & 0.353 \\
\hline
\texttt{YOLOv12x} (Patched, Box)    & 0.618 & 0.594 & 0.619 & 0.453 & --    & --    \\
\texttt{YOLOv12x} (Patched, Mask)   & 0.602 & 0.562 & 0.586 & 0.287 & --    & --    \\
\hline
\textbf{\texttt{YOLOv11x} (Patched, Box)}  & \textbf{0.953} & \textbf{0.890} & \textbf{0.934} & \textbf{0.733} & --    & --    \\
\textbf{\texttt{YOLOv11x} (Patched, Mask)} & \textbf{0.926} & \textbf{0.860} & \textbf{0.910} & \textbf{0.593} & \textbf{0.647} & \textbf{0.782} \\
\bottomrule
\end{tabular}
\subcaption{\textbf{\texttt{YOLO}-Based Image Segmentation Model Performance:} Comparison of YOLO-based models on \texttt{PTB-XL}. \texttt{YOLOv11x} (Patched) achieved the best overall segmentation results, with mask-based overlap metrics reaching IoU=0.647 and Dice=0.782 on noisy digitized ECG signals.}
\label{tab:sub_segmentation}
\end{subtable}

\vspace{10pt}

\begin{minipage}[t]{0.48\textwidth}
\centering
\begin{tabular}{lcc}
\toprule
\textbf{Metric} & \textbf{Lead Detection} & \textbf{Reference Detection} \\
\midrule
Precision  & 0.997 & 0.999 \\
Recall     & 0.991 & 0.999 \\
mAP@50     & 0.995 & 0.995 \\
mAP@50–95  & 0.930 & 0.987 \\
\bottomrule
\end{tabular}
\subcaption{\textbf{\texttt{YOLOv11x} Lead \& Reference Pulse Detection Model Performance:} Performance for \texttt{YOLOv11x} on \texttt{PTB-XL} (bounding-box based evaluation).}
\label{tab:sub_leaddetect}
\end{minipage}
\hfill
\begin{minipage}[t]{0.48\textwidth}
\centering
\begin{tabular}{lcc}
\toprule
\textbf{Layout} & \textbf{Correct (/400)} & \textbf{Accuracy} \\
\midrule
$3\times4$  & 398 & 0.995 \\
$4\times3$  & 338 & 0.845 \\
$6\times2$  & 393 & 0.983 \\
$12\times1$ & 368 & 0.920 \\
\bottomrule
\end{tabular}
\subcaption{\textbf{Layout Detection Accuracy:} Overall accuracy of Layout identification workflow using \texttt{YOLOv11x} Lead Name Detection model across four ECG configurations ($N=1600$).}
\label{tab:sub_layout}
\end{minipage}

\vspace{10pt}

\begin{subtable}[t]{\textwidth}
\centering
\begin{tabular}{lcccc}
\toprule
\textbf{Lead} & \textbf{Pearson} & \textbf{RMSE (mV)} & \textbf{SNR (dB)} & \textbf{$p$-value} \\
\midrule
I   & 0.813 & 0.016 & 4.40 & $8.35\times10^{-111}$ \\
II  & 0.819 & 0.017 & 4.41 & $8.11\times10^{-133}$ \\
III & 0.835 & 0.012 & 4.81 & $1.23\times10^{-99}$ \\
aVR & 0.813 & 0.013 & 4.42 & $4.90\times10^{-120}$ \\
aVL & 0.833 & 0.009 & 4.65 & $6.65\times10^{-22}$ \\
aVF & 0.841 & 0.010 & 4.85 & $4.85\times10^{-123}$ \\
V1  & 0.859 & 0.018 & 5.78 & $6.37\times10^{-158}$ \\
V2  & 0.844 & 0.041 & 5.59 & $9.26\times10^{-160}$ \\
V3  & 0.819 & 0.050 & 4.97 & $5.95\times10^{-139}$ \\
V4  & 0.785 & 0.057 & 4.14 & $6.11\times10^{-107}$ \\
V5  & 0.780 & 0.048 & 3.90 & $6.69\times10^{-121}$ \\
V6  & 0.805 & 0.032 & 4.39 & $4.04\times10^{-114}$ \\
\midrule
\textbf{Overall Avg.} & \textbf{0.806} & \textbf{0.043} & \textbf{4.54} & \textbf{11.94\% Failure Rate} \\
\bottomrule
\end{tabular}
\subcaption{\textbf{Lead-Wise Digitization Performance On Final Evaluation Subset ($N=1600$):}  
Metrics computed against \texttt{PTB-XL} ground truth show consistent correlation across all 12 leads, with a mean Pearson coefficient of 0.806, RMSE of 0.043~mV, and SNR of 4.54~dB. Failures (11.94\%) were primarily attributed to overlapping traces and baseline drift.}
\label{tab:sub_digitization}
\end{subtable}

\vspace{10pt}

\caption{\textbf{Detection, Segmentation, \& Digitization Results On \texttt{PTB-XL}.}  
\textbf{A.} Segmentation comparison between \texttt{YOLOv11x} and \texttt{YOLOv12x}.  
\textbf{B.} Lead and reference pulse detection metrics.  
\textbf{C.} Layout configuration accuracy.  
\textbf{D.} Per-lead digitization evaluation, presenting detailed correlation, error, and signal-to-noise metrics across standard ECG leads.  
The \texttt{YOLOv11x}-based pipeline demonstrated high segmentation precision and robust waveform reconstruction across heterogeneous paper ECG conditions.
}
\label{tab:results}
\end{table*}

\subsection*{Lead Name Detection \& ECG Layout Identification}\label{sec:lead-layout}
To map segmented traces to individual leads, we trained a dedicated \texttt{YOLOv11x} detector to localize lead-name text on full ECG pages (\texttt{PTB-XL}~\cite{ptb_xl}; see \href{tab:dataset}{Table~\ref{tab:dataset}}). Training on full pages preserves global lead--label geometry; robustness was improved with the same synthetic augmentations used elsewhere (font/orientation changes, rotation, creases, noise; Figure~\ref{fig:objdet}). Lead-name detection was near-perfect (precision \textbf{0.997}, recall \textbf{0.991}, mAP@50 \textbf{0.995}; Table~\ref{tab:results}B), and remained reliable under typographic and geometric distortions~\cite{Liao2018, Khan2020}.

Layout inference combined lead-name bounding boxes with segmentation masks to assign each sheet to \textbf{one of 16} standard templates (e.g., 3$\times$4, 4$\times$3 \texttt{Cabrera}, 6$\times$2, 12$\times$1; Figures~\ref{fig:layout}--\ref{fig:cabrera}). On \textbf{1,600} synthetic samples, overall accuracy was \textbf{93\%}, with highest performance on 3$\times$4 (\textbf{0.995}) and 6$\times$2 (\textbf{0.983}) and lower accuracy on 4$\times$3 \texttt{Cabrera} (\textbf{0.845}; mainly aVF/V6 confusion) and 12$\times$1 (\textbf{0.920}; overlap/compression affecting row detection) (Table~\ref{tab:results}C).

\subsection*{Reference Pulse Detection \& Scaling Estimation}\label{sec:ref-pulse}

Reference pulses were detected using a \texttt{YOLOv11}-based model (precision/recall \textbf{1.000}, mAP@50 \textbf{0.995}; Table~\ref{tab:results}B). Within each detected region, a peak-detection method (adaptive thresholding + envelope tracking)~\cite{Scholkmann2012} estimated pulse height relative to baseline to derive a pixel-to-mV conversion (typically \textbf{1~mV/10~mm}), providing calibration that does not rely on grid visibility.

Pixel-to-time was estimated by binarizing the pulse with \texttt{Otsu} multi-thresholding~\cite{otsu_threshold_1979}, isolating vertical structure via morphological opening~\cite{gonzalez_digital_2009} (1$\times$25), and measuring pulse dimensions with probabilistic \texttt{Hough}/LSD line detection~\cite{kiryati_probabilistic_1991, grompone_von_gioi_lsd_2010} to obtain pixels per \textbf{1~mV} and \textbf{200~ms} (Figure~\ref{fig:ref}). Finally, centroid-based vectorization extracted the trace centerline per column, with second-derivative reweighting to better preserve peaks (Figure~\ref{fig:vec}B,D).

\subsection*{Overall ECG Digitization Quality}

Digitization fidelity was evaluated against \texttt{PTB-XL} ground truth using Pearson $r$ (Equation~\ref{eq:pearson})~\cite{benesty_pearson_2009}, RMSE (Equation~\ref{eq:rmse}), and SNR (Equation~\ref{eq:snr}) on \textbf{1,600} recordings spanning six diagnostic superclasses (\textbf{1,574} patients; age $62.26\pm31.17$ years; 51.6\% male), including synthetically degraded variants to emulate scanning artifacts (Table~\ref{tab:dataset}C).

Across all 12 leads, the full pipeline (patched segmentation + lead/layout inference + reference-pulse scaling) achieved mean \textbf{$r=0.806$}, \textbf{RMSE=0.043~mV}, and \textbf{SNR=4.54~dB} (\textbf{$p<0.05$}), with best correlations in \texttt{V1--V3} and \texttt{aVF} (Table~\ref{tab:results}D). The overall failure rate was \textbf{11.94\%}, primarily due to overlapping traces and severe baseline drift.

\begin{table*}[t]
\centering
\footnotesize
\begin{subtable}[t]{\textwidth}
\centering
\begin{tabular}{lcc}
\toprule
\textbf{Digitization Metrics} & \textbf{Normal} & \textbf{Myocardial Infarction} \\
\midrule
Number of Images & 1,307 & 1,182 \\
Processing Time (s/image) & 10.92 & 10.69 \\
Total Processing Time (min) & 237.83 & 210.66 \\
RMSE (mV) & 0.0219 & 0.0225 \\
SNR (dB) & 4.8862 & 6.0335 \\
Pearson Correlation ($r$) & 0.8428 & 0.8705 \\
$p$-value & 0.0001 & 0.0001 \\
\bottomrule
\end{tabular}
\subcaption{\textbf{Noisy ECG Image Digitization Performance:}  
Performance and processing statistics for the digitized \texttt{PTB-XL} subsets.  
Results are reported separately for \textit{Normal} and \textit{Myocardial Infarction} samples.  
Average processing time represents \texttt{YOLOv11x}-based inference per image.  
Metrics include RMSE, signal-to-noise ratio (SNR), Pearson correlation coefficient, and statistical significance ($p$-value).}
\label{tab:digitized_ptbxl}
\end{subtable}

\vspace{8pt}

\begin{subtable}[t]{\textwidth}
\footnotesize
\resizebox{\textwidth}{!}{%
\begin{tabular}{
  >{\centering\arraybackslash}p{2.85cm}
  c c
  r r r
  c c
  r r r
}
\toprule
\textbf{Dataset} & \textbf{Setting} & \textbf{Split} & \textbf{Samples} & \makecell[c]{\textbf{Unique}\\\textbf{Subjects}} & \makecell[c]{\textbf{Avg.}\\\textbf{Timesteps}} & \makecell[c]{\textbf{Male}\\(\%)} & \makecell[c]{\textbf{Female}\\(\%)} & \makecell[c]{\textbf{Age}\\(years)} & \makecell[c]{\textbf{Height}\\(m)} & \makecell[c]{\textbf{Weight}\\(kg)} \\
\midrule
\multirow{4}{=}{\makecell[c]{\textbf{\textit{MI vs Normal}}\\(\texttt{PTB-XL})}}
  & \multirow{2}{*}{Full seq.} & Train & 1,866 & 1,866 & \multirow{2}{*}{1,633} & \multirow{4}{*}{52.01\%} & \multirow{4}{*}{47.99\%} & \multirow{4}{*}{\makecell[c]{62.67\\$\pm$\\35.25}} & \multirow{4}{*}{\makecell[c]{1.67\\$\pm$\\0.09}} & \multirow{4}{*}{\makecell[c]{70.40\\$\pm$\\15.12}} \\
  &  & Test  & 623  & 623  &  &  &  &  &  &  \\
\cmidrule(lr){2-6}
  & \multirow{2}{*}{Seg.} & Train & 7,045 & 1,866 & \multirow{2}{*}{280} &  &  &  &  &  \\
  &  & Test  & 2,349 & 623 &  &  &  &  &  &  \\
\specialrule{0.9pt}{0.6ex}{0.6ex}
\multirow{4}{=}{\makecell[c]{\textbf{\textit{Pre-Procedural vs}}\\\textbf{\textit{Post-Procedural MI}}\\(\texttt{ECG-Matrix})}}
  & \multirow{2}{*}{Full seq.} & Train & 176 & 176 & \multirow{2}{*}{3,167} & \multirow{4}{*}{77.69\%} & \multirow{4}{*}{22.31\%} & \multirow{4}{*}{\makecell[c]{66.31\\$\pm$\\12.12}} & \multirow{4}{*}{\makecell[c]{1.69\\$\pm$\\0.08}} & \multirow{4}{*}{\makecell[c]{78.05\\$\pm$\\12.74}} \\
  &  & Test  & 59  & 59  &  &  &  &  &  &  \\
\cmidrule(lr){2-6}
  & \multirow{2}{*}{Seg.} & Train & 1,032 & 176 & \multirow{2}{*}{280} &  &  &  &  &  \\
  &  & Test  & 344 & 59 &  &  &  &  &  &  \\
\specialrule{0.9pt}{0.6ex}{0.6ex}
\multirow{4}{=}{\makecell[c]{\textbf{\textit{OMI vs non-OMI}}\\(\texttt{ECG-Matrix})}}
  & \multirow{2}{*}{Full seq.} & Train & 319 & 319 & \multirow{2}{*}{4,046} & \multirow{4}{*}{76.24\%} & \multirow{4}{*}{23.76\%} & \multirow{4}{*}{\makecell[c]{64.30\\$\pm$\\11.11}} & \multirow{4}{*}{\makecell[c]{1.69\\$\pm$\\0.08}} & \multirow{4}{*}{\makecell[c]{77.71\\$\pm$\\14.35}} \\
  &  & Test  & 107 & 107 &  &  &  &  &  &  \\
\cmidrule(lr){2-6}
  & \multirow{2}{*}{Seg.} & Train & 1,997 & 319 & \multirow{2}{*}{280} &  &  &  &  &  \\
  &  & Test  & 666 & 107 &  &  &  &  &  &  \\
\bottomrule
\end{tabular}%
}
\subcaption{\textbf{Myocardial Infarction Classification Dataset Composition:}  
Dataset split and characteristics for full-sequence and segmented ECG samples (12 leads, 500~Hz).  
To avoid overloading the table with full-sequence details, we primarily highlight the \textbf{number of unique subjects} for the full-sequence settings; segmented rows report sample-level counts and input shapes. Demographics are reported as mean$\pm$SD based on the processed metadata described in the text.}
\label{tab:train_test_split}
\end{subtable}
\vspace{6pt}

\caption{\textbf{Overview Of Digitized ECG Evaluation \& Classification Datasets:}  
Subtable~(A) reports the quantitative results of the \texttt{YOLOv11x}-based ECG digitization pipeline across \textit{Normal} and \textit{Myocardial Infarction} subsets.  
Subtable~(B) details the dataset splits used for training and evaluating myocardial infarction classifiers on segmented and full-sequence ECG signals.  
Together, these datasets form the foundation for segmentation validation, lead name detection, and downstream diagnostic modeling.}
\label{tab:digitization_combined}
\end{table*}

Compared with leading PhysioNet Challenge submissions, our \texttt{YOLOv11x}-based pipeline showed stronger robustness under degradation (see Table~\ref{tab:combined_results}); e.g., \texttt{SignalSavants} reported \textbf{0.514} (deteriorated) and \texttt{BAPORLab} \textbf{0.358}, whereas our approach achieved \textbf{4.54~dB} SNR with \textbf{Pearson $r=0.806$} and \textbf{RMSE=0.043~mV} on degraded sheets. This robustness stems from combining patch-wise segmentation with reference-pulse calibration, yielding a fully automated and physically interpretable digitization pipeline suitable for downstream modeling. While certain competing methods attained higher peak SNRs under idealized clean conditions, their performance degraded significantly in deteriorated scenarios. In contrast, the proposed  digitization pipeline maintained a balanced trade-off between signal fidelity and robustness, yielding consistent results across heterogeneous data sources. These findings confirm that the framework not only reconstructs amplitude-accurate waveforms but also sustains high structural and morphological fidelity, making it a reliable foundation suitable for downstream diagnostic modeling and pathology-specific feature extraction.

\subsection*{\texttt{MLP-SHAP}-Based Feature Analysis \& Pathophysiological Attribution}

For interpretability, we digitized class-stratified \texttt{PTB-XL} cohorts (\textit{Normal} $N=\mathbf{1{,}307}$; \textit{MI} $N=\mathbf{1{,}182}$) and confirmed that reconstructed signals retained high agreement with ground truth (Normal: \textbf{$r=0.8428$}; MI: \textbf{$r=0.8705$}; RMSE \textbf{0.0219--0.0225}~mV; SNR \textbf{4.886--6.034}~dB; all \textbf{$p=0.0001$}), at similar runtime (\textbf{10.92} vs. \textbf{10.69}~s/image; Table~\ref{tab:digitized_ptbxl}). Since \texttt{ECG Matrix} lacks paired waveform ground truth, we digitized all \texttt{ECG Matrix} records using the \texttt{PTB-XL}-validated pipeline and used the reconstructions for downstream analysis.

\begin{figure*}[t]
\centering
\vspace*{-1.0cm} 
\begingroup
\renewcommand{\thesubfigure}{\Alph{subfigure}}

\begin{subfigure}[t]{0.32\textwidth}
  \centering
  \includegraphics[width=\linewidth]{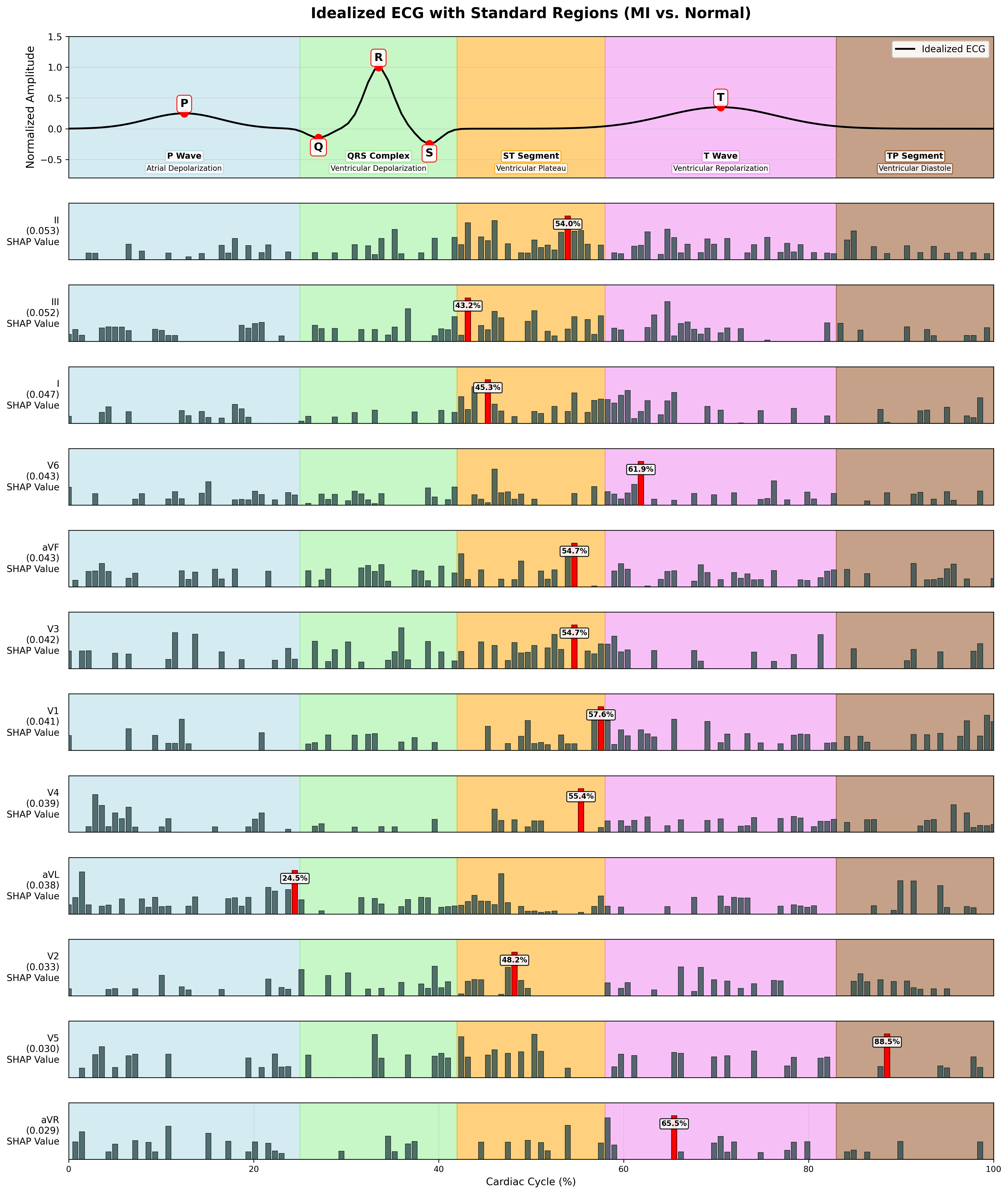}
  \subcaption{\textbf{\texttt{Normal} vs. \texttt{MI} (\texttt{PTB-XL}).} Idealized 12-lead heartbeat with overlaid \texttt{SHAP} importance traces from an \texttt{MLP} trained on digitized \texttt{PTB-XL} signals. Higher-magnitude regions indicate waveform segments (QRS and ST-T) and leads that most strongly drive predictions toward the \texttt{MI} class.}
\end{subfigure}
\hfill
\begin{subfigure}[t]{0.32\textwidth}
  \centering
  \includegraphics[width=\linewidth]{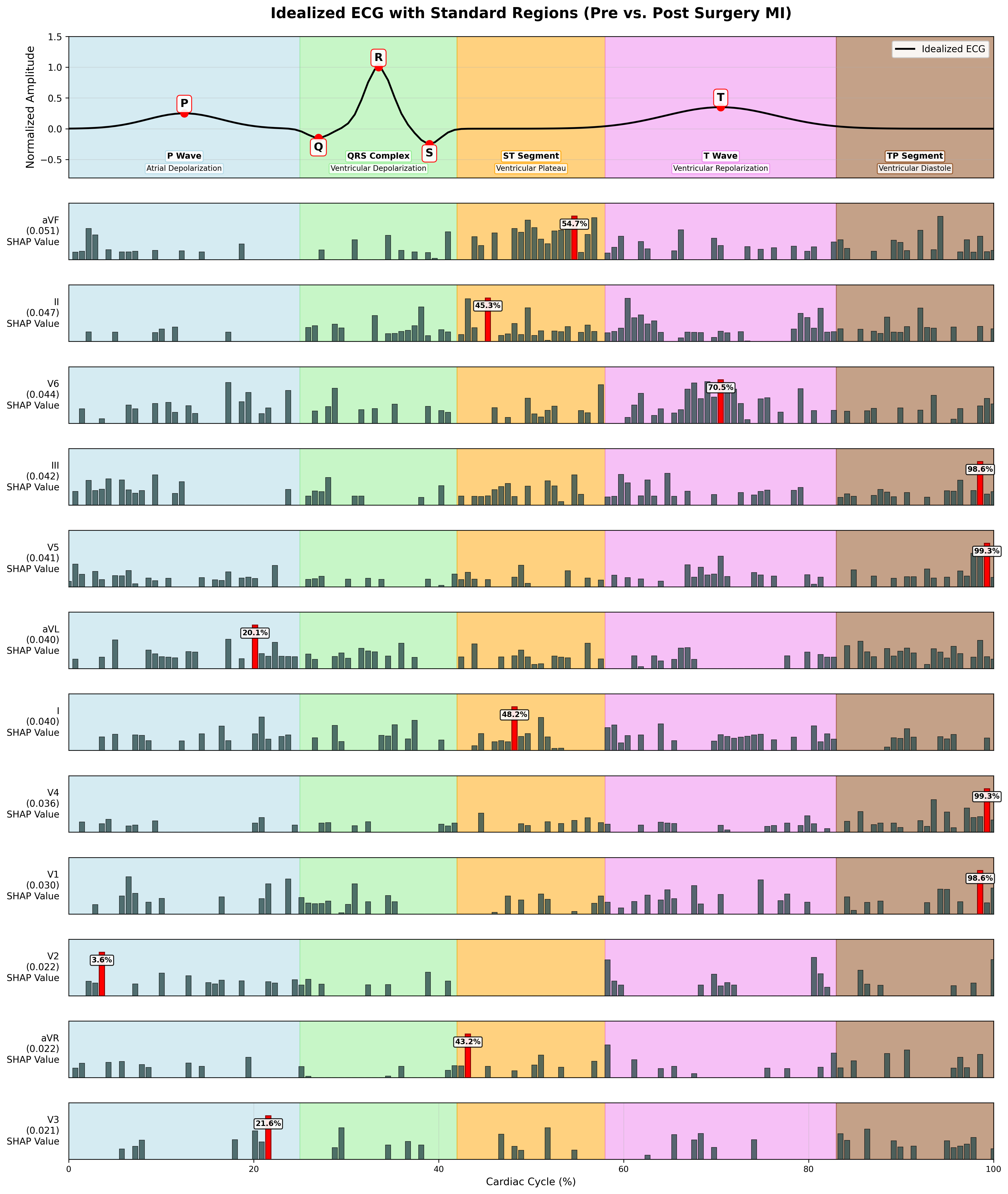}
  \subcaption{\textbf{\texttt{Pre} vs. \texttt{Post} Procedural MI (\texttt{ECG Matrix}).} The same \texttt{MLP}+\texttt{SHAP} visualization for the \texttt{ECG Matrix} pre-/post procedural label definition. The attribution pattern highlights which leads and temporal intervals are most informative for discriminating pre- from post-intervention recordings.}
\end{subfigure}
\hfill
\begin{subfigure}[t]{0.32\textwidth}
  \centering
  \includegraphics[width=\linewidth]{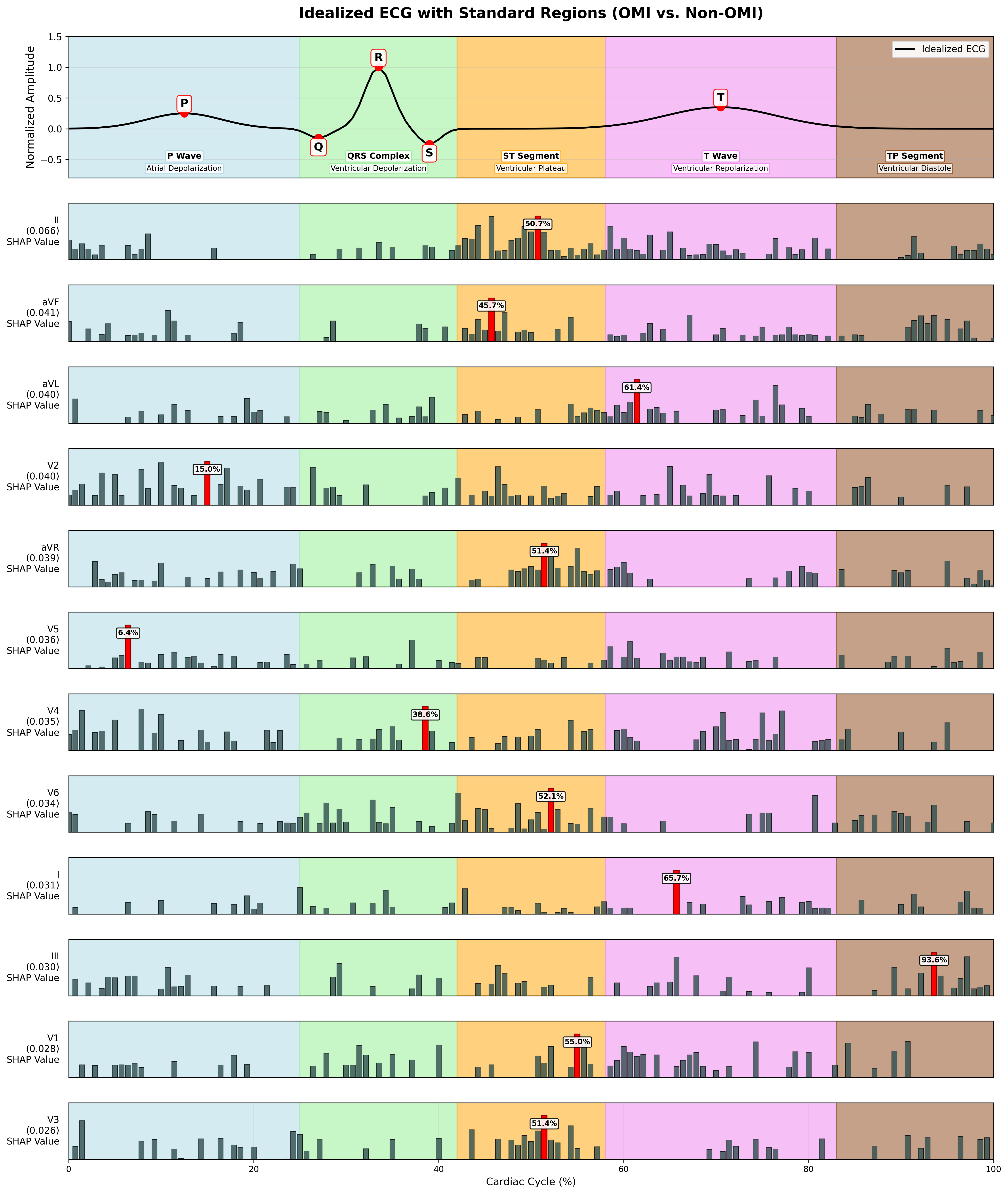}
  \subcaption{\textbf{\texttt{OMI} vs. \texttt{Non-OMI} (\texttt{ECG Matrix}).} \texttt{SHAP} attribution map for occlusion MI labeling on \texttt{ECG Matrix}. Compared with panel~(B), importance is redistributed across leads/time, reflecting morphology that is specific to the \texttt{OMI} definition (repolarization and ST-segment changes).}
\end{subfigure}

\caption{\textbf{Idealized 12-lead \texttt{SHAP} attribution maps for myocardial infarction classification under three clinical label definitions.} Each panel summarizes post-hoc feature attribution from an \texttt{MLP} classifier using \texttt{SHapley Additive exPlanations (SHAP)} on digitized 12-lead ECG time-series (standardized per lead and evaluated on held-out test samples). \texttt{SHAP} values quantify the marginal contribution of each lead’s waveform samples toward the positive class. Panel~(A) reports \texttt{Normal} vs.~\texttt{MI} on \texttt{PTB-XL}; Panel~(B) reports \texttt{Pre} vs.~\texttt{Post} Procedural MI on \texttt{ECG Matrix}; Panel~(C) reports \texttt{OMI} vs.~\texttt{Non-OMI} on \texttt{ECG Matrix}.}
\label{fig:mlp_shap}
\endgroup
\end{figure*}

We then studied three binary tasks: \textit{MI vs Normal} (\texttt{PTB-XL}), \textit{Pre- vs Post-procedural MI} (\texttt{ECG Matrix}), and \textit{OMI vs non-OMI} (\texttt{ECG Matrix}) (Table~\ref{tab:train_test_split}). Explanations were generated with an \texttt{MLP}+\texttt{SHAP} setup: reconstructed recordings were trimmed (first/last \textbf{2\%}), segmented into R-peak-aligned heartbeat windows, z-score normalized per lead, and used to compute timepoint-level \texttt{SHAP} attributions on held-out samples (Figure~\ref{fig:mlp_shap}).

As a quantitative reference point for these explanations, the \texttt{MLP} on \textit{MI} vs.~\textit{Normal} (\texttt{PTB-XL}) achieved an accuracy of Accuracy $\textbf{0.8318}$ (Precision $0.8423$ Recall $0.8188$ Specificity $0.8394$ F1-score $\textbf{0.8297}$ latency $1.25$~ms), On \textit{Pre-procedural MI} vs.~\textit{Post procedural MI} (\texttt{ECG Matrix}) it achieved an accuracy of $\textbf{0.8343}$ (Precision $0.8129$, Recall $0.8235$, Specificity $0.8432$, F1-score $\textbf{0.8182}$, latency $0.26$~ms), and on \textit{OMI} vs.~non-\textit{OMI} (\texttt{ECG Matrix}), it achieved an accuracy of $\textbf{0.8213}$ (Precision $0.8421$ Recall $0.7643$ Specificity $0.8722$ F1-score $\textbf{0.8013}$ latency $0.26$~ms) (see \href{tab:performance_comparison}{Table~\ref{tab:performance_comparison}}).

To compute the \texttt{SHAP} maps, we used \textbf{50} PCA components for dimensionality reduction and \textbf{50} background samples for \texttt{SHAP} baseline estimation, then evaluated \texttt{SHAP} at every timepoint across all \textbf{12} leads. Leads were subsequently ordered (ascending) by the number of timepoints exhibiting the strongest attribution magnitudes, and the resulting lead ranking was used to generate the three attribution visualizations corresponding to the three classification tasks (see \href{fig:mlp_shap}{Figure~\ref{fig:mlp_shap} A-C}).

For \texttt{PTB-XL} (\textit{Normal} vs.~\textit{MI}), the strongest \texttt{SHAP} attributions concentrate around the QRS complex and early ST segment, indicating that the \texttt{MLP} primarily relies on depolarization and early ischemia-sensitive morphology to separate infarction from normal rhythm\cite{goldberger_clinical_2017,thygesen_universal_2018}. While alterations in the QRS complex may reflect past infarction events (pathological Q waves) or conduction delays, the prominence of the ST interval is consistent with its central role in detecting acute ischemic injury\cite{goldberger_clinical_2017,thygesen_universal_2018}. High-attribution regions are typically localized to anatomically informative lead groups (anterior and inferior--lateral channels), yielding a compact set of timepoints that consistently drive positive-class evidence (see \href{fig:mlp_shap}{Figure~\ref{fig:mlp_shap} A}). In \texttt{the ECG Matrix}, the attribution structure changes with the clinical endpoint. In the \textit{Pre-procedural MI} vs.~\textit{Post procedural MI} setting (see \href{fig:mlp_shap}{Figure~\ref{fig:mlp_shap} B}), importance becomes more distributed across the beat, with contributions spanning both QRS and ST--T intervals, consistent with peri-intervention shifts that may not be confined to classic ischemic segments\cite{bhatt_diagnosis_2022}. In the occlusion-focused \textit{OMI} vs.\textit{non-OMI} task (see \href{fig:mlp_shap}{Figure~\ref{fig:mlp_shap} C}), \texttt{SHAP} emphasis shifts toward repolarization-sensitive regions (ST segment and T wave) and appears more coherent across contiguous lead groups, reflecting spatially correlated patterns expected under acute coronary occlusion\cite{meyers_omi_2021}. Overall, the task-dependent redistribution of \texttt{SHAP} supports that the digitized signals preserve physiologically meaningful variation rather than noise-driven artifacts.

The \texttt{SHAP} distributions align with established MI electrophysiology. Anterior leads (\texttt{V1--V4}) show the strongest attributions around the QRS complex and early ST segment (LAD territory), lateral leads (\texttt{I}, \texttt{aVL}, \texttt{V5--V6}) contribute mainly in the ST--T segment (repolarization abnormalities), and inferior leads (\texttt{II}, \texttt{III}, \texttt{aVF}) exhibit moderate importance consistent with inferior-wall MI patterns most often involving the RCA (or LCx depending on dominance)\cite{goldberger_clinical_2017}. Across leads, \texttt{SHAP} peaks localize to the QRS and ST segments rather than the \textit{P} wave or baseline, indicating reliance on depolarization--repolarization abnormalities and supporting that the digitization/scaling preserved clinically meaningful diagnostic features\cite{goldberger_clinical_2017,thygesen_universal_2018}. Despite its shallow architecture, the \texttt{MLP} therefore leverages physiologically grounded information across the 12-lead system, enabling anatomically interpretable model attributions.

To corroborate the attribution maps, we also evaluated \emph{single-lead} classification using the same \texttt{MLP} (see \href{tab:per_lead_performance}{Supplementary Table \ref{tab:per_lead_performance}}). The \textbf{top-ranked leads} by single-lead performance matched those with the highest aggregate \texttt{SHAP} importance, demonstrating agreement between post-hoc explanations and direct predictive utility and increasing confidence that the attributions reflect true MI/OMI morphology rather than spurious artifacts.

\subsection*{Benchmarking Time-Series Classifiers Using Digitized ECGs}

Across experiments, ECGs were sampled at 500~Hz and per-lead z-score normalized. Benchmarking used an \texttt{Intel Core i9-13900H} CPU. Deep models trained for 100 epochs (batch size 64); kernel methods used 20{,}000 kernels; other settings followed \texttt{sktime} defaults\cite{sktime}. With \texttt{SHAP}, per-lead normalization encourages attributions to reflect waveform \emph{shape} (e.g., ST deviation, T-wave asymmetry, QRS notching) rather than amplitude/baseline artifacts.

\textbf{\textit{MI vs.~Normal} (\texttt{PTB-XL}).} On segmented beats, \texttt{InceptionTime} \cite{Fawaz2020InceptionTime} (accuracy \textbf{0.8906}, \textbf{2.94~ms}) and \texttt{MCDCNN} \cite{Cui2016MCDCNN} (\textbf{0.8842}, \textbf{2.75~ms}) were strongest deep baselines; \texttt{LSTM-FCN} \cite{Karim2018LSTMFCN} (\textbf{0.8574}, \textbf{0.79~ms}) and \texttt{ResNet} \cite{Wang2017TSCResNet} (\textbf{0.8544}, \textbf{1.72~ms}) reduced latency at modest cost, while \texttt{GRU} \cite{Cho2014GRU} underperformed (\textbf{0.5683}). Full sequences improved multiple methods (e.g., \texttt{InceptionTime} \textbf{0.9294} at \textbf{11.12~ms}). Kernel models were particularly strong but slower: \texttt{Rocket} \cite{Dempster2020} reached \textbf{0.9551} on full sequences (\textbf{304.05~ms}), and \texttt{Arsenal} \cite{Middlehurst2021} achieved the best full-sequence accuracy (\textbf{0.9583}) at high latency (\textbf{6{,}314.01~ms}), illustrating the accuracy--runtime trade-off. \texttt{SHAP} can help verify that models focus on physiologically plausible regions (ST/early T-wave) rather than digitization seams or borders.

\textbf{\textit{Pre-procedural} vs.~\textit{Post procedural} (\texttt{ECG-Matrix}).} On segmented beats, \texttt{InceptionTime} provided the best balance (accuracy \textbf{0.9053}, recall \textbf{0.8954}, \textbf{2.00~ms}), while \texttt{MCDCNN} was more precision-oriented (precision \textbf{0.9071}, accuracy \textbf{0.8846}). Kernel methods were competitive and high-specificity: \texttt{Rocket} and \texttt{Arsenal} both achieved \textbf{0.8639} accuracy with specificity \textbf{0.9376}/\textbf{0.9338}. Because pre/post status is not a standardized endpoint, performance likely reflects mixed factors (reperfusion, procedural effects, evolving ischemia, medications, inter-ECG interval), so we interpret this task as systematic peri-procedural ECG change in paired recordings rather than intervention success.

\begin{table*}[t]
\centering
\resizebox{\textwidth}{!}{%
\begin{tabular}{@{}>{\centering\arraybackslash}p{3.0cm} c l S[table-format=4.2] S[table-format=1.4] S[table-format=1.4] S[table-format=1.4] S[table-format=1.4] S[table-format=1.4]@{}}
\toprule
\textbf{Dataset} & \textbf{Model Type} & \textbf{Model} & {\textbf{Latency (ms)}} & {\textbf{Accuracy}} & {\textbf{Precision}} & {\textbf{Recall}} & {\textbf{Specificity}} & {\textbf{F1}} \\
\midrule

\multirow{9}{*}{\makecell[c]{\textbf{\textit{MI}}\\\textbf{vs}\\\textbf{\textit{Normal}}\\(\texttt{PTB-XL})\\\textit{Segmented}}}
& \multirow{7}{*}{\textbf{Deep Learning}}
& MLP          & 1.25  & 0.8318 & 0.8423 & 0.8188 & 0.8394 & 0.8297 \\
&  & GRU          & 11.28 & 0.5683 & 0.5774 & 0.5275 & 0.6038 & 0.5513 \\
&  & CNN          & 0.44 & 0.8161 & 0.8433 & 0.7790 & 0.8533 & 0.8099 \\
&  & MCDCNN       & 2.75  & 0.8842 & 0.9076 & 0.8569 & 0.9160 & 0.8815 \\
&  & LSTM-FCN     & 0.79  & 0.8574 & 0.8603 & 0.8552 & 0.8599 & 0.8577 \\
&  & ResNet       & 1.72  & 0.8544 & 0.8570 & 0.8527 & 0.8559 & 0.8548 \\
&  & InceptionTime & 2.94 & 0.8906 & 0.8915 & 0.8908 & 0.8907 & 0.8911 \\
\cmidrule(lr){2-9}
& \multirow{2}{*}{\textbf{Kernel-Based}}
& Rocket       & 17.70 & 0.9106 & 0.9278 & 0.8916 & 0.9297 & 0.9093 \\
&  & \cellcolor{cyan!15}\textbf{Arsenal}
& \multicolumn{1}{>{\columncolor{cyan!15}}S[table-format=4.2]}{295.20}
& \multicolumn{1}{>{\columncolor{cyan!15}}S[table-format=1.4]}{\textbf{0.9221}}
& \multicolumn{1}{>{\columncolor{cyan!15}}S[table-format=1.4]}{\textbf{0.9408}}
& \multicolumn{1}{>{\columncolor{cyan!15}}S[table-format=1.4]}{\textbf{0.9018}}
& \multicolumn{1}{>{\columncolor{cyan!15}}S[table-format=1.4]}{\textbf{0.9411}}
& \multicolumn{1}{>{\columncolor{cyan!15}}S[table-format=1.4]}{\textbf{0.9209}} \\

\midrule
\multirow{8}{*}{\makecell[c]{\textbf{\textit{MI}}\\\textbf{vs}\\\textbf{\textit{Normal}}\\(\texttt{PTB-XL})\\\textit{Full sequence}}}
& \multirow{6}{*}{\textbf{Deep Learning}}
& GRU          & 33.58 & 0.5875 & 0.5666 & 0.5608 & 0.6103 & 0.5637 \\
&  & CNN          & 0.64 & 0.7897 & 0.7405 & 0.8581 & 0.7402 & 0.7950 \\
&  & MCDCNN       & 1.83  & 0.7769 & 0.7774 & 0.7432 & 0.8017 & 0.7599 \\
&  & LSTM-FCN     & 4.45  & 0.8796 & 0.8553 & 0.8986 & 0.8585 & 0.8764 \\
&  & ResNet       & 7.45  & 0.8989 & 0.8896 & 0.8986 & 0.8986 & 0.8941 \\
&  & InceptionTime & 11.12 & 0.9294 & 0.9468 & 0.9020 & 0.9585 & 0.9239 \\
\cmidrule(lr){2-9}
& \multirow{2}{*}{\textbf{Kernel-Based}}
& Rocket       & 304.05 & 0.9551 & 0.9685 & 0.9358 & \textbf{0.9845} & 0.9519 \\
&  & \cellcolor{cyan!15}\textbf{Arsenal}
& \multicolumn{1}{>{\columncolor{cyan!15}}S[table-format=4.2]}{6314.01}
& \multicolumn{1}{>{\columncolor{cyan!15}}S[table-format=1.4]}{\textbf{0.9583}}
& \multicolumn{1}{>{\columncolor{cyan!15}}S[table-format=1.4]}{\textbf{0.9688}}
& \multicolumn{1}{>{\columncolor{cyan!15}}S[table-format=1.4]}{\textbf{0.9426}}
& \multicolumn{1}{>{\columncolor{cyan!15}}S[table-format=1.4]}{0.9842}
& \multicolumn{1}{>{\columncolor{cyan!15}}S[table-format=1.4]}{\textbf{0.9555}} \\

\midrule
\multirow{9}{*}{\makecell[c]{\textbf{\textit{Pre-Procedural MI}}\\\textbf{vs}\\\textbf{\textit{Post-Procedural MI}}\\(\texttt{ECG-Matrix})\\\textit{Undersampled}}}
& \multirow{7}{*}{\textbf{Deep Learning}}
& MLP          & 0.26 & 0.8343 & 0.8129 & 0.8235 & 0.8432 & 0.8182 \\
&  & GRU           & 2.44  & 0.5651 & 0.5188 & 0.5425 & 0.5928 & 0.5304 \\
&  & CNN           & 0.32  & 0.7160 & 0.7143 & 0.6209 & 0.7534 & 0.6643 \\
&  & MCDCNN        & 2.04  & 0.8846 & \textbf{0.9071} & 0.8301 & 0.9206 & 0.8669 \\
&  & LSTM-FCN      & 0.78  & 0.8669 & 0.8600 & 0.8431 & 0.8900 & 0.8515 \\
&  & ResNet        & 1.85  & 0.8373 & 0.8403 & 0.7908 & 0.8555 & 0.8148 \\
&  & \cellcolor{cyan!15}\textbf{InceptionTime}
& \multicolumn{1}{>{\columncolor{cyan!15}}S[table-format=4.2]}{2.00}
& \multicolumn{1}{>{\columncolor{cyan!15}}S[table-format=1.4]}{\textbf{0.9053}}
& \multicolumn{1}{>{\columncolor{cyan!15}}S[table-format=1.4]}{0.8954}
& \multicolumn{1}{>{\columncolor{cyan!15}}S[table-format=1.4]}{\textbf{0.8954}}
& \multicolumn{1}{>{\columncolor{cyan!15}}S[table-format=1.4]}{0.9135}
& \multicolumn{1}{>{\columncolor{cyan!15}}S[table-format=1.4]}{\textbf{0.8954}} \\
\cmidrule(lr){2-9}
& \multirow{2}{*}{\textbf{Kernel-Based}}
& Rocket        & 342.11  & 0.8639 & 0.8794 & 0.8105 & \textbf{0.9376} & 0.8435 \\
&  & Arsenal       & 1462.74 & 0.8639 & 0.8741 & 0.8170 & 0.9338 & 0.8446 \\

\midrule
\multirow{9}{*}{\makecell[c]{\textbf{\textit{OMI}}\\\textbf{vs}\\\textbf{\textit{non-OMI}}\\(\texttt{ECG-Matrix})}}
& \multirow{7}{*}{\textbf{Deep Learning}}
& MLP          & 0.26 & 0.8213 & 0.8421 & 0.7643 & 0.8722 & 0.8013 \\
&  & GRU           & 7.00 & 0.5691 & 0.5440 & 0.5318 & 0.5900 & 0.5378 \\
&  & CNN           & 0.26 & 0.7012 & 0.6825 & 0.6847 & 0.7159 & 0.6836 \\
&  & MCDCNN        & 0.90 & 0.8544 & 0.8401 & 0.8535 & 0.8553 & 0.8468 \\
&  & LSTM-FCN      & 0.69 & 0.8529 & \textbf{0.8673} & 0.8121 & 0.8892 & 0.8388 \\
&  & ResNet        & 1.01 & 0.8063 & 0.8033 & 0.7803 & 0.8295 & 0.7916 \\
&  & InceptionTime & 1.08 & 0.8634 & 0.8474 & 0.8662 & 0.8608 & 0.8567 \\
\cmidrule(lr){2-9}
& \multirow{2}{*}{\textbf{Kernel-Based}}
& \cellcolor{cyan!15}\textbf{Rocket}
& \multicolumn{1}{>{\columncolor{cyan!15}}S[table-format=4.2]}{540.81}
& \multicolumn{1}{>{\columncolor{cyan!15}}S[table-format=1.4]}{\textbf{0.8889}}
& \multicolumn{1}{>{\columncolor{cyan!15}}S[table-format=1.4]}{0.8571}
& \multicolumn{1}{>{\columncolor{cyan!15}}S[table-format=1.4]}{\textbf{0.9172}}
& \multicolumn{1}{>{\columncolor{cyan!15}}S[table-format=1.4]}{0.8636}
& \multicolumn{1}{>{\columncolor{cyan!15}}S[table-format=1.4]}{\textbf{0.8862}} \\
&  & Arsenal       & 2741.94 & 0.8829 & 0.8554 & 0.9045 & 0.8633 & 0.8793 \\

\bottomrule
\end{tabular}%
}

\caption{\textbf{ECG time-series classifier benchmarking on digitized 12-lead signals.} Deep-learning and kernel-based models are compared by per-sample CPU latency and standard metrics across four blocks: \texttt{PTB-XL} (MI vs.~Normal) on segmented beats and full sequences, \texttt{ECG-Matrix} (Pre- vs.~Post-procedural MI) under undersampling/balancing, and \texttt{ECG-Matrix} (OMI vs.~non-OMI) on segmented beats. Best-in-column values are bold; the best model per block is highlighted (cyan): \texttt{Arsenal} \cite{Middlehurst2021} (\texttt{PTB-XL}), \texttt{InceptionTime} \cite{Fawaz2020InceptionTime} (pre/post), and \texttt{Rocket} \cite{Dempster2020} (OMI). Full-sequence context improves \texttt{PTB-XL}, while \texttt{ECG-Matrix} tasks are harder and reward models capturing subtle ST--T morphology at low latency.}
\label{tab:performance_comparison}
\end{table*}

\textbf{\textit{OMI vs.~non-OMI} (\texttt{ECG-Matrix}).} The lowest-latency baseline was \texttt{MLP} \cite{Rumelhart1986} (\textbf{0.26~ms}, accuracy \textbf{0.8213}). \texttt{InceptionTime} improved accuracy to \textbf{0.8634} (\textbf{1.08~ms}), while \texttt{Rocket} \cite{Dempster2020} provided the best overall discrimination (accuracy \textbf{0.8889}, recall \textbf{0.9172}, F1 \textbf{0.8862}). Clinically, high recall is important because missed occlusion delays reperfusion; \texttt{SHAP}-style lead/time attributions can support clinician review by highlighting which ST--T segments (and leads) drove an ``OMI'' decision.

\section{DISCUSSION}

We extend learning-based digitization approaches \cite{Tomaszewski2020, Pham2022, wu_fully-automated_2022} by using patched \texttt{YOLOv11} instance segmentation with padding and multi-pass patch fusion, improving pixel-level boundary fidelity in high-resolution or degraded sheets (\textbf{IoU $=0.647$}, \textbf{Dice $=0.782$} on \texttt{PTB-XL}) while reducing seam artifacts during full-page recomposition. By explicitly segmenting inked waveforms (rather than relying on hand-tuned binarization), the pipeline avoids the recurring problem of \textbf{manual thresholding} that often accompanies grid suppression and foreground extraction in classic workflows, while keeping compute demands modest.

We further reduce reliance on grid visibility by using \textbf{reference-pulse detection} for physically grounded calibration, and we integrate \textbf{waveform segmentation}, \textbf{lead-name detection}, and \textbf{explicit layout inference} across common report formats, including challenging variants such as \texttt{Cabrera} (see \href{tab:results}{Table~\ref{tab:results}} \& \href{fig:layouterr}{Figure~\ref{fig:layouterr}}). Several recent systems—including multi-stage pipelines proposed for the \texttt{2024 George B. Moody PhysioNet Challenge}—demonstrate that learning-based digitization components can work well \cite{a_reyna_digitization_2024, ecgimagekit2024, krones_combining_2024, a_verlyck_wavie_2024, yoon_segmentation-based_2024}; however, these efforts typically benchmark digitization sub-tasks in isolation and do not report a complete CPU-constrained, end-to-end pathway from scanned (and potentially deteriorated) reports through calibrated waveforms to downstream diagnostic modeling. Moreover, challenge reporting is not consistently stratified by \textbf{severely deteriorated} image conditions (faded ink, low contrast, wrinkles, and scanning noise), leaving open how robust ``best'' methods are under the failure modes most common in legacy archives.

To test whether digitization preserved diagnostically meaningful morphology, we trained deep-learning and kernel-based classifiers on reconstructed signals from \texttt{PTB-XL} and \texttt{ECG Matrix}. Best performance was: \texttt{PTB-XL} \textit{MI vs.~Normal}---\texttt{Arsenal} \cite{Middlehurst2021} (accuracy \textbf{0.9583}, \textbf{6{,}314.01\,ms} CPU latency); \texttt{ECG Matrix} \textit{Pre-procedural MI} vs.~\textit{Post procedural MI}---\texttt{InceptionTime} \cite{Fawaz2020InceptionTime} (\textbf{0.9053}, \textbf{2.00\,ms}); and \texttt{ECG Matrix} \textit{OMI vs.~non-OMI}---\texttt{Rocket} (accuracy \textbf{0.8889}).

For interpretability, we ran a lightweight \texttt{MLP}+\texttt{SHAP} attribution analysis to localize influential \textbf{leads} and \textbf{temporal regions} (\href{fig:mlp_workflow}{Figure~\ref{fig:mlp_workflow}}). \texttt{PTB-XL} \textit{MI vs.~Normal} attributions concentrate around the QRS complex and early ST segment, peaking in clinically informative lead groups (anterior \texttt{V1--V4}, lateral \texttt{I}/\texttt{aVL}/\texttt{V5--V6}, inferior \texttt{II}/\texttt{III}/\texttt{aVF}; \href{fig:mlp_shap}{Figure~\ref{fig:mlp_shap}A}). In \texttt{ECG Matrix}, \textit{Pre-} vs.~\textit{Post-surgery MI} is more distributed from QRS through ST--T, whereas \textit{OMI vs.~non-OMI} emphasizes repolarization-sensitive regions (ST segment and T wave) with coherent patterns across contiguous leads (\href{fig:mlp_shap}{Figure~\ref{fig:mlp_shap}B,C}). Overall, these physiologically plausible attributions suggest reconstruction preserves clinically meaningful morphology and provide an audit bridge from model decisions to actionable ECG intervals and territories.

Our experiments also highlight cross-source transfer where although the digitization models were trained only on synthetic \texttt{PTB-XL} sheet renderings, they generalized to real hospital-style \texttt{ECG Matrix} scans acquired through a different imaging pipeline, producing signals at a fidelity sufficient to train downstream models (see \href{tab:performance_comparison}{Table~\ref{tab:performance_comparison}}). End-to-end inference required approximately \textbf{25--30\,s per ECG} on CPU (see \href{fig:pipeline_detailed}{Figure~\ref{fig:pipeline_detailed}}), dominated by high-resolution patched instance segmentation and fusion/recomposition, while the MI classifiers operated in the millisecond-to-second range depending on model family.

Most quantitative digitization validation is performed on synthetic paper ECG images rendered from \texttt{PTB-XL} waveforms (using ECG sheet generators), because \texttt{PTB-XL} provides paired image--signal ground truth for direct measurement of reconstruction error\cite{ptb_xl}. Although we also apply the full pipeline to \texttt{ECG-Matrix}, the dataset provides report images and clinical labels but no paired digital waveforms; consequently, for real-world scans we evaluate digitization indirectly through downstream diagnostic performance and interpretability rather than signal-level reconstruction error. In addition, the \textit{pre-procedural vs.~post procedural} comparison is an exploratory proxy based on paired serial ECGs and may reflect multiple peri-procedural factors (timing, medications, evolving ischemia, and procedural effects) rather than a standardized endpoint, and the smaller sample size in \texttt{ECG Matrix} may limit calibration and partially explain modest performance differences (see \href{tab:performance_comparison}{Table~\ref{tab:performance_comparison}}).

Crucially, the same digitization--calibration--layout--classification pipeline can be extended to additional ECG pathologies ( arrhythmias, conduction disease, hypertrophy, electrolyte abnormalities) as datasets with paired waveforms and richer labels become available through routine clinical collection. In this sense, our contribution is not only improved digitization robustness but an explicit \emph{end-to-end} connection between reconstruction quality and diagnostic utility under \textbf{compute constraints}. Unlike prior work that validates individual components or assumes GPU/server-class resources, we provide and benchmark a fully offline, CPU-only pathway (see \href{fig:pipeline_detailed}{Figure~\ref{fig:pipeline_detailed}}), aligning with edge deployment needs in remote and resource-limited settings\cite{howard_mobilenets_2017, banbury_tinyml_2021}.

\section{CONCLUSION}

This work presents an end-to-end paper-ECG digitization, classification, and analysis framework that jointly addresses waveform extraction, reference-pulse calibration, layout recovery, downstream MI-related classification, and interpretability within a single pipeline (see \href{fig:pipeline_detailed}{Figure~\ref{fig:pipeline_detailed}}). The proposed workflow supports scalable paper-ECG digitization with exploratory MI-related decision support while localizing which waveform components and cardiac-cycle phases (P wave, QRS complex, ST segment, and T wave) most strongly contribute to a given prediction.

A defining feature of the proposed system is its suitability for low-resource, offline deployment, where both ECG digitization and downstream inference can be performed on CPU-only hardware. This reduces reliance on GPUs and cloud infrastructure while aligning with the practical constraints of low-power TinyML and edge-computing environments \cite{banbury_tinyml_2021}. In remote or under-resourced clinical settings—such as community health centers and mobile outreach programs, where ECGs are often printed, photographed, and reviewed intermittently—the pipeline can convert paper ECGs into calibrated digital time series at the point of care and automatically flag high-risk patterns consistent with MI or ACS, enabling faster clinical escalation and referral.

More broadly, bridging the digitization-to-diagnosis gap allows longitudinal paper ECG archives to be transformed into analyzable signals for population-scale screening, retrospective risk modeling, and clinical research. By facilitating earlier identification of time-sensitive cardiac pathology, the system may help reduce missed diagnoses and delays in treatment. Ultimately, by transforming legacy paper ECGs into actionable digital biomarkers through an end-to-end, resource-efficient workflow, the proposed approach expands access to automated lightweight AI-enabled cardiovascular screening and helps bring advanced diagnostic capabilities to settings where theyin settings where digital ECG export, connectivity, or high-end compute are needed the most.



\section*{Data and Code availability}
The \texttt{PTB-XL} dataset\cite{ptb_xl} (v1.0.3) is publicly available on PhysioNet at \url{https://physionet.org/content/ptb-xl/1.0.3/}. Additional study data, including \texttt{ECG-Matrix}, are available from the corresponding author upon reasonable request and subject to institutional approvals and data-sharing agreements. Code for the end-to-end paper-ECG workflow (pre-processing, digitization, post-processing, classification, and analyses) is available at \url{https://github.com/SCAI-Lab/ECGLight}, including a web dashboard, pretrained models/weights, and stage-specific configuration files.

\section*{Acknowledgments}
This work was funded by the \textit{Schweizer Paraplegiker Stiftung} (SPS) and the \textit{ETH Foundation} through grant \textbf{2021-HS-348} (\textit{Digital Transformation in Personalized Healthcare for Spinal Cord Injury (SCI) individuals}). The authors gratefully acknowledge this support, which enabled the development and evaluation of the proposed offline ECG digitization and time-series classification pipeline. We thank the clinical collaborators at the University of Campania ``Luigi Vanvitelli'' for their role in collecting and curating the \texttt{ECG-Matrix} dataset and for their continued scientific and clinical support throughout this study. The funders had no role in study design; data collection, analysis, or interpretation; manuscript preparation; or the decision to submit the work for publication.

%
\IEEEpeerreviewmaketitle

\bibliographystyle{IEEEtran}
\bibliography{bibtex/bib/IEEEabrv,bibtex/bib/IEEEexample}


%


\newpage

\renewcommand{\thefigure}{S\arabic{figure}}
\setcounter{figure}{0}
\renewcommand{\thetable}{S\arabic{table}}
\setcounter{table}{0}

\section*{Supplementary Information}

\subsection*{Synthetic ECG Image Generation and Dataset Preparation}

\begin{figure*}[!t]
\centering
\includegraphics[width=0.75\textwidth]{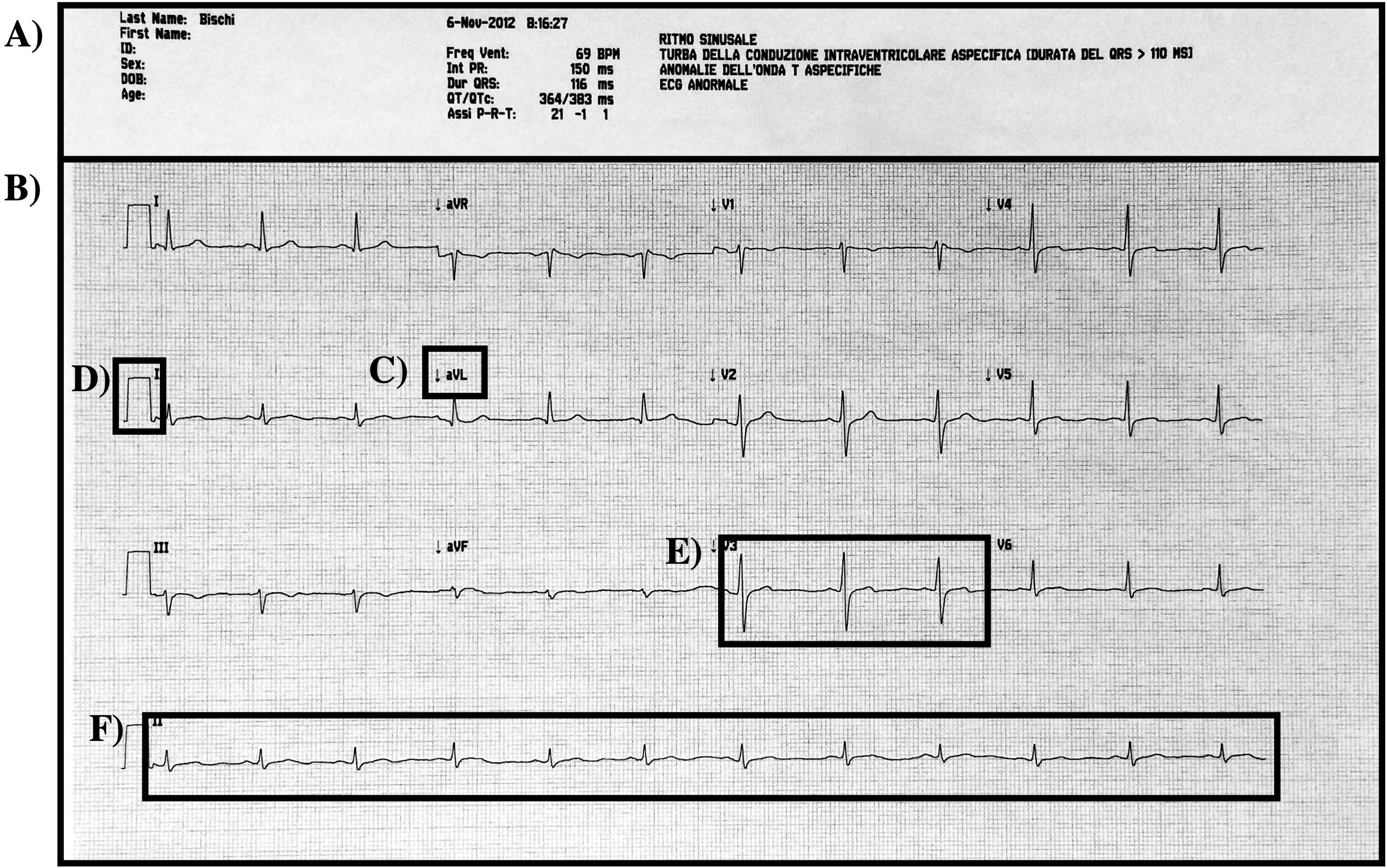}
\vspace{-4pt}
\caption{\textbf{Typical synthetic 12-lead paper ECG sheet derived from \texttt{PTB-XL}.} The rendered report replicates the standard clinical printout structure used throughout this study: \textbf{(A)} header/metadata region, \textbf{(B)} millimeter grid supporting visual time--amplitude reading, \textbf{(C)} printed lead identifiers used for automated lead-name detection and layout inference, \textbf{(D)} calibration/reference pulse used to recover physical scaling (mV and ms), \textbf{(E)} lead waveform traces that are segmented and subsequently vectorized, and \textbf{(F)} long rhythm strip providing extended temporal context. These components define the cues exploited by the digitization pipeline for layout-aware reconstruction and calibration.}
\label{fig:ecg_ex}
\end{figure*}

To rigorously evaluate the performance of the proposed end-to-end ECG digitization pipeline—including segmentation, lead-name and layout identification, reference pulse detection, and amplitude scaling—an extensive synthetic image dataset was generated from the \texttt{PTB-XL} waveform database~\cite{ptb_xl}. The \texttt{PTB-XL} dataset, one of the most comprehensive publicly available ECG corpora, provides 12-lead recordings stored in \texttt{WFDB} (WaveForm DataBase) format, sampled at 500~Hz and each lasting $10~\mathrm{s}$. Each ECG signal is accompanied by diagnostic metadata and hierarchical superclass annotations (\texttt{NORM}, \texttt{MI}, \texttt{STTC}, \texttt{CD}, \texttt{HYP}, and \texttt{OTHER}), ensuring the inclusion of diverse cardiac pathologies and morphological variations essential for model generalization.

The first stage of dataset creation involved converting these WFDB signals into high-resolution image representations that replicate the visual characteristics of clinical paper ECG printouts. This conversion was carried out using the \texttt{ECG-Image-Kit} rendering library, which automatically plots all 12 leads in standardized configurations—such as $4\times3$, $3\times4$, $6\times2$, and $12\times1$ layouts—while embedding essential structural elements including patient headers, calibration pulses, lead labels, and millivolt–millisecond grid lines. Each rendered image followed the standard clinical format, displaying the individual lead signals with consistent amplitude and time scaling across the page. A representative example of such an image is shown in \hyperref[fig:ecg_ex]{Figure~\ref*{fig:ecg_ex}}, which illustrates the main structural components of a 12-lead ECG: the patient and recording header (A), the calibration and scaling grid (B), the lead identifiers (C), the reference pulse used for amplitude calibration (D), the waveform segments corresponding to each lead (E), and the rhythm strip commonly used for temporal continuity analysis (F).

\begin{figure*}[t]
  \centering
  \begin{minipage}[t]{0.32\textwidth}%
    \centering
    \includegraphics[width=\linewidth]{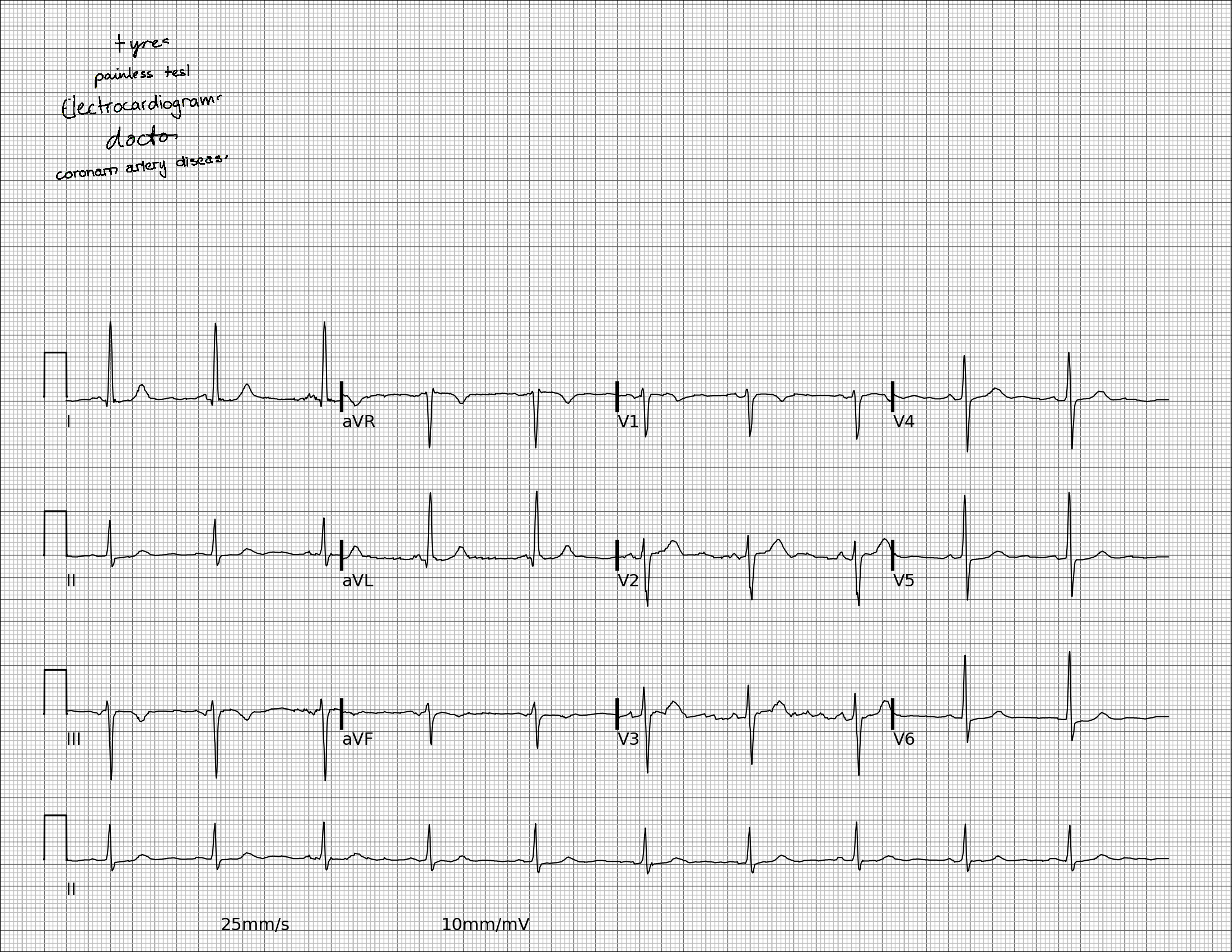}\\[-2pt]
    {\small (A)}
  \end{minipage}%
  \hfill
  \begin{minipage}[t]{0.32\textwidth}%
    \centering
    \includegraphics[width=\linewidth]{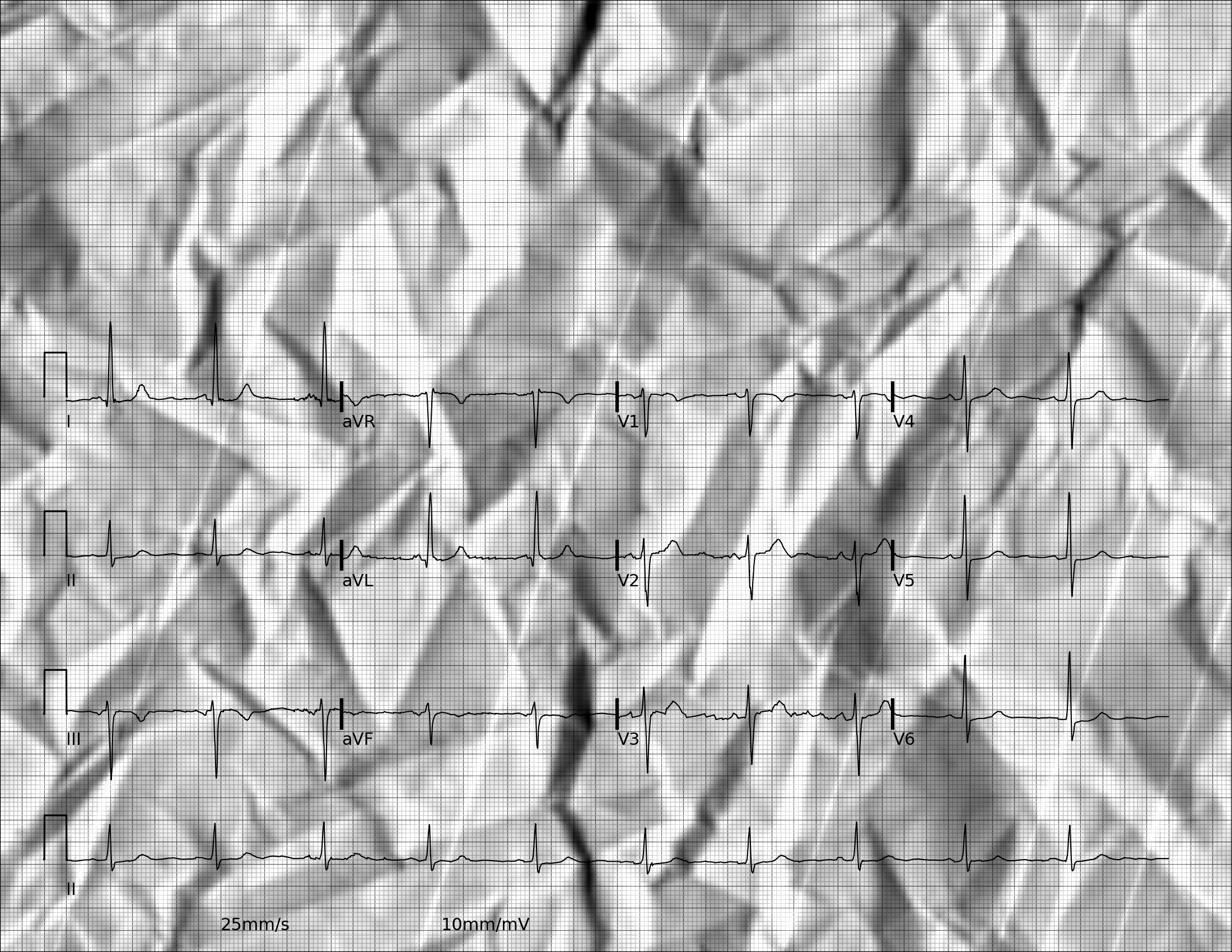}\\[-2pt]
    {\small (B)}
  \end{minipage}%
  \hfill
  \begin{minipage}[t]{0.32\textwidth}%
    \centering
    \includegraphics[width=\linewidth]{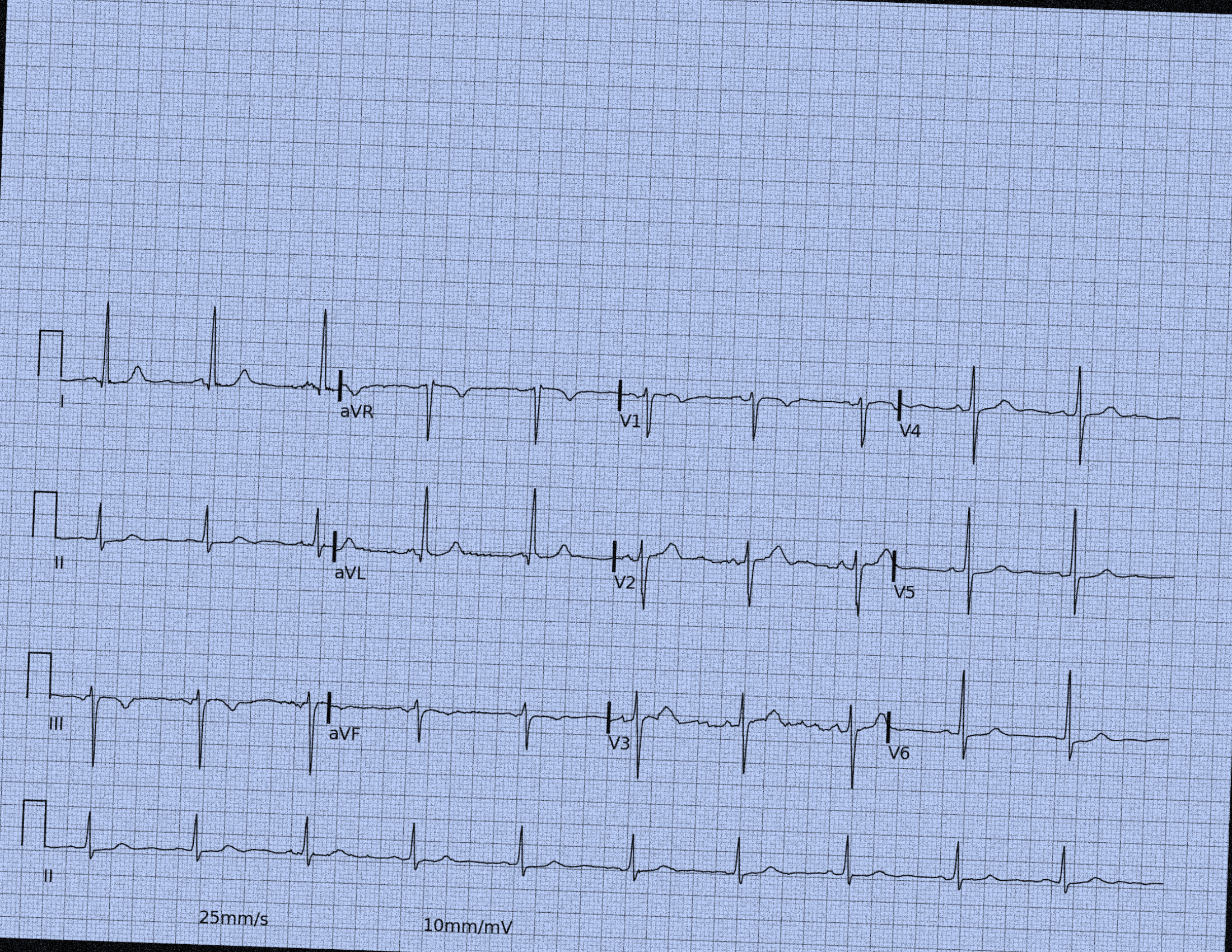}\\[-2pt]
    {\small (C)}
  \end{minipage}
  \vspace{-4pt}
  \caption{\textbf{Synthetic paper-ECG augmentations used to emulate real scanning and archival artifacts.} Examples of augmentations applied to the same rendered ECG sheet using \texttt{ECG-Image-Kit}: \textbf{(A)} handwritten annotations/overprints that can occlude waveforms and text labels, \textbf{(B)} paper creases, folds, and wrinkle-like shading that introduce non-uniform background texture, and \textbf{(C)} combined photometric and geometric distortions (temperature/contrast shifts, additive noise, and rotation) that mimic camera capture, compression, and imperfect alignment. These degradations are applied during training to improve robustness of segmentation, detection, and calibration under heterogeneous image quality.}
  \label{fig:augs}
\end{figure*}

To emulate real-world variations in paper-based ECGs and scanner artifacts, each rendered image was further augmented through a series of controlled degradations. The augmentation pipeline introduced illumination variation, Gaussian and impulse noise, random geometric distortions, motion blur, rotation, handwritten annotations, paper creases, and faded grid lines. These transformations were implemented using the \texttt{ECG-Image-Kit} augmentation module, designed specifically for ECG domain data to ensure physiologically plausible distortions. \hyperref[fig:augs]{Figure~\ref*{fig:augs}} depicts typical examples of these synthetic augmentations, including the introduction of handwritten notes and smudges (A), simulated folds and wrinkles resembling scanned paper artifacts (B), and color and contrast shifts with additive thermal and rotational noise (C). Collectively, these augmentations were instrumental in creating a robust training and validation corpus that captured the variability present in real-world ECG scans obtained from clinical archives and mobile imaging devices.

The synthetically generated dataset served as the foundation for all model development and evaluation tasks in this study. The images were first utilized for the training and validation of the \texttt{YOLOv11x-seg} model for patch-based segmentation, where each ECG sheet was divided into overlapping tiles to extract fine-grained waveform boundaries. The same dataset was subsequently employed for lead-name detection using a \texttt{YOLOv11x} object detection model trained on synthetically augmented lead label images, followed by automated layout identification based on detected lead positions. Finally, the images containing calibration markers were used for reference pulse detection and amplitude scaling, enabling precise conversion from pixel units to physical ECG measures in millivolts and milliseconds.

Through this multi-stage synthesis and augmentation process, the \texttt{PTB-XL}-derived image dataset provided a consistent and controllable framework for assessing the digitization pipeline under diverse degradation conditions. By combining clinical signal fidelity from the original WFDB recordings with realistic image-level perturbations, the dataset enabled reproducible evaluation of segmentation, detection, and scaling algorithms, ultimately supporting the quantitative analyses summarized in \href{tab:results}{Table~\ref{tab:results}}.

\subsection*{Segmentation Methodology and Optimization}

\begin{figure*}[!t]
  \centering
  \begin{minipage}[t]{0.31\textwidth}%
    \centering
    \includegraphics[width=\linewidth]{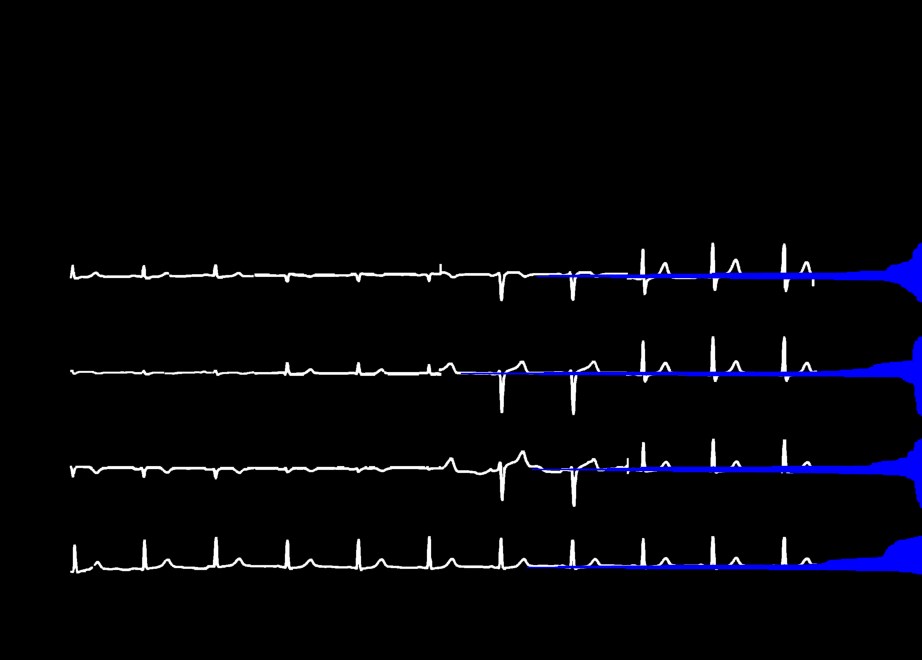}\\[-2pt]
    {\small (A)}
  \end{minipage}%
  \hfill
  \begin{minipage}[t]{0.31\textwidth}%
    \centering
    \includegraphics[width=\linewidth]{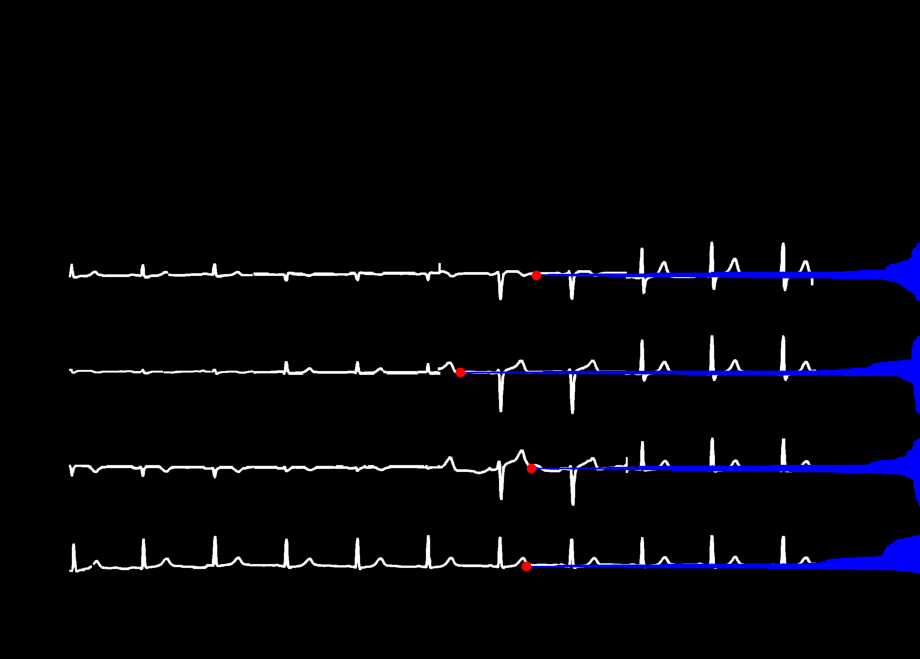}\\[-2pt]
    {\small (B)}
  \end{minipage}%
  \hfill
  \begin{minipage}[t]{0.31\textwidth}
    \centering
    \includegraphics[width=\linewidth]{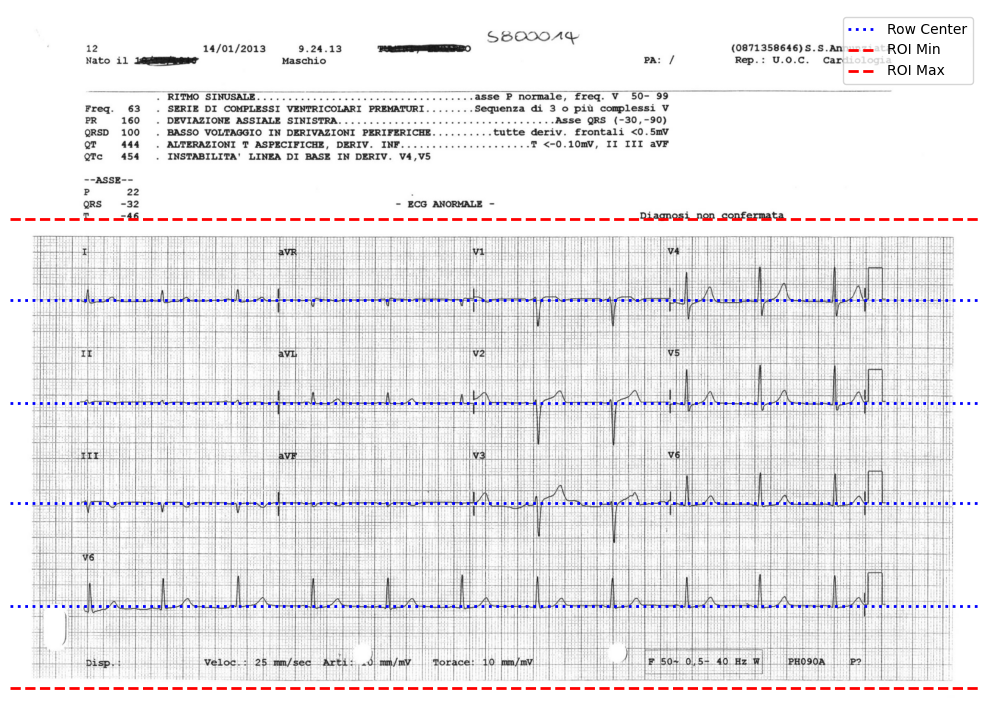}\\[-2pt]
    {\small (C)}
  \end{minipage}
  \caption{\textbf{Row-center estimation and region-of-interest (ROI) extraction for layout-aware processing.} \textbf{(A)} A 1D horizontal projection (sum of non-zero pixels per row) is computed from the segmentation mask to highlight lead rows. \textbf{(B)} After smoothing, peaks in the projection correspond to candidate row centers and provide an adaptive estimate of inter-row spacing. \textbf{(C)} Detected row centers (blue) define the vertical layout; the ROI bounds (red) restrict subsequent lead-name association and vectorization to signal-bearing regions while excluding headers and margins, improving robustness across formats and resolutions.}
  \label{fig:peak}
\end{figure*}

\begin{figure*}[!t]
  \centering
  \begin{minipage}[t]{0.32\textwidth}%
    \centering
    \includegraphics[width=\linewidth]{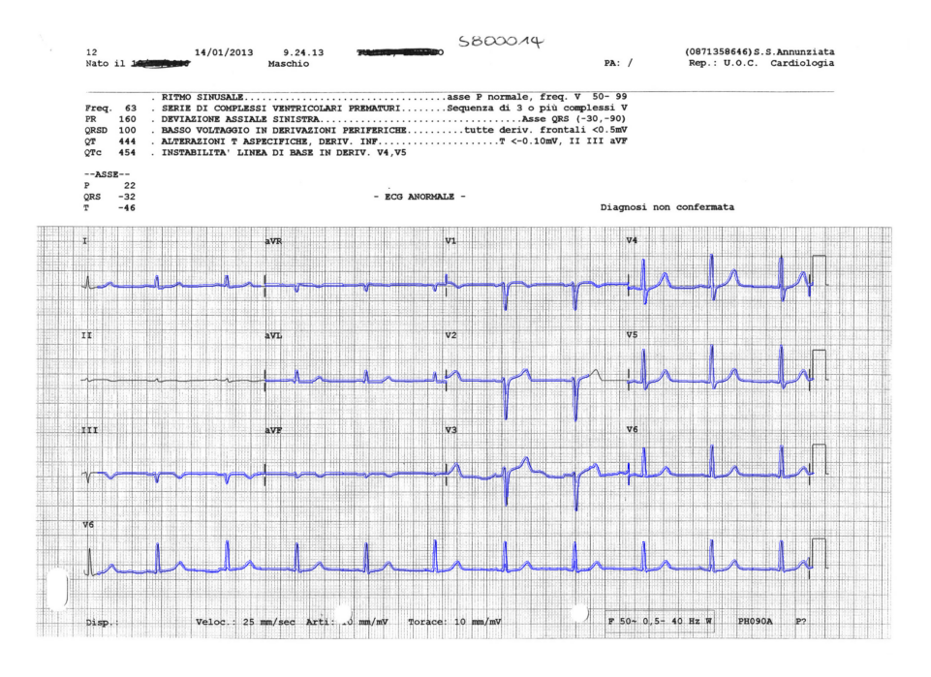}\\
    \includegraphics[width=\linewidth]{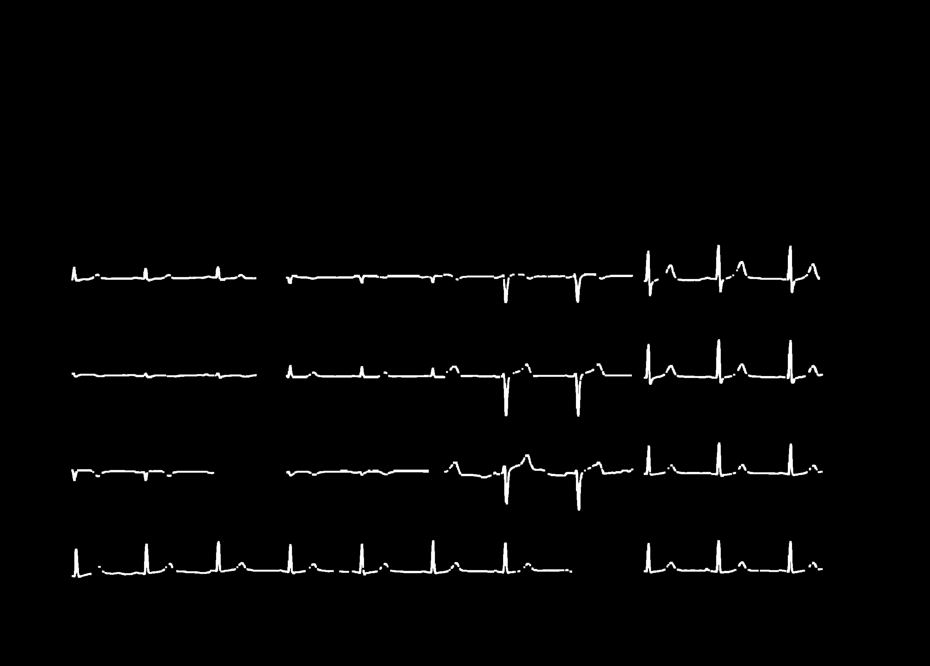}\\[-2pt]
    \textbf{(A)}
  \end{minipage}%
  \hfill
  \begin{minipage}[t]{0.32\textwidth}%
    \centering
    \includegraphics[width=\linewidth]{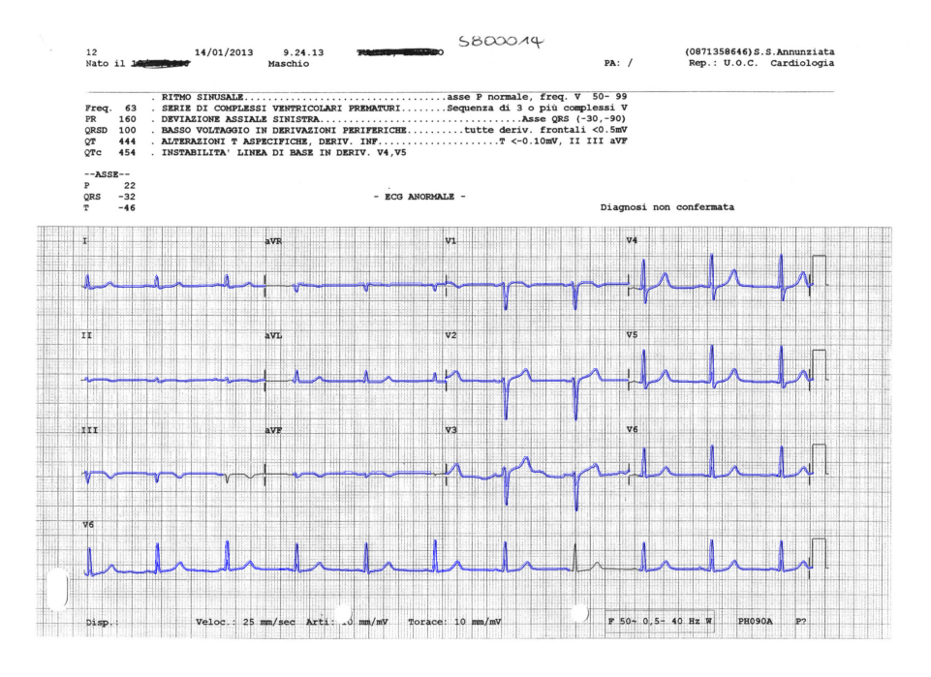}\\
    \includegraphics[width=\linewidth]{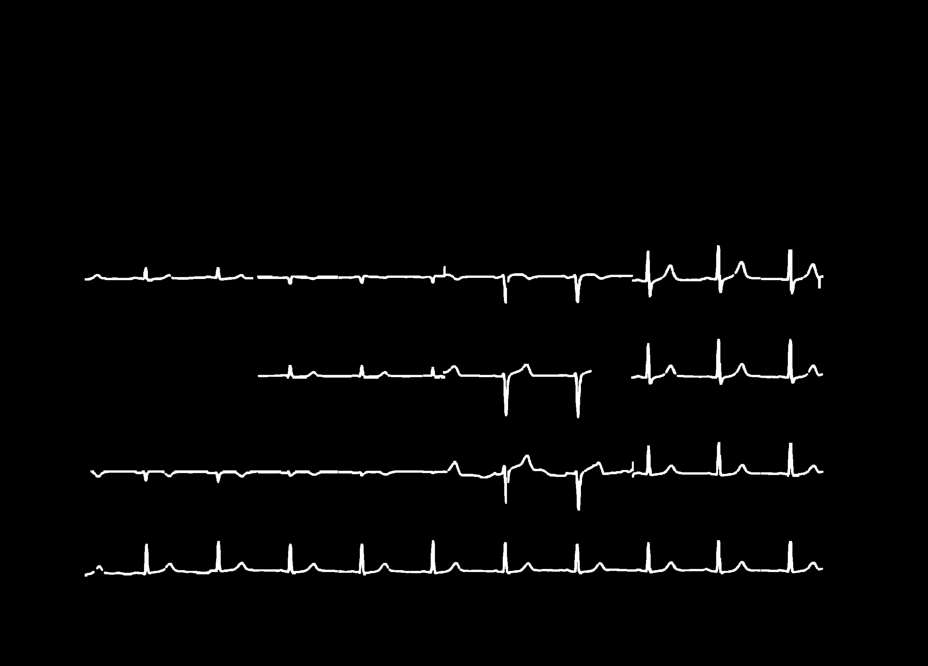}\\[-2pt]
    \textbf{(B)}
  \end{minipage}%
  \hfill
  \begin{minipage}[t]{0.32\textwidth}
    \centering
    \includegraphics[width=\linewidth]{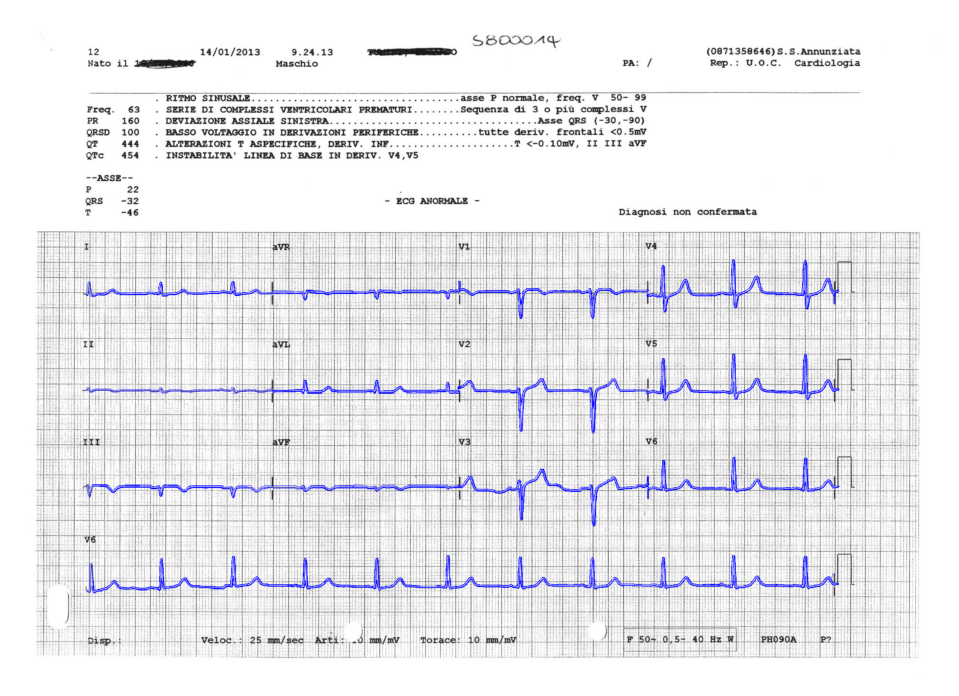}\\
    \includegraphics[width=\linewidth]{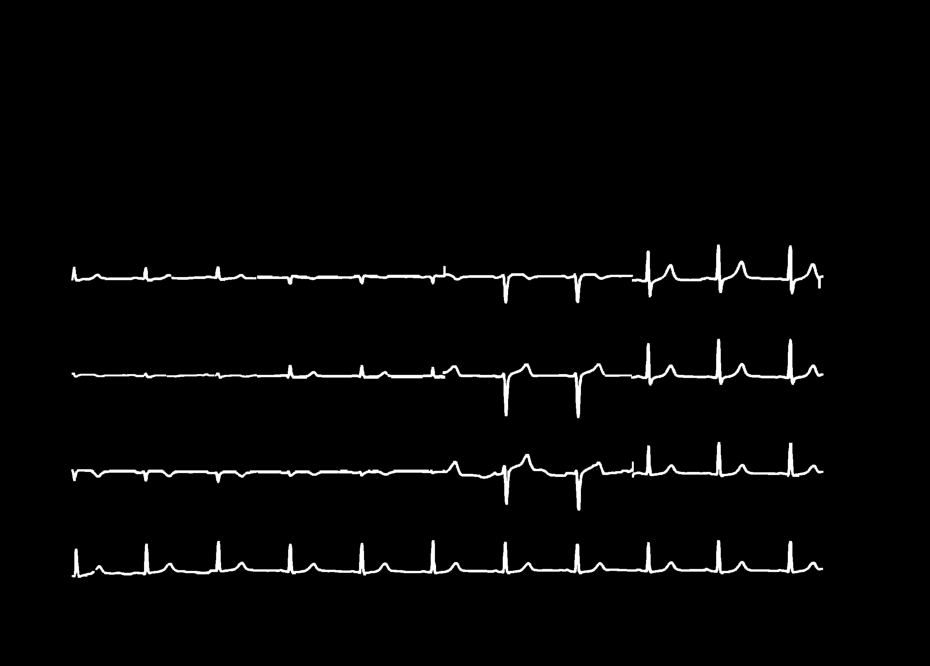}\\[-2pt]
    \textbf{(C)}
  \end{minipage}

  \caption{\textbf{Effect of patch scale and multi-scale fusion on waveform segmentation.} Predicted masks (top) and binarized post-processed masks (bottom) for different patch-size regimes: \textbf{(A)} smaller patches (20\%) can miss long contiguous structures (e.g., incomplete detection of lead~II) due to reduced context; \textbf{(B)} moderately larger patches (25\%) improve continuity but may still show localized gaps under severe degradations; \textbf{(C)} union/fusion of multi-scale predictions increases coverage and reduces seam artifacts, providing a more reliable global mask for downstream lead mapping and vectorization.}
  \label{fig:patchsz}
\end{figure*}

Accurate segmentation of ECG waveforms from synthetically generated paper-like images is a critical step in the proposed digitization pipeline, serving as the foundation for subsequent lead identification, scaling, and signal reconstruction. Given the variability introduced by synthetic degradations such as noise, rotation, and illumination imbalance, a hybrid segmentation strategy combining geometric preprocessing and deep learning–based mask prediction was adopted.

\begin{figure*}[!t]
  \centering
  \begin{minipage}[t]{0.48\textwidth}%
    \centering
    \includegraphics[width=\linewidth]{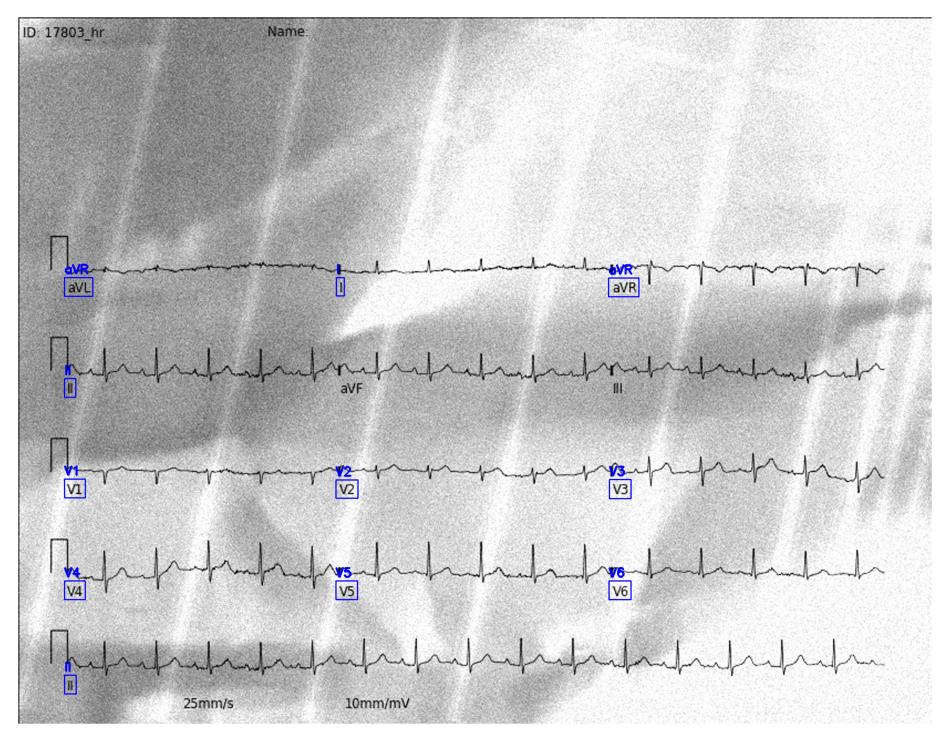}\\[-2pt]
    {\small (A)}
  \end{minipage}%
  \hfill
  \begin{minipage}[t]{0.48\textwidth}
    \centering
    \includegraphics[width=\linewidth]{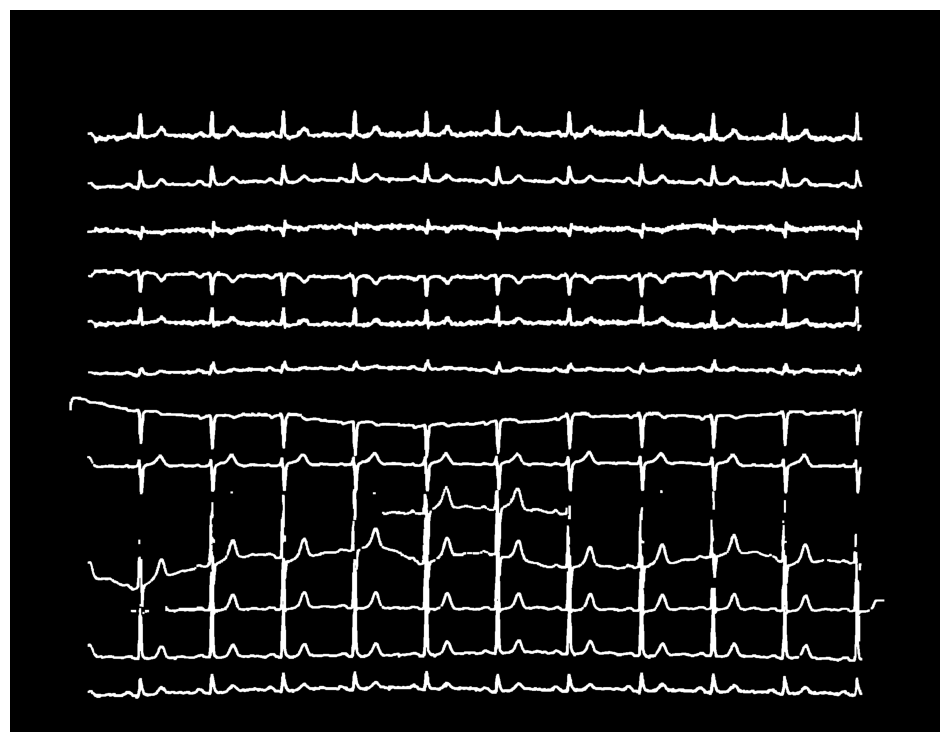}\\[-2pt]
    {\small (B)}
  \end{minipage}
  \caption{\textbf{Representative layout-inference failure cases.} \textbf{(A)} A 4$\times$3 \texttt{Cabrera} sheet in which a missed \texttt{aVF} lead-label detection disrupts the inferred limb-lead ordering and therefore the layout classification. \textbf{(B)} A 12$\times$1 long-strip format in which incomplete waveform segmentation reduces the quality of the horizontal projection used for row detection, leading to incorrect row/lead assignment. These examples motivate the use of confidence thresholds and quality-control filters for deployment.}
  \label{fig:layouterr}
\end{figure*}

The initial stage of segmentation involved the identification of row centers and regions of interest (ROIs) corresponding to each lead. As shown in \href{fig:peak}{Figure~\ref{fig:peak}}, this was achieved through horizontal summation of non-zero pixel values across the image to form a one-dimensional intensity profile (\textbf{A}). After applying a smoothing filter to suppress spurious fluctuations, local maxima in this profile were detected as potential lead row centers (\textbf{B}). These centers were used to delineate the bounding regions for each lead, producing the final ROI boundaries (red lines) and center positions (blue lines) shown in \textbf{(C)}. This geometric preprocessing ensured that each ECG waveform was properly localized despite variable image resolutions or scanning distortions.

\begin{table*}[!t]
\centering
\begin{tabular}{@{}lccc@{}}
\toprule
\textbf{Category / Characteristic} & \textbf{(A) Training Subset} & \textbf{(B) Patched Training Subset} & \textbf{(C) Evaluation Subset} \\
\midrule
\multicolumn{4}{@{}l}{\textbf{Dataset Overview}} \\
Total ECG Recordings              & 21,799 & 11,000 & 1,600 \\
Unique Patients                   & 18,869 & 10,115 & 1,574 \\
\midrule
\multicolumn{4}{@{}l}{\textbf{Diagnostic Superclass Distribution}} \\
NORM (Normal)                     & 9,514  & 4,780  & 698 \\
MI (Myocardial Infarction)        & 5,469  & 2,749  & 398 \\
STTC (ST/T Change)                & 5,235  & 2,678  & 387 \\
CD (Conduction Disturbance)       & 4,898  & 2,455  & 345 \\
HYP (Hypertrophy)                 & 2,649  & 1,315  & 184 \\
OTHER                             & 411    & 221    & 33 \\
\midrule
\multicolumn{4}{@{}l}{\textbf{Patch Augmentation Statistics (Subset B)}} \\
Total Patches Generated           & — & 323,000 & — \\
Mean $\pm$ SD Patches per Patient & — & $29.36 \pm 7.42$ & — \\
Median Patches per Patient        & — & 30.0 & — \\
Range (min--max)                  & — & 20--48 & — \\
\midrule
\multicolumn{4}{@{}l}{\textbf{Patch Count Distribution (Subset B)}} \\
20 Patches $\rightarrow$ Patients & — & 4,750 & — \\
30 Patches $\rightarrow$ Patients & — & 4,000 & — \\
48 Patches $\rightarrow$ Patients & — & 2,250 & — \\
\bottomrule
\end{tabular}
\caption{
\textbf{Dataset Composition \& Demographic Statistics Used Across The Model Development Pipeline.}
\textbf{(A)} Full filtered \texttt{PTB-XL} cohort comprising 21,799 12-lead ECG recordings from 18,869 unique patients, used for training/validation of the \texttt{YOLO}-based segmentation, lead-name detection, and reference-pulse detection components. We report the per-label sample counts for the major diagnostic superclasses (\texttt{NORM, MI, STTC, CD, HYP, OTHER}) across splits, highlighting the inherent class imbalance typical of clinical ECG datasets.
\textbf{(B)} Patched augmentation subset ($N = 11{,}000$ recordings) used for \texttt{YOLOv11x}-based segmentation training and validation, with a total of 323,000 cropped patches generated to increase layout and noise diversity. The patching scheme yielded a mean of $29.36 \pm 7.42$ patches per patient (median 30; range 20--48)
\textbf{(C)} Final evaluation subset ($N = 1{,}600$ ECGs from 1,574 patients) used to benchmark end-to-end digitization performance using Pearson correlation, SNR, and RMSE.
Across all subsets, this table provides patient-level demographic statistics (mean age $\pm$ SD and sex distribution), showing a near-balanced male/female proportion (male:female ratio $\approx 1.06$--$1.09$). All recordings were standardized 12-lead ECGs sampled at 500~Hz with consistent annotation structure and verified patient-level metadata.
}
\label{tab:dataset}
\end{table*}

Following ROI extraction, the localized patches were processed using the \texttt{YOLOv11x-seg} model trained in a patch-based configuration. Each full ECG image was divided into overlapping tiles of varying patch sizes (20\%–30\% of the full resolution), enabling the model to focus on local waveform morphology while maintaining context across adjacent leads. As illustrated in \href{fig:patchsz}{Figure~\ref{fig:patchsz}}, smaller patch sizes such as 20\% (\textbf{A}) occasionally failed to capture complete lead structures—most notably, the lead II waveform was missed in several instances—while slightly larger patch sizes such as 25\% (\textbf{B}) improved continuity but still exhibited minor detection gaps under severe degradation. To mitigate this, predictions from multiple patch scales were merged using pixel-wise logical aggregation, resulting in combined segmentation masks (\textbf{C}) that achieved near-complete waveform coverage even in the presence of local occlusions or gridline interference.

Quantitative evaluation of the segmentation models, detailed in \href{tab:results}{Table~\ref{tab:results}}, demonstrated that the patched \texttt{YOLOv11x} configuration achieved the highest overall accuracy. While the full-image model attained superior object-level precision (0.995) and recall (0.991) due to its access to broader spatial context, its mask-level overlap metrics—IoU and Dice—were considerably lower (0.221 and 0.353, respectively), reflecting poor boundary precision. In contrast, the patch-based \texttt{YOLOv11x-seg} model achieved a balanced performance with IoU = \textbf{0.647} and Dice = \textbf{0.782}, outperforming both full-image and \texttt{YOLOv12x} variants. This improvement arises from the ability of localized patches to better model fine-grained transitions between waveform and background regions, especially in noisy or deteriorated ECG scans.

By combining robust geometric localization with patch-based segmentation and multi-scale aggregation, the proposed approach effectively mitigates the common challenges of waveform overlap, partial occlusion, and paper texture interference. The resulting segmentation masks form the structural backbone for all downstream stages—lead-name detection, layout classification, and reference pulse–based amplitude scaling—ensuring that each waveform is accurately isolated and preserved for signal-level reconstruction.

\subsection*{Lead Name Detection and Layout Estimation}

Following segmentation of waveform regions, the next critical stage in the digitization pipeline involved \textbf{lead-name detection} and \textbf{layout estimation}, both of which define the spatial organization of the 12-lead ECG recording. Accurate identification of lead labels and their spatial arrangement is essential for correctly mapping waveform segments to their anatomical and functional counterparts, ensuring the reconstructed digital signal preserves clinical interpretability.

Lead-name detection was performed using a fine-tuned \texttt{YOLOv11x} model trained on localized text patches extracted from the \texttt{PTB-XL} dataset. The training data underwent extensive visual augmentation to simulate real-world scanning and printing variability, including rotation, fading, compression artifacts, and handwritten overlays. \href{fig:objdet}{Figure~\ref{fig:objdet}} illustrates examples of these augmentations: (\textbf{A}) depicts diverse augmentations applied to the aVR lead label, while (\textbf{B}) shows augmentation strategies used for reference pulse detection. These augmentations were crucial for improving robustness against heterogeneous visual conditions encountered in paper ECGs. The final model achieved near-perfect precision and recall (\textbf{0.997} and \textbf{0.991}, respectively), as reported in \href{tab:results}{Table~\ref{tab:results}}, demonstrating reliable detection of both text and structural calibration cues under varied degradations.

\begin{figure*}[!t]
  \centering
  \begin{minipage}[t]{0.485\textwidth}%
    \centering
    \includegraphics[width=\linewidth]{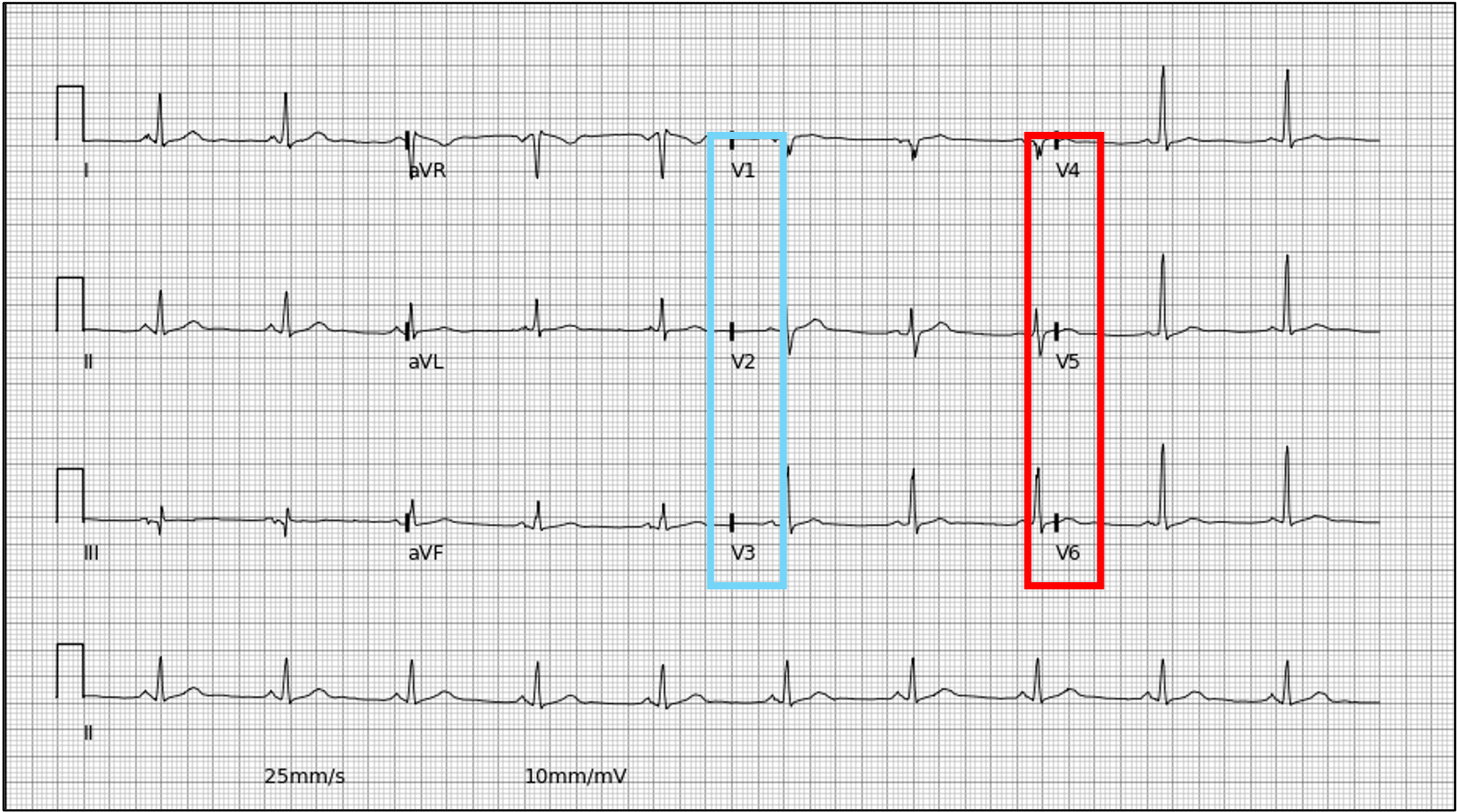}\\[-2pt]
    {\small (A)}
  \end{minipage}%
  \hfill
  \begin{minipage}[t]{0.485\textwidth}
    \centering
    \includegraphics[width=\linewidth]{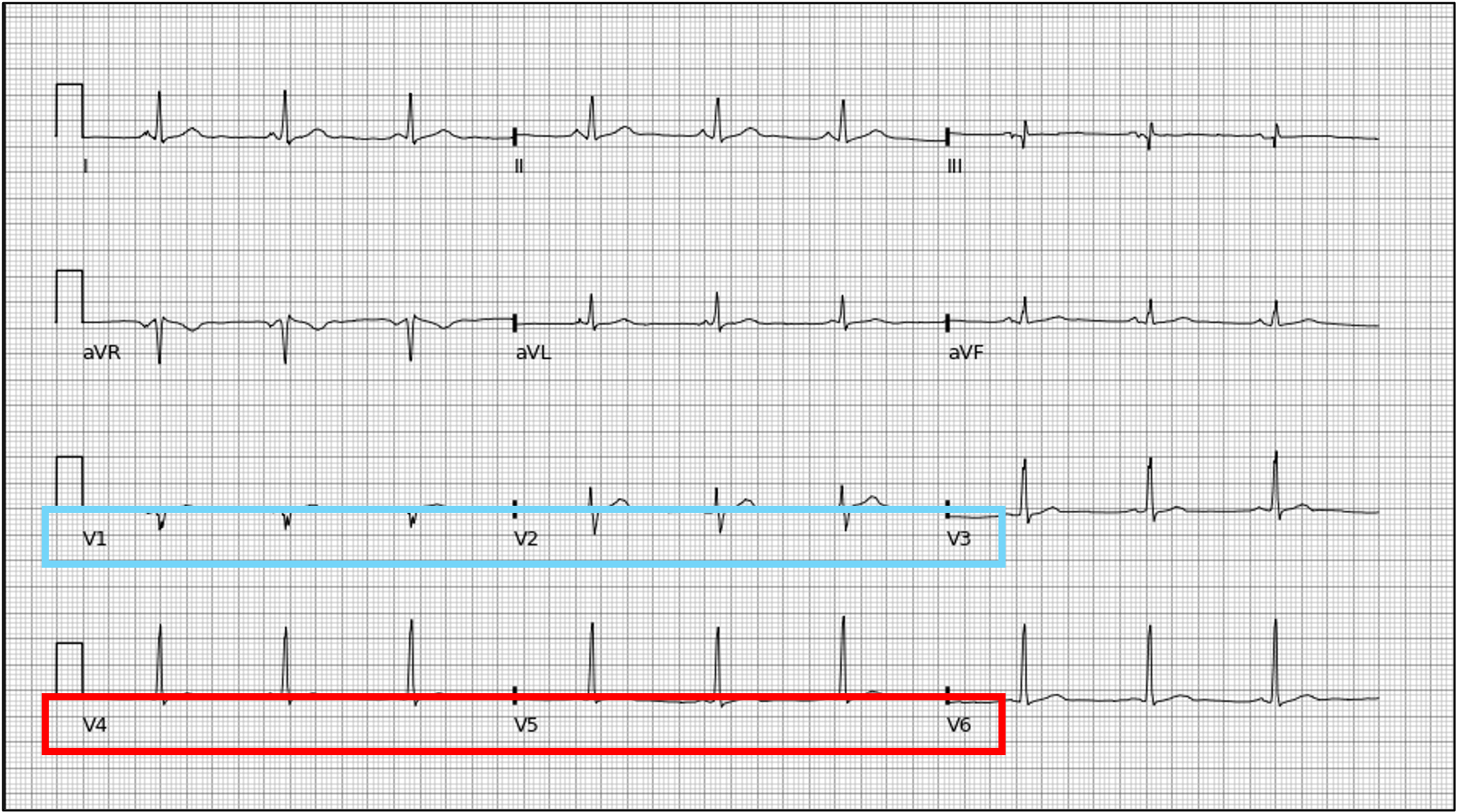}\\[-2pt]
    {\small (B)}
  \end{minipage}
  \caption{\textbf{Disambiguating 4-row layouts using precordial-lead grouping.} When four signal rows are detected, two common configurations are plausible: \textbf{(A)} a 3$\times$4 grid with an additional rhythm strip row, and \textbf{(B)} a 4$\times$3 grid without a rhythm strip. The relative alignment of precordial leads \texttt{V1--V3} (blue) and \texttt{V4--V6} (red) provides a robust cue: their row/column grouping differs systematically across these layouts, enabling rule-based resolution of the ambiguity after lead-name detection.}
  \label{fig:layout}
\end{figure*}

Once lead names were detected, a spatial association algorithm linked each label to its corresponding waveform segment using bounding-box proximity and alignment derived from the segmentation outputs. This ensured correct mapping of all 12 leads—including limb, augmented unipolar, and precordial leads—regardless of print or scan format differences. The same detection framework also identified reference pulses, which were later used for amplitude scaling and calibration in the final reconstruction phase.

The \textbf{layout estimation} stage subsequently determined the overall configuration of the ECG grid, identifying whether the recording followed a $3\times4$, $4\times3$, $6\times2$, or $12\times1$ arrangement. Layout inference relied on geometric ordering and alignment of the detected lead names. For instance, when four vertical groupings were detected, the model distinguished between a $3\times4$ layout (with a rhythm strip at the bottom) and a $4\times3$ layout (without a rhythm strip) by analyzing inter-row distances and rhythm lead positioning (see \href{fig:layout}{Figure~\ref{fig:layout}}). This classification step was essential for reconstructing ECGs in their canonical diagnostic order and ensuring temporal continuity across concatenated leads.

\begin{figure*}[!t]
  \centering
  \begin{minipage}{0.44\textwidth}
    \centering
    \includegraphics[width=\textwidth]{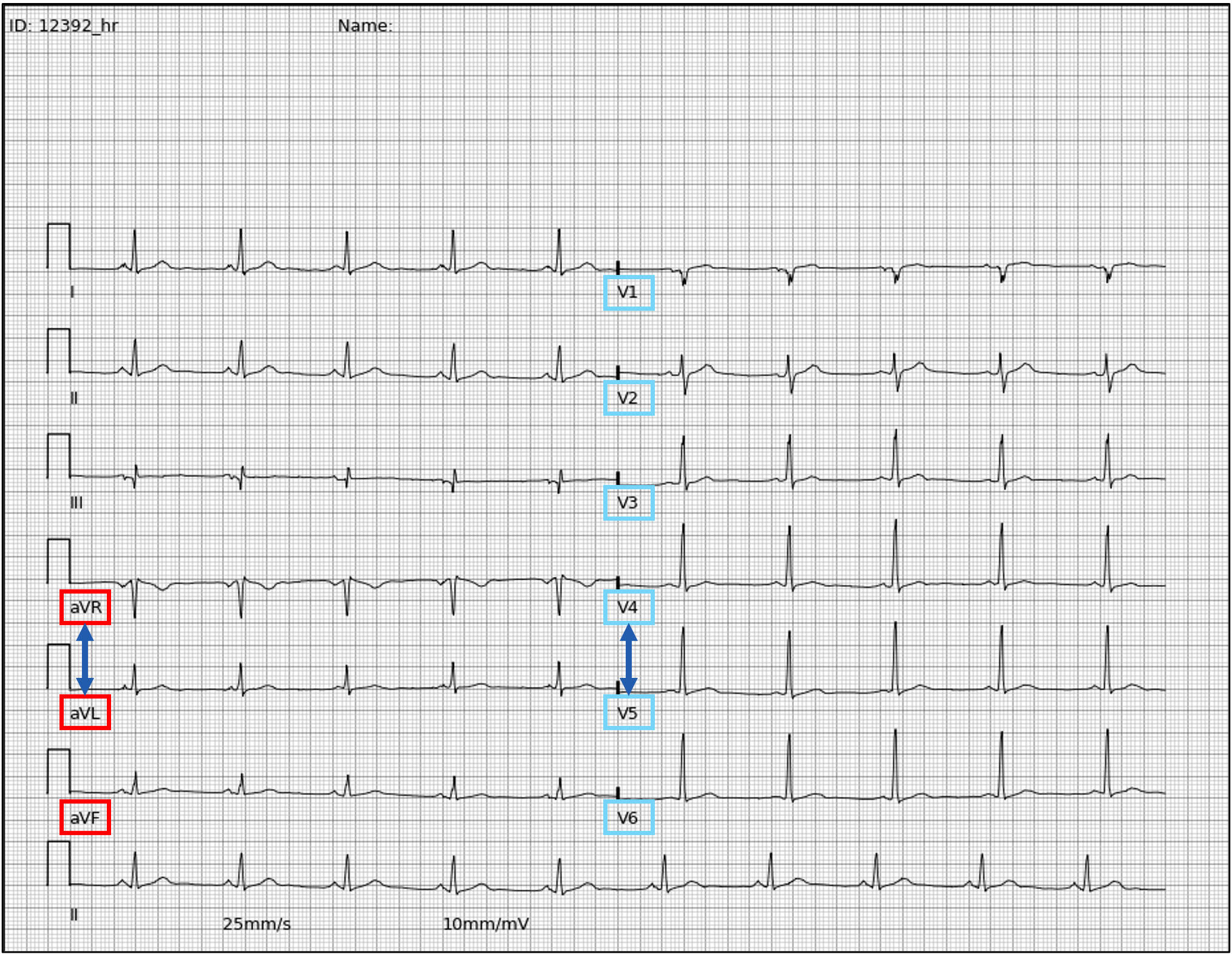}\\
    \textbf{(A)}
    \includegraphics[width=\textwidth]{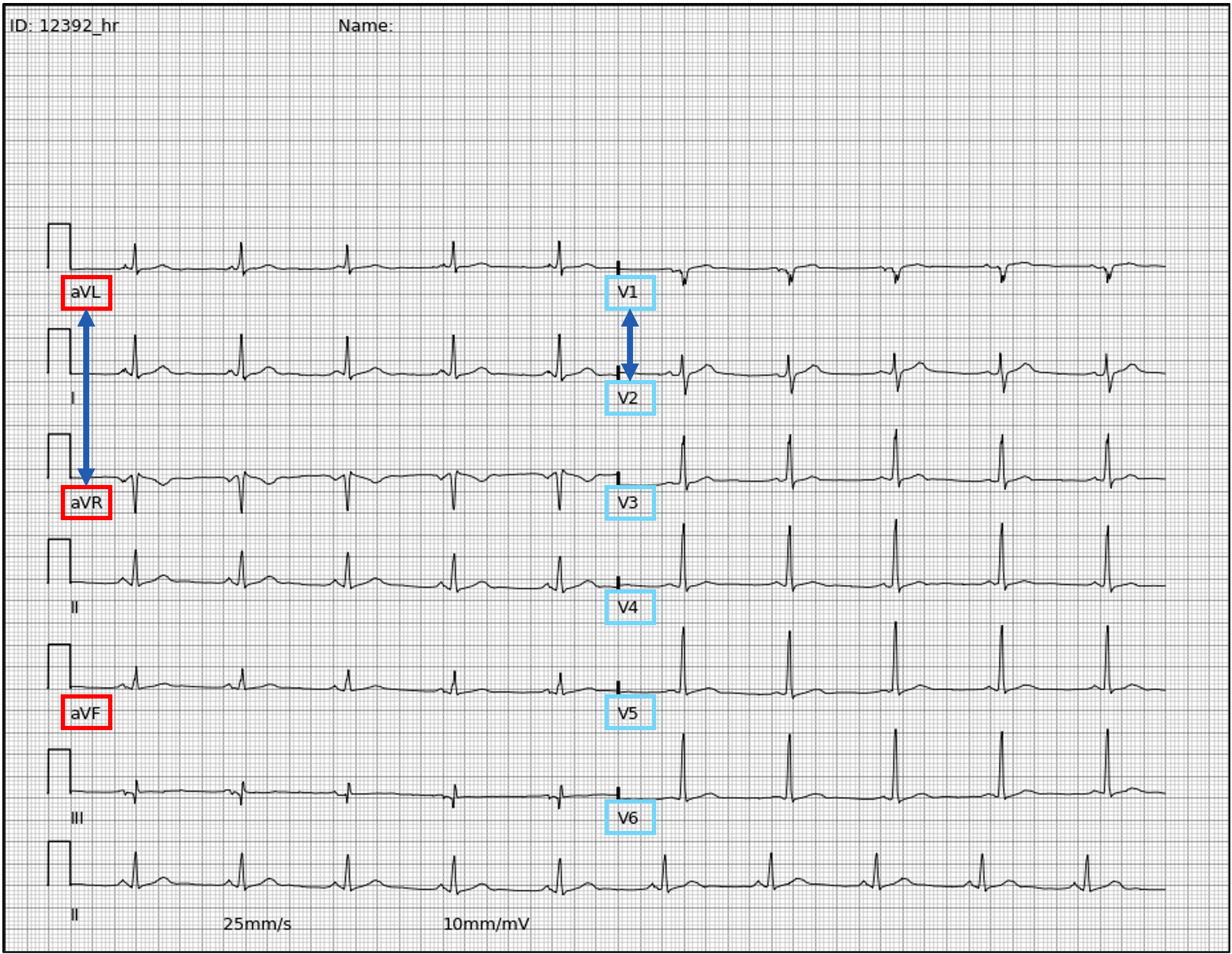}\\
    \textbf{(B)}
  \end{minipage}%
  \hfill
  \begin{minipage}{0.44\textwidth}
    \centering
    \includegraphics[width=\textwidth]{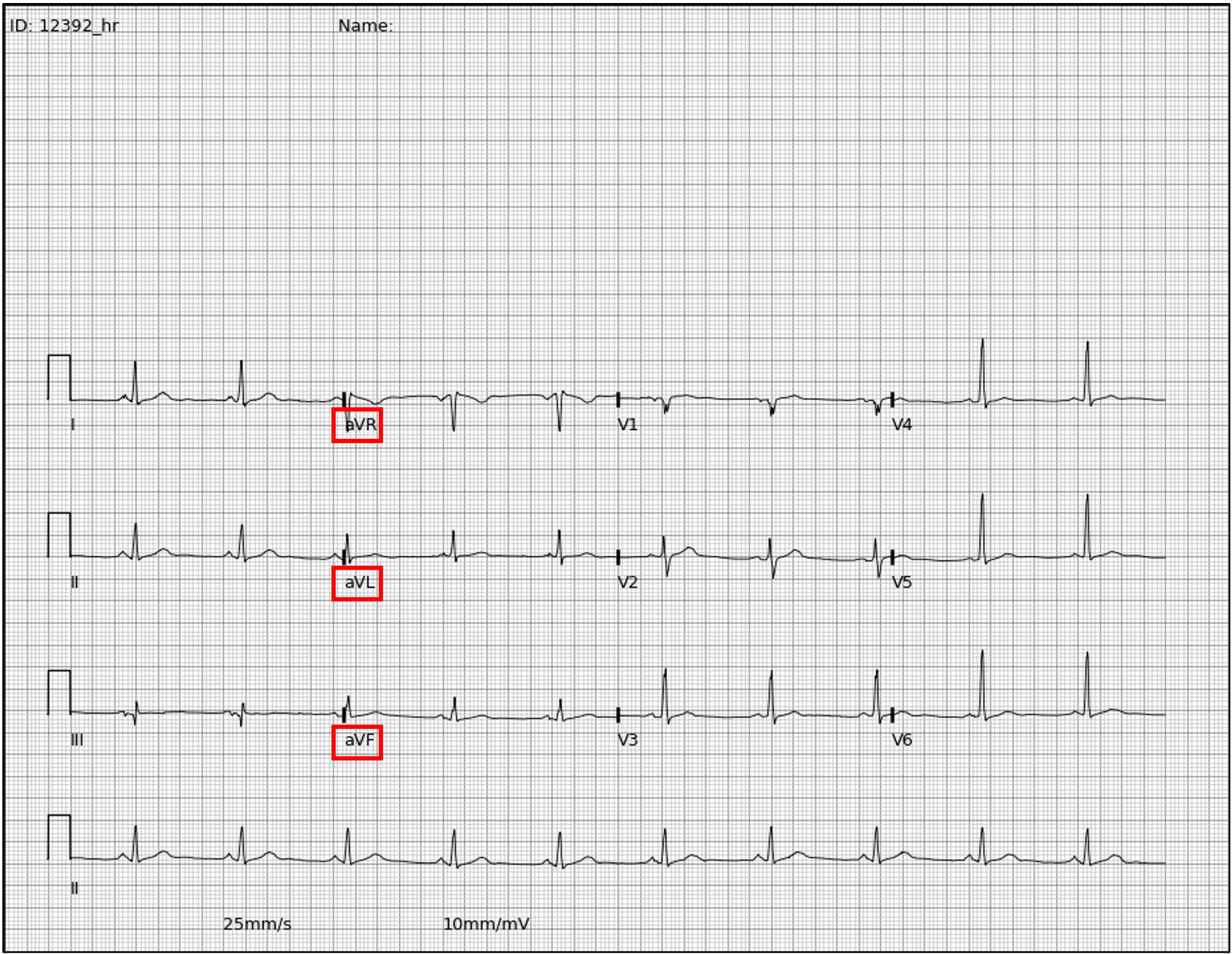}\\
    \textbf{(C)}
    \includegraphics[width=\textwidth]{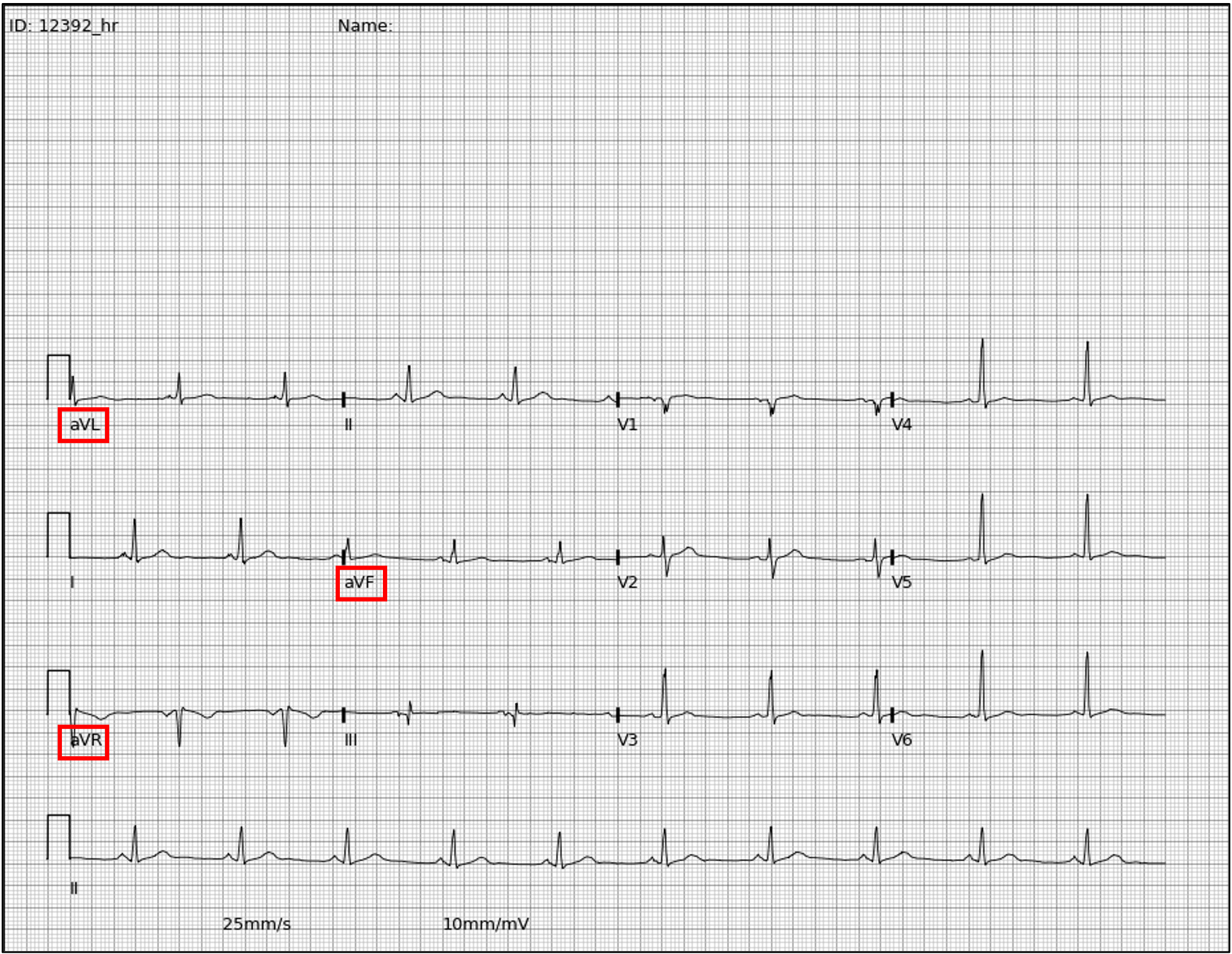}\\
    \textbf{(D)}
  \end{minipage}
  \caption{\textbf{Normal vs. \texttt{Cabrera} format cues from augmented unipolar-lead geometry.} Augmented unipolar limb leads (red; \texttt{aVL}, \texttt{aVR}, \texttt{aVF}) and precordial leads (blue; \texttt{V1--V6}) exhibit characteristic spacing/alignment patterns that enable automated \texttt{Cabrera} detection. \textbf{(A)} In a normal 6$\times$2 format, augmented unipolar leads follow uniform spacing comparable to precordial blocks. \textbf{(B)} In a \texttt{Cabrera} 6$\times$2 format, the limb-lead ordering introduces alternating/doubled spacing. \textbf{(C)} In a normal 3$\times$4 grid, augmented unipolar leads are aligned in their expected positions. \textbf{(D)} In a \texttt{Cabrera} 3$\times$4 grid, misalignment (notably of \texttt{aVF}) relative to \texttt{aVL}/\texttt{aVR} provides a discriminative geometric signature.}
  \label{fig:cabrera}
\end{figure*}

A particular challenge arose due to the coexistence of two presentation standards—the \textbf{Normal} and the \textbf{Cabrera} formats. As shown in \href{fig:cabrera}{Figure~\ref{fig:cabrera}}, these differ in the placement and spacing of augmented unipolar leads (aVR, aVL, aVF). In the normal format (\textbf{A}, \textbf{C}), augmented unipolar and precordial leads appear uniformly spaced, while in the Cabrera format (\textbf{B}, \textbf{D}), augmented unipolar leads exhibit double spacing or offset alignment. To address this, the system incorporated both spatial heuristics and lead-order validation, allowing it to distinguish between the two configurations with high reliability across both $3\times4$ and $6\times2$ layouts.

Despite strong performance—achieving over \textbf{99\%} accuracy for the $3\times4$ and $6\times2$ formats—some edge cases resulted in misclassifications. \href{fig:layouterr}{Figure~\ref{fig:layouterr}} illustrates two representative failure scenarios: (\textbf{A}) a $4\times3$ Cabrera-format ECG where the missing aVF label caused misalignment in row inference, and (\textbf{B}) a $12\times1$ layout where incomplete segmentation led to missing waveform associations. These failures underscore the dependence of layout estimation accuracy on precise segmentation and label extraction.

Overall, the integrated lead-name detection and layout estimation framework enabled fully automated spatial organization of ECG waveforms, bridging the visual-to-structural gap in the digitization process. By combining robust object detection with geometric reasoning, it effectively handled heterogeneous formats, printing artifacts, and noise, forming a critical foundation for the subsequent reference pulse–based scaling and waveform reconstruction stages.

\subsection*{Reference Pulse Detection and Scaling Estimation}

Following lead-name and layout identification, the next crucial stage of the pipeline involved the detection of the \textbf{reference calibration pulse} and the estimation of physical scaling parameters for amplitude (in millivolts) and time (in milliseconds). These calibration pulses are rectangular markers, typically corresponding to a $1~\mathrm{mV}$ vertical deflection and $200~\mathrm{ms}$ horizontal duration, and are printed on nearly all clinical ECG sheets. They serve as an absolute reference for signal quantification, allowing the conversion of extracted waveform pixel coordinates into physiologically meaningful units.

\begin{figure*}[!t]
  \centering
  \begin{minipage}{0.45\textwidth}
    \centering
    \includegraphics[width=\textwidth]{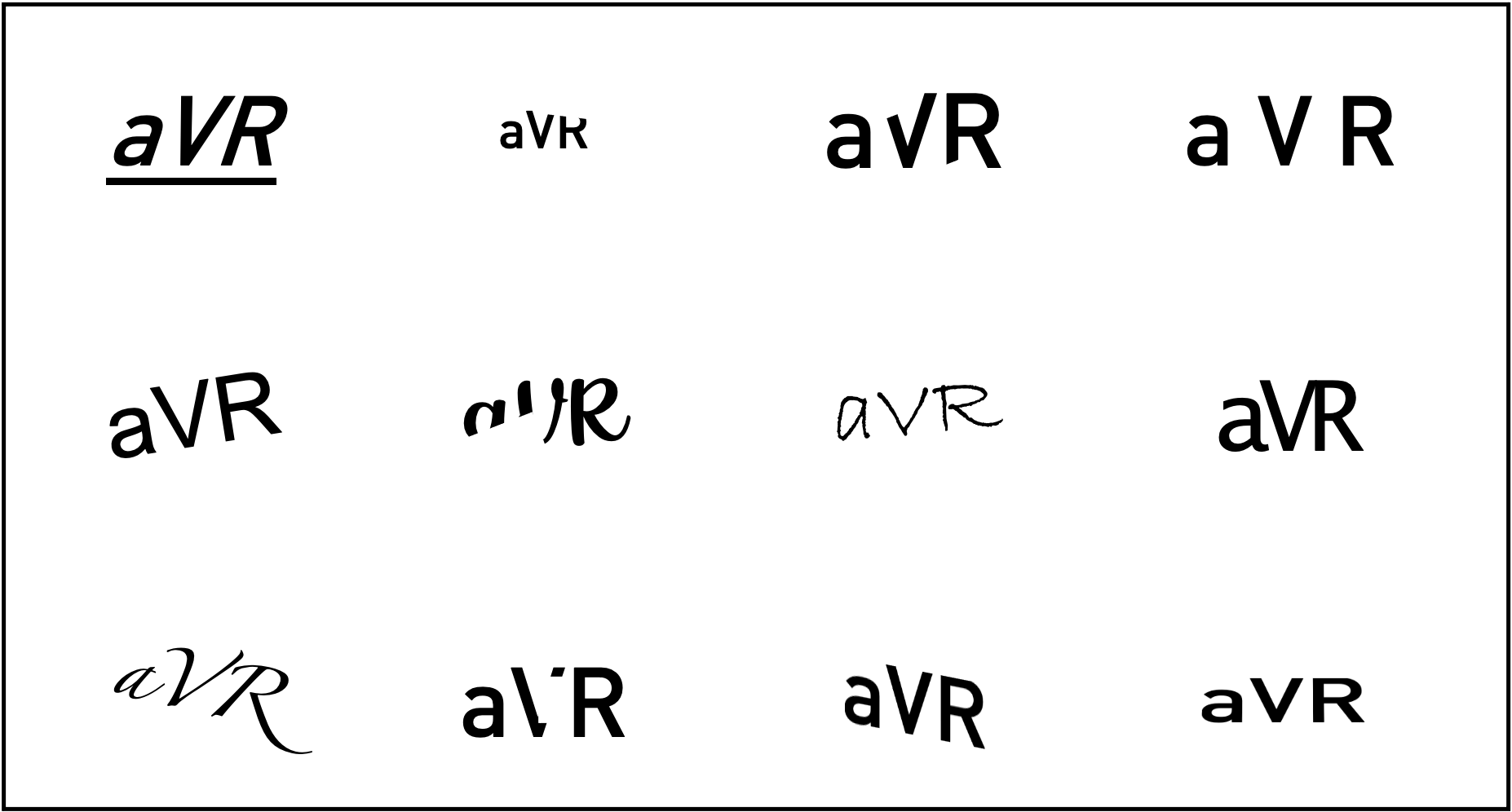}\\
    \vspace{0.5em}
    \includegraphics[width=\textwidth]{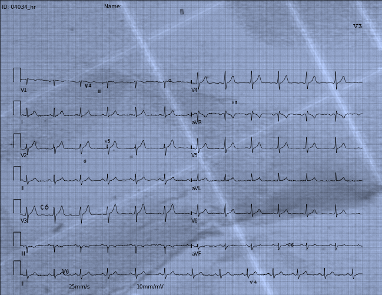}\\
    \textbf{(A)}
  \end{minipage}%
  \hfill
  \begin{minipage}{0.45\textwidth}
    \centering
    \includegraphics[width=\textwidth]{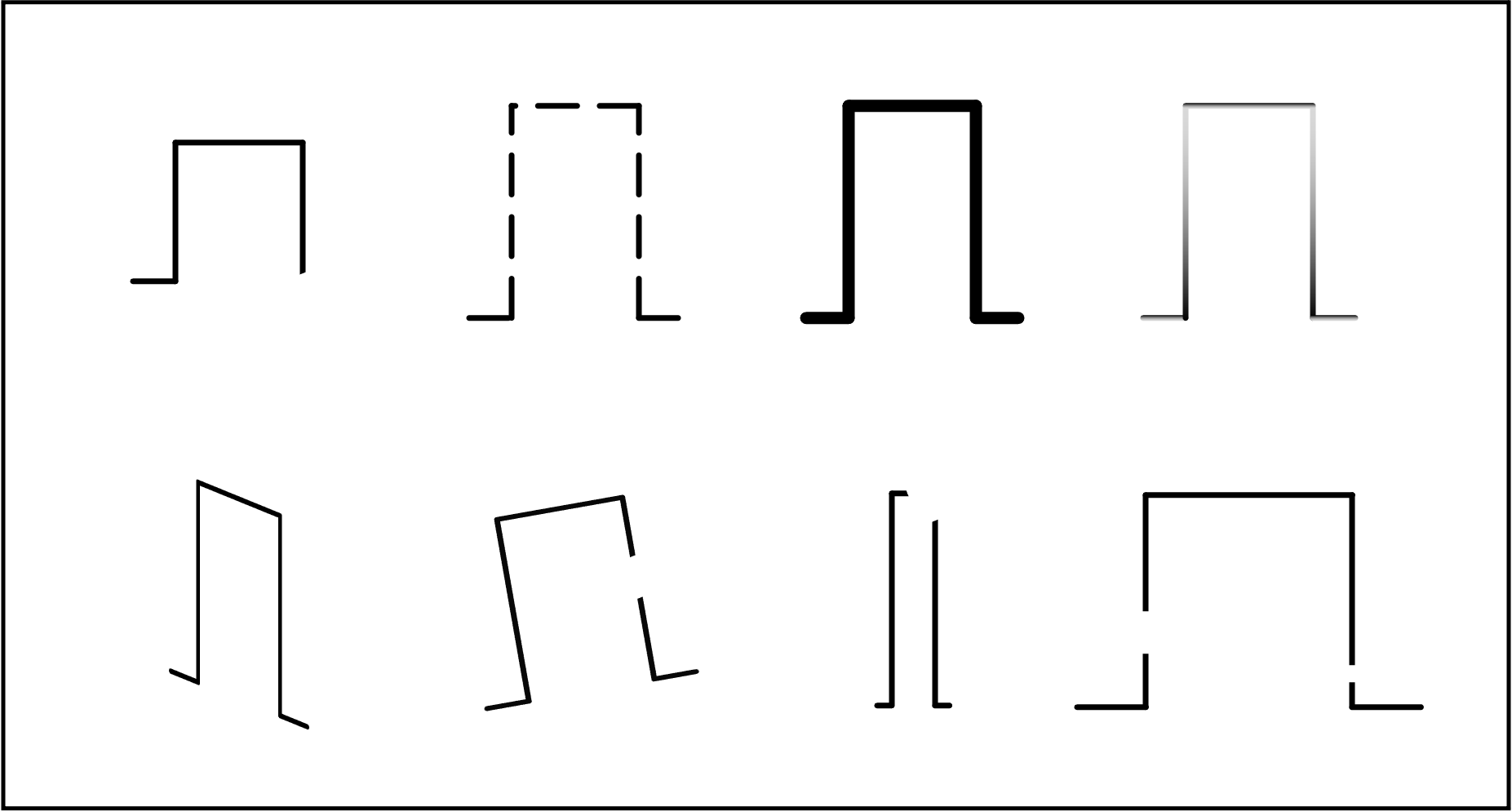}\\
    \vspace{0.5em}
    \includegraphics[width=\textwidth]{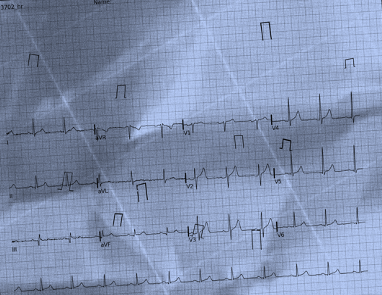}\\
    \textbf{(B)}
  \end{minipage}
  \caption{\textbf{Training-time augmentations for lead-name and reference-pulse detectors.} \textbf{(A)} Lead-label crops (example: \texttt{aVR}) are augmented with font/size changes, rotation, blur, contrast variation, noise, and partial occlusion to improve robustness to printing and scanning variability. \textbf{(B)} Reference-pulse crops are augmented with similar photometric/geometric distortions and overprint artifacts to ensure reliable calibration-marker localization even when pulses are faint, blurred, or partially occluded. These augmentations reduce false positives in text-dense regions and improve generalization to heterogeneous paper ECG styles.}
  \label{fig:objdet}
\end{figure*}

Reference pulse localization was performed using a fine-tuned \texttt{YOLOv11x} model trained on synthetic ECG patches derived from the \texttt{PTB-XL} dataset. To ensure robust generalization across different paper qualities and print configurations, the model was trained on heavily augmented samples (see \href{fig:objdet}{Figure~\ref{fig:objdet}}).  
\href{fig:objdet}{Figure~\ref{fig:objdet}B} depicts the variety of augmentations applied to reference pulse regions—such as blurring, contrast variation, rotation, and overprinting artifacts—while \href{fig:objdet}{Figure~\ref{fig:objdet}A} shows similar augmentation diversity for lead labels. These augmentations allowed the detector to remain robust under diverse imaging conditions, including faded ink, scanner noise, and handwritten annotations.  

The resulting detector achieved a box precision and recall of \textbf{1.000} and \textbf{1.000}, with a mean average precision (mAP@50) of \textbf{0.995}, indicating that the reference pulses are visually distinct and highly separable from surrounding ECG structures (see \href{tab:results}{Table~\ref{tab:results}B}). This exceptional detection reliability arises from the rectangular geometry, high contrast, and spatial consistency of calibration markers relative to lead waveforms.

\begin{figure*}[!t]
\centering
\begin{subfigure}[t]{0.48\textwidth}
  \centering
  \includegraphics[width=0.90\linewidth]{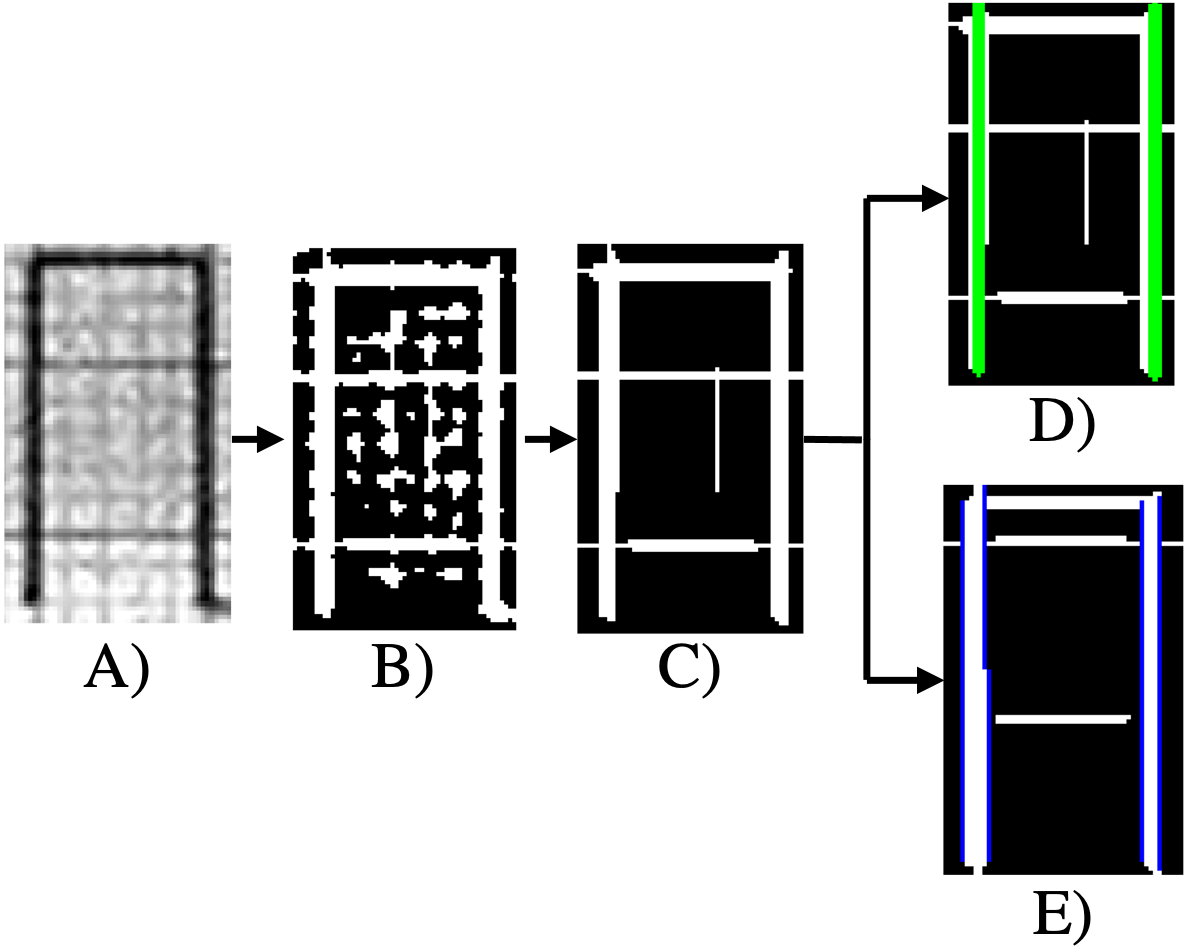}
  \caption{\textbf{Scale determination from reference pulse.} Otsu thresholding, morphological opening, and line extraction are used to estimate pixel-to-mV and pixel-to-ms scaling.}
  \label{fig:ref}
\end{subfigure}\hfill
\begin{subfigure}[t]{0.48\textwidth}
  \centering
  \includegraphics[width=0.75\linewidth]{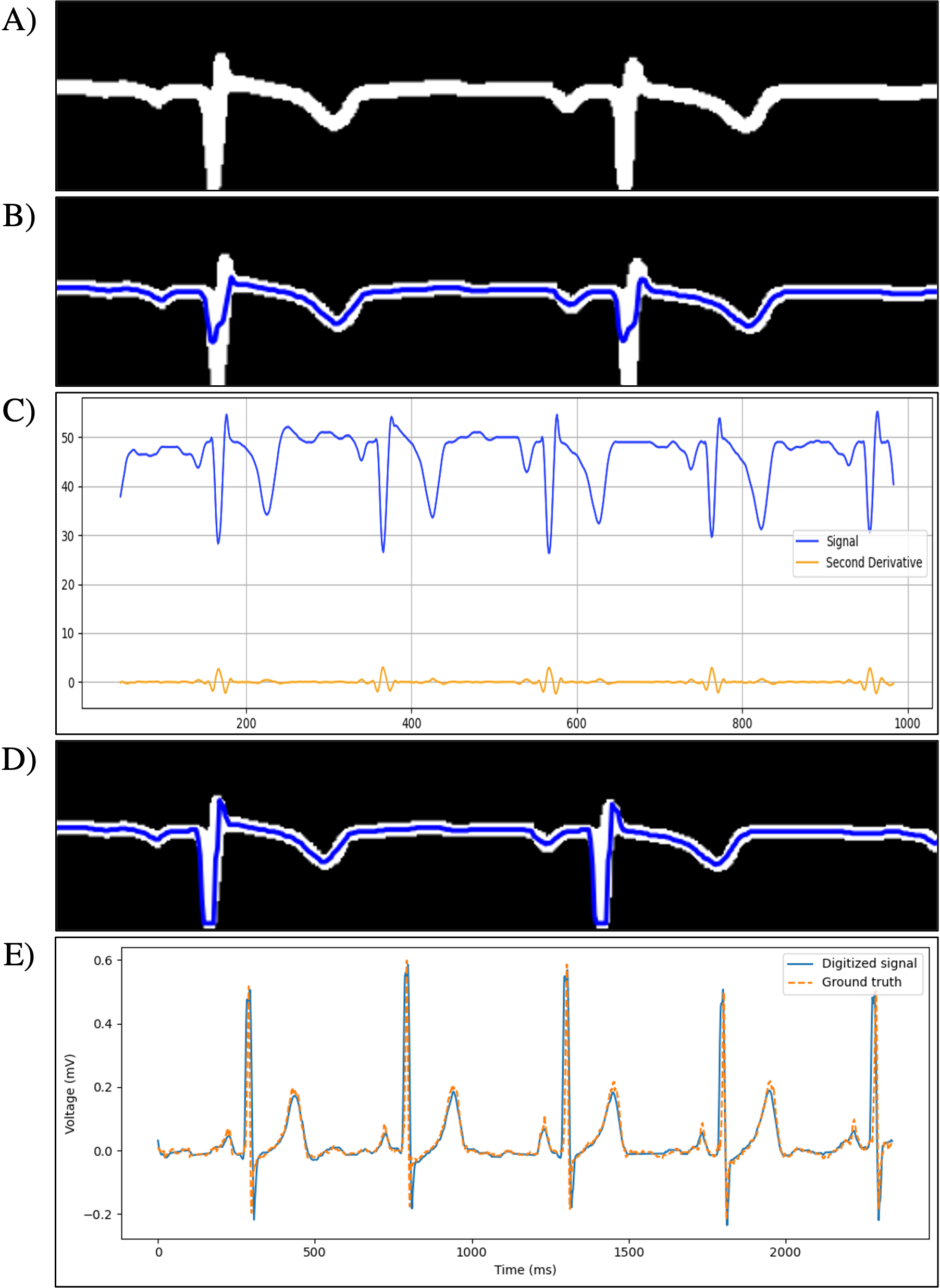}
  \caption{\textbf{Signal vectorization.} Centroid tracing with peak-aware reweighting converts the segmented trace into a calibrated time series.}
  \label{fig:vec}
\end{subfigure}
\caption{\textbf{Reference-pulse calibration and waveform vectorization (paper ECG image $\rightarrow$ physically calibrated time-series).} \textbf{(A)} The calibration pulse provides a device-independent anchor to convert pixel distances into clinical units (mV and ms), enabling consistent amplitude/time scaling across layouts and scan qualities. \textbf{(B)} Given the calibrated scale and the segmentation mask, a centroid-based trace extraction with peak-aware reweighting converts the rasterized waveform into a continuous 1D signal while preserving sharp QRS peaks and ST--T morphology. Together, these steps produce the lead-aligned, physically interpretable 12-lead matrix used in downstream analyses.}
\label{fig:ref_vec}
\end{figure*}

\textbf{Scaling Estimation.}  
Once detected, each reference pulse region underwent a multi-step image processing pipeline to extract quantitative scaling parameters, as shown in \href{fig:ref}{Figure~\ref{fig:ref}}. The process began with \textbf{Otsu multi-thresholding}~\cite{otsu_threshold_1979} to binarize the pulse region and suppress background grid lines (\href{fig:ref}{Figure~\ref{fig:ref} B}). Subsequently, \textbf{morphological opening}~\cite{gonzalez_digital_2009} with a $1\times25$ kernel was applied to remove residual horizontal noise and retain only the vertical pulse edges (\href{fig:ref}{Figure~\ref{fig:ref} C}). The refined binary mask was then analyzed using a combination of a \textbf{probabilistic Hough line transform}~\cite{kiryati_probabilistic_1991} and a \textbf{line segment detector (LSD)}~\cite{grompone_von_gioi_lsd_2010}, which accurately measured the pulse’s height and width (\href{fig:ref}{Figure~\ref{fig:ref} D-E}). These measurements provided the pixel-to-millivolt and pixel-to-millisecond scaling ratios used in all subsequent waveform conversions.

To further refine amplitude quantification, the localized pulse signal was analyzed using an adaptive envelope-tracking algorithm~\cite{Scholkmann2012}, which determined the peak-to-baseline vertical displacement. The resulting calibration factor—typically $\sim\!1~\mathrm{mV}/h_{\text{pixels}}$—was applied uniformly across all segmented leads, ensuring consistent scaling throughout the reconstructed 12-lead recording.

\textbf{Signal Vectorization and Reconstruction.}  
After amplitude and time scaling were established, each segmented lead waveform was vectorized into a continuous time series. This step involved centroid-based tracing of nonzero pixels within the segmented region, generating an initial estimate of the waveform’s trajectory (\href{fig:vec}{Figure~\ref{fig:vec} B}). The second derivative of this signal was computed to identify rapid curvature changes corresponding to ECG peaks and troughs (\href{fig:vec}{Figure~\ref{fig:vec} C}). These peak regions were then used to adjust the centroid weighting, refining waveform accuracy and preserving fine morphological details (\href{fig:vec}{Figure~\ref{fig:vec} D}). The final calibrated signal was rescaled using the pixel-to-millivolt and pixel-to-millisecond factors derived from the reference pulse, producing a fully digitized ECG waveform (\href{fig:vec}{Figure~\ref{fig:vec} E}).

Unlike traditional grid-based scaling methods that rely on visible background grids, this approach leverages the inherent geometric stability of calibration pulses. Grid-based techniques often fail under real-world conditions—such as uneven illumination, fading ink, or compression artifacts—that obscure grid visibility. By contrast, reference-pulse detection provides a layout-independent calibration mechanism that remains invariant to such degradations. This method enables precise amplitude reconstruction even when grid lines are missing or distorted, allowing for robust and fully automated signal scaling in diverse scanning environments.

Together, the detection and scaling stages form the quantitative foundation of the entire digitization pipeline. They bridge the visual and analytical domains, converting segmented waveforms into clinically interpretable signals that preserve both morphological structure and true physiological amplitude.

\begin{figure*}[!t]
\centering
\includegraphics[width=1\textwidth]{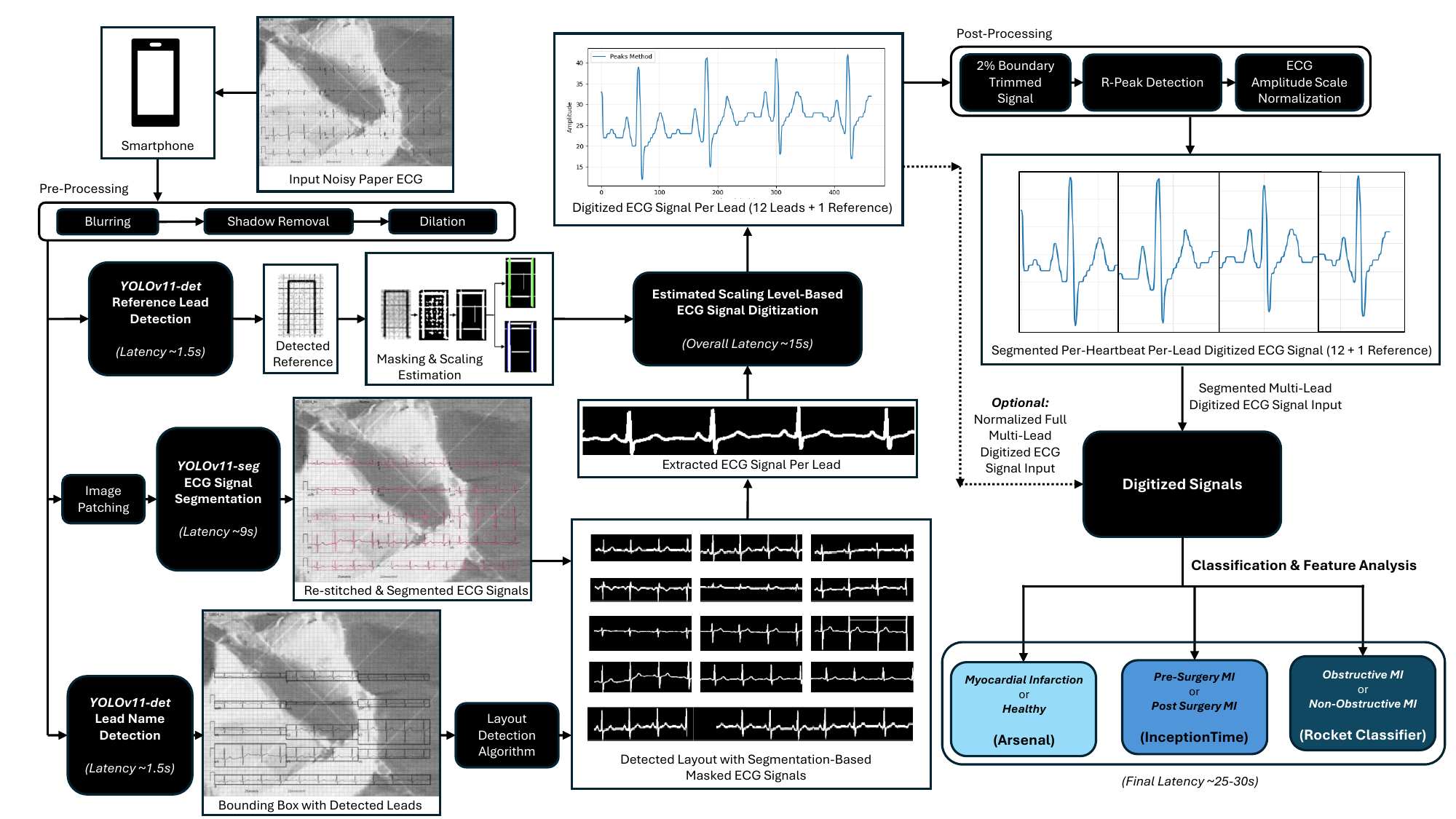}
\caption{\textbf{End-to-End ECG Workflow:} Input noisy paper ECG images undergo pre-processing (shadow removal, dilation and blurring), followed by \texttt{YOLOv11}-based segmentation and detection for signal extraction, lead identification, reference detection, and layout estimation. Scale is determined via reference pulse detection, contour detection and masking to digitize re-stitched ECG signals into multi-lead time-series data. Post-processing comprises trimming, amplitude normalization and R-peak detection. The processed signals are classified using an \texttt{ARSENAL} classifier. Latency measurements were conducted on an Intel Core i9-13900H laptop CPU, which achieved 25-30 seconds using PyTorch Native \texttt{YOLOv11}.}
\label{fig:pipeline_detailed}
\end{figure*}

\subsection*{Noisy Paper ECG Digitization Evaluation}

\subsubsection*{Evaluation cohorts and dataset composition}
To contextualize the performance of the proposed ECG digitization pipeline within the current state of the art, we conducted a comparative analysis against several leading submissions from recent ECG digitization and reconstruction challenges. All benchmarking was performed on a well-characterized cohort with clearly defined training and evaluation splits (\href{tab:dataset}{Table~\ref{tab:dataset}}). This includes (i) the full filtered \texttt{PTB-XL} development cohort ($N=21{,}799$ recordings; 18,869 patients), (ii) a patched training subset ($N=11{,}000$ recordings; 323,000 patches; mean $29.36 \pm 7.42$ patches/patient) used to increase robustness to layout variability and common scanning artifacts, and (iii) a final held-out end-to-end evaluation subset of $N=1{,}600$ ECGs from 1,574 patients with near-balanced sex distribution (male:female ratio $\approx 1.06$--$1.09$) and reported mean age $\pm$ SD. \href{tab:dataset}{Table~\ref{tab:dataset}} further summarizes diagnostic superclass composition (\texttt{NORM}, \texttt{MI}, \texttt{STTC}, \texttt{CD}, \texttt{HYP}, \texttt{OTHER}) and the inherent class imbalance typical of clinical ECG corpora.

\subsubsection*{Evaluation metrics}
Digitization fidelity was quantified using three complementary signal-level metrics relative to the \texttt{PTB-XL} ground-truth waveforms: \textbf{Signal-to-Noise Ratio (SNR)} under both \textit{clean} and \textit{deteriorated} imaging conditions, \textbf{Pearson correlation coefficient} ($r$) to assess morphological agreement, and \textbf{Root Mean Square Error (RMSE)} to assess amplitude reconstruction accuracy (\href{tab:combined_results}{Table~\ref{tab:combined_results}}). Together, these metrics capture robustness to degradation, preservation of clinically relevant waveform shape, and calibration correctness in physical units.

\subsubsection*{Evaluation protocol and comparability}
All comparisons were performed on the same held-out evaluation subset described above, and all metrics were computed at the signal level after converting image-space traces to physically calibrated units (mV and ms). The clean vs.~deteriorated SNR reporting follows the challenge convention and is particularly informative for paper ECG digitization because it isolates robustness to visual degradations (e.g., blur, noise, contrast shifts, and grid visibility) from the intrinsic difficulty of reconstructing the underlying waveform morphology. Where applicable, significance testing was performed using standard thresholds (e.g., $p<0.05$) to confirm that performance differences were not attributable to random variation.

\subsubsection*{Comparative benchmarking against challenge baselines}
The comparison in \href{tab:combined_results}{Table~\ref{tab:combined_results}} includes top-performing teams such as \texttt{USST\_Med}, \texttt{Ahus AI Lab}, \texttt{wavie\_ABI}, \texttt{BAPORLab}, and \texttt{SignalSavants}, each of which employed distinct strategies for ECG vectorization and denoising. While some models, notably \texttt{SignalSavants}, achieved the highest peak SNR on clean inputs (\textbf{12.151~dB}), their performance deteriorated substantially under degraded conditions (\textbf{3.479~dB}; relative degradation \textbf{0.514}). Similar sensitivity was observed for \texttt{BAPORLab} (\textbf{5.493~dB} clean; \textbf{4.735~dB} deteriorated), illustrating that strong performance on pristine images does not necessarily translate to robust paper-ECG digitization.

\subsubsection*{Interpretation of reconstruction fidelity and robustness}
In contrast, the proposed \textbf{YOLOv11x}-based digitization framework achieved an overall \textbf{SNR of 4.54~dB} together with strong waveform agreement (\textbf{Pearson $r=0.806$}) and low amplitude error (\textbf{RMSE $=0.043$~mV}). The relatively stable behavior under degradations indicates that the pipeline preserves waveform integrity even when segmentation is challenged by grid remnants, illumination variability, and compression artifacts. These results are consistent with the design of the pipeline: patch-based segmentation increases local contour fidelity, lead-name/layout inference enforces correct topological reconstruction of the 12-lead matrix, and reference-pulse calibration anchors amplitude and time scaling in physical units, reducing sensitivity to grid visibility.

\subsubsection*{Practical implications and failure modes}
Although aggregate metrics summarize average fidelity, digitization failures in practice are typically driven by a small set of recurring issues: (i) severe overlap between adjacent lead traces, (ii) missing or occluded calibration pulses that prevent reliable scaling, (iii) extreme baseline drift or saturation that breaks single-valued trace assumptions, and (iv) non-standard report layouts that violate the learned priors. These conditions motivate the inclusion of explicit quality-control checks (e.g., calibration plausibility bounds and lead-completeness requirements) before downstream clinical modeling. Overall, the results in \href{tab:combined_results}{Table~\ref{tab:combined_results}} support that the proposed method achieves a robust accuracy--robustness trade-off suitable for real-world digitization scenarios, particularly when the downstream goal is to retain clinically meaningful morphology for diagnostic modeling.

\begin{table*}[!t]
\centering
\begin{tabular}{lcc}
\toprule
\textbf{Team / Method} & \textbf{SNR Leaderboard  (Clean, Deteriorated)} & \textbf{Overall Performance (Pearson, RMSE)} \\
\midrule
USST\_Med       & 2.202 ($-$0.058, $-$0.375) & -- \\
Ahus AI Lab     & 3.047 (2.777, $-$0.320) & -- \\
wavie\_ABI      & 5.469 (--, --) & -- \\
BAPORLab        & 5.493 (4.735, 0.358) & -- \\
SignalSavants   & 12.151 (3.479, 0.514) & -- \\
\midrule
\textbf{Proposed Method} & \textbf{4.54} & \textbf{0.806, 0.043} \\
\bottomrule
\end{tabular}
\caption{\textbf{ECG digitization benchmarking versus leading challenge submissions.} Comparison of the proposed \texttt{YOLOv11x}-based digitization pipeline against representative top submissions from recent ECG reconstruction challenges. The SNR column reports the official leaderboard metric and SNR measured on \textbf{clean} and \textbf{deteriorated} test conditions (in parentheses; higher is better). The final column reports overall waveform agreement with the ground truth using Pearson correlation ($r$) and RMSE (mV) when available (``--'' indicates not reported). Smaller SNR drop from clean to deteriorated inputs indicates greater robustness to visual degradation.}
\label{tab:combined_results}
\end{table*}

Overall, the results summarized in \href{tab:combined_results}{Table~\ref{tab:combined_results}} establish that the proposed pipeline provides a balanced compromise between signal fidelity and robustness. While some challenge entries achieve higher peak SNRs in ideal conditions, they exhibit higher degradation under real-world noise. In contrast, the presented method demonstrates consistently reliable performance, marking a significant step toward practical, fully automated digitization of heterogeneous paper ECGs suitable for downstream clinical and diagnostic applications.

\subsection*{\texttt{MLP}-Based SHAP Feature Importance Analysis}

\subsubsection*{Dataset and preprocessing}
\begin{figure*}[!t]
\centering
\includegraphics[width=\textwidth]{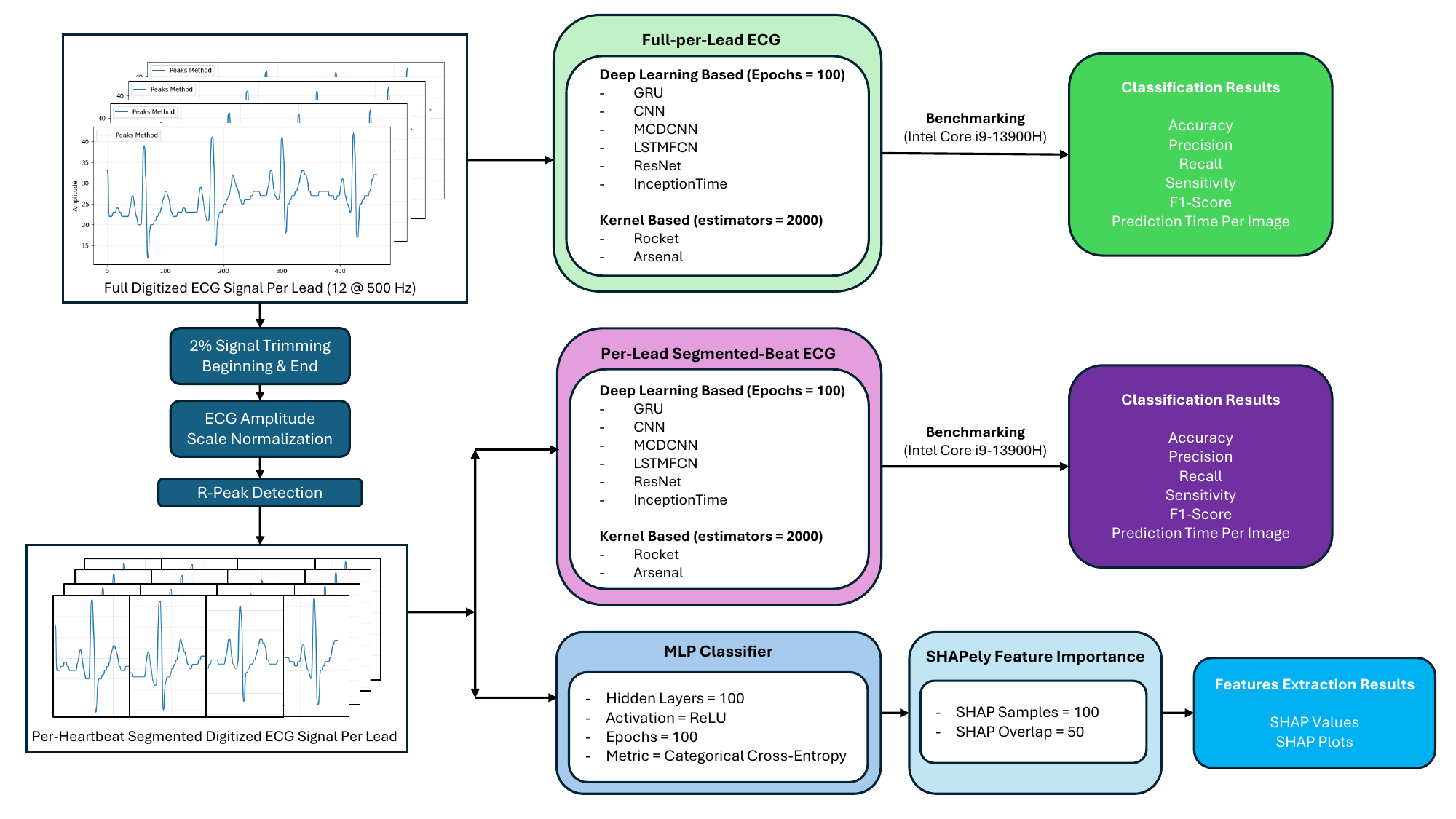}
\caption{\textbf{Workflow for beat-level \texttt{MLP} classification and \texttt{SHAP} attribution.} Digitized 12-lead ECGs are standardized to 500~Hz and segmented into R-peak-centered beat windows~\cite{9707247}; beats are normalized/denoised and used to train an \texttt{MLP}. \texttt{SHAP} values are then computed to localize predictive evidence by lead and time (e.g., QRS and ST--T regions).}
\label{fig:mlp_workflow}
\end{figure*}

Following digitization, the reconstructed 12-lead time-series were standardized to \textbf{500~Hz} and segmented into beat-level windows using an R-peak detection algorithm~\cite{9707247} (\href{tab:train_test_split}{Table~\ref{tab:train_test_split}}). Each beat window was z-score normalized per lead to reduce amplitude offsets across recordings and lightly denoised to attenuate low-frequency baseline wander. These beat tensors (one tensor per beat, with all 12 leads) were used to train a shallow \texttt{MLP} for \texttt{Normal} vs.~\texttt{MI} classification on digitized \texttt{PTB-XL} (workflow in \href{fig:mlp_workflow}{Figure~\ref{fig:mlp_workflow}}).

\subsubsection*{MLP model and training configuration}
The \texttt{MLP} architecture was intentionally lightweight to support transparent attribution: two fully connected hidden layers (100 and 50 neurons), ReLU nonlinearities, and dropout regularization (0.2). Training used the Adam optimizer with learning rate $10^{-3}$ and a fixed random seed. Although the \texttt{MLP} is not the strongest-performing classifier in our benchmark suite, it provides a stable and computationally efficient surrogate model for interpretability (average latency $\approx 1.25$~ms per beat). On the segmented \texttt{PTB-XL} \textit{MI vs.~Normal} task, it achieved accuracy $0.8318$, with errors partly attributable to the fact that record-level MI labels can include morphologically normal beats within an MI-labeled recording.

\subsubsection*{SHAP attribution protocol and clinical interpretation}
To quantify which leads and time regions contributed most to the \texttt{MLP} decision, we computed \texttt{SHAP} values on held-out test beats using a representative background set (as described in the main Methods). Because each input feature corresponds to a specific lead and time index, the resulting attributions can be aggregated (i) across time to rank leads by global importance and (ii) within-lead to localize evidence to clinically interpretable ECG segments. Across samples, the attribution maps consistently emphasized physiologically plausible regions: the QRS complex and the ST--T segment carried the highest marginal contributions, while baseline and atrial components contributed less. In terms of leads, the strongest SHAP contributions were concentrated in precordial leads (\texttt{V1}--\texttt{V3}) and lateral limb leads (\texttt{I}, \texttt{aVL}), which aligns with clinical expectations for infarction-related depolarization and repolarization abnormalities.

\subsection*{\texttt{SMOTE-Tomek} Oversampling in \textit{Pre-} vs.\textit{~Post-Procedural} MI Classification Benchmarking}
\label{sec:supp_smote_surgery}

\href{tab:ecg_surgery_undersampling_vs_oversampling}{Table~\ref{tab:ecg_surgery_undersampling_vs_oversampling}} reports \textit{Pre- vs.~Post-Procedural MI} benchmarking on the \texttt{ECG Matrix} dataset with \texttt{Synthetic Minority Over-sampling Technique) SMOTE-Tomek} applied to mitigate class imbalance in the \emph{training split only} (the held-out test split was unchanged). All models operated on the digitized outputs of our paper-to-signal pipeline (physically calibrated 12-lead time-series in mV and seconds). Each record was represented as fixed-length \textbf{$140\times 12$} segments (140 time steps per lead) to standardize inputs and enable controlled accuracy--latency comparisons. \texttt{SMOTE} synthesizes minority-class samples via nearest-neighbor interpolation, while \texttt{Tomek} links remove ambiguous near-boundary majority/minority pairs to reduce local class overlap.

\begin{table*}[!t]
\centering
\setlength{\tabcolsep}{6.0pt}
\renewcommand{\arraystretch}{1.15}
\resizebox{\textwidth}{!}{%
\begin{tabular}{@{}l@{\hspace{6pt}\vrule\hspace{6pt}}l
S[table-format=5.2]
S[table-format=1.4] S[table-format=1.4] S[table-format=1.4] c S[table-format=1.4]@{}}
\toprule
& & \multicolumn{6}{c}{\texttt{ECG-Matrix}-Oversampling (\texttt{SMOTE-Tomek})} \\
\cmidrule(lr){3-8}
\textbf{Group} & \textbf{Model}
& {\textbf{Inference Time (ms)}} & {\textbf{Accuracy}} & {\textbf{Precision}} & {\textbf{Recall}} & {\textbf{Specificity}} & {\textbf{F1-score}} \\
\midrule
\multirow[t]{7}{*}{\cellcolor{white}\textbf{Deep Learning}\strut}
& InceptionTime & 1.38 & 0.8637 & 0.6215 & 0.6429 & 0.9131 & 0.6320 \\
& LSTM-FCN      & 1.69 & 0.8845 & 0.6798 & 0.6914 & 0.9271 & 0.6856 \\
& MCDCNN        & 0.75 & 0.8684 & 0.6570 & 0.5800 & 0.9337 & 0.6161 \\
& ResNet        & 1.34 & 0.8580 & 0.5960 & \textbf{0.6829} & 0.9064 & 0.6365 \\
& CNN           & 0.76 & 0.7279 & 0.3383 & 0.5171 & 0.7735 & 0.4090 \\
& GRU           & 15.14 & 0.5624 & 0.2245 & 0.5714 & 0.5602 & 0.3223 \\
\midrule
\multirow[t]{2}{*}{\cellcolor{white}\textbf{Kernel-Based}\strut}
& Rocket        & 905.99 & 0.9058 & 0.7735 & \textbf{0.6829} & 0.9581 & 0.7253 \\
\rowcolor{blue!12}\cellcolor{white}
& Arsenal       & 10736.04 & \textbf{0.9089} & \textbf{0.8028} & 0.6629 & \textbf{0.9637} & \textbf{0.7261} \\
\bottomrule
\end{tabular}%
}

\caption{\textbf{\textit{ECG-Matrix} benchmarking (Pre- vs.~Post-Procedural MI) with SMOTE--Tomek.} Models are evaluated on fixed-length 12-lead segments (140 timesteps/lead). We report average CPU inference time (ms) and Accuracy/Precision/Recall/Specificity/F1; best values are \textbf{bold} and the top model row is shaded.}
\label{tab:ecg_surgery_undersampling_vs_oversampling}
\end{table*}

We compared compact deep-learning baselines (\texttt{InceptionTime}, \texttt{LSTM-FCN}, \texttt{MCDCNN}, \texttt{ResNet}, \texttt{CNN}, \texttt{GRU}) against kernel-based time-series methods (\texttt{Rocket}, \texttt{Arsenal}), reporting accuracy, precision, recall, specificity, and F1-score together with average per-sample CPU inference time. In this oversampled setting, kernel-based ensembles achieved the strongest overall metrics (e.g., \texttt{Arsenal} accuracy \textbf{0.9089}, precision \textbf{0.8028}, specificity \textbf{0.9637}) but at substantially higher inference cost than the deep-learning baselines, illustrating a persistent performance--latency trade-off. For consistency and comparability with the primary benchmarks, the main manuscript emphasizes results on the original patient-wise splits without synthetic oversampling (\href{tab:train_test_split}{Table~\ref{tab:train_test_split}}).

\subsection*{Per-Lead Classification Performance}

To further characterize lead-specific diagnostic informativeness, we performed a per-lead classification study in which the \emph{same} \texttt{MLP} architecture was trained and evaluated using each lead independently (i.e., single-lead input). All inputs were obtained from the digitized, physically calibrated 12-lead reconstructions produced by the proposed pipeline, but only one lead vector was provided to the classifier at a time. This setup intentionally removes multi-lead cues (e.g., contiguous-lead concordance and reciprocal changes), so any remaining predictive signal is attributable to morphology within the single channel.

To preserve the clinical validity of the evaluation, we used the same subject-level splitting strategy as the main beat-level experiments so that all beats from a given subject remain within the same split. The cohort definitions and segmented train/test sizes for each endpoint are reported in \href{tab:train_test_split}{Table~\ref{tab:train_test_split}}.

\subsubsection*{Datasets, segmentation, and feature construction}
Per-lead benchmarking was performed on the segmented-heartbeat representation to (i) increase the number of training/evaluation instances and (ii) isolate lead-local morphology around depolarization and early repolarization. Digitized signals were standardized to 500~Hz and segmented via R-peak detection into fixed windows of \textbf{280} time steps (0.56\,s) centered on each detected peak. For each lead, this yields a length-280 vector; for \texttt{MLP} input, this vector was flattened into a 1D feature array of length 280.

\begin{table*}[!t]
\centering
\setlength{\tabcolsep}{4.0pt}
\renewcommand{\arraystretch}{1.06}
\begin{tabular}{@{}>{\centering\arraybackslash}p{2.7cm} c ccccc@{}}
\toprule
\textbf{Dataset} & \textbf{Lead} & \textbf{Accuracy} & \textbf{Precision} & \textbf{Recall} & \textbf{Specificity} & \textbf{F1 Score} \\
\midrule
\multirow{12}{*}{\makecell[c]{\textit{MI}\\\textbf{vs}\\\textit{Normal}\\\footnotesize\texttt{(PTB-XL)}}}
& II  & 0.798 & 0.814 & 0.776 & 0.821 & 0.795 \\
& I   & 0.762 & 0.796 & 0.709 & 0.817 & 0.750 \\
& aVF & 0.734 & 0.743 & 0.721 & 0.748 & 0.732 \\
& III & 0.735 & 0.758 & 0.693 & 0.777 & 0.724 \\
& aVR & 0.736 & 0.779 & 0.662 & 0.810 & 0.716 \\
& aVL & 0.701 & 0.704 & 0.698 & 0.704 & 0.701 \\
& V6  & 0.706 & 0.720 & 0.681 & 0.732 & 0.700 \\
& V5  & 0.712 & 0.736 & 0.665 & 0.759 & 0.699 \\
& V2  & 0.708 & 0.748 & 0.633 & 0.784 & 0.686 \\
& V3  & 0.702 & 0.736 & 0.636 & 0.769 & 0.682 \\
& V4  & 0.708 & 0.766 & 0.602 & 0.814 & 0.674 \\
& V1  & 0.647 & 0.648 & 0.649 & 0.644 & 0.649 \\
\midrule
\multirow{12}{*}{\makecell[c]{\textit{Pre-Procedural MI}\\\textbf{vs}\\\textit{Post-Procedural MI}\\\footnotesize\texttt{(ECG Matrix)}}}
& aVF & 0.698 & 0.689 & 0.608 & 0.783 & 0.646 \\
& III & 0.630 & 0.667 & 0.366 & 0.855 & 0.473 \\
& V2  & 0.615 & 0.672 & 0.294 & 0.886 & 0.409 \\
& I   & 0.595 & 0.551 & 0.562 & 0.643 & 0.557 \\
& aVL & 0.592 & 0.552 & 0.523 & 0.668 & 0.537 \\
& aVR & 0.589 & 0.552 & 0.484 & 0.694 & 0.516 \\
& V5  & 0.589 & 0.552 & 0.484 & 0.694 & 0.516 \\
& V3  & 0.589 & 0.537 & 0.667 & 0.549 & 0.595 \\
& II  & 0.586 & 0.580 & 0.307 & 0.829 & 0.402 \\
& V6  & 0.568 & 0.528 & 0.431 & 0.699 & 0.475 \\
& V1  & 0.553 & 0.512 & 0.288 & 0.787 & 0.368 \\
& V4  & 0.536 & 0.490 & 0.614 & 0.497 & 0.545 \\
\midrule
\multirow{12}{*}{\makecell[c]{\textit{OMI}\\\textbf{vs}\\\textit{non-OMI}\\\footnotesize\texttt{(ECG Matrix)}}}
& II  & 0.625 & 0.621 & 0.496 & 0.730 & 0.552 \\
& III & 0.624 & 0.605 & 0.551 & 0.676 & 0.577 \\
& aVF & 0.624 & 0.601 & 0.569 & 0.662 & 0.584 \\
& aVR & 0.619 & 0.593 & 0.574 & 0.648 & 0.584 \\
& V2  & 0.602 & 0.567 & 0.608 & 0.585 & 0.587 \\
& aVL & 0.589 & 0.587 & 0.389 & 0.756 & 0.468 \\
& V1  & 0.580 & 0.567 & 0.407 & 0.722 & 0.474 \\
& I   & 0.578 & 0.555 & 0.462 & 0.670 & 0.504 \\
& V6  & 0.578 & 0.543 & 0.577 & 0.568 & 0.559 \\
& V4  & 0.573 & 0.550 & 0.446 & 0.673 & 0.493 \\
& V3  & 0.573 & 0.535 & 0.619 & 0.520 & 0.574 \\
& V5  & 0.566 & 0.532 & 0.540 & 0.577 & 0.536 \\
\bottomrule
\end{tabular}

\caption{\textbf{Per-lead classification performance (single-lead \texttt{MLP} inputs).} Accuracy/Precision/Recall/Specificity/F1 are reported for each lead on: \texttt{PTB-XL} (MI vs.~Normal) and \texttt{ECG Matrix} (Pre- vs.~Post-Procedural MI; OMI vs.~non-OMI). Leads are sorted by F1 within each block.}
\label{tab:per_lead_performance}
\end{table*}

Concretely, for \texttt{PTB-XL} \textit{MI vs.~Normal} the segmented dataset contained \textbf{7,045} training beats and \textbf{2,349} test beats; for \texttt{ECG Matrix} \textit{Pre- vs.~Post-Procedural MI} it contained \textbf{1,032} training beats and \textbf{344} test beats; and for \texttt{ECG Matrix} \textit{OMI vs.~non-OMI} it contained \textbf{1,997} training beats and \textbf{666} test beats (\href{tab:train_test_split}{Table~\ref{tab:train_test_split}}). Each beat window was normalized (z-score) within lead to reduce amplitude offsets introduced by heterogeneous acquisition and digitization conditions, and the same light denoising/baseline suppression used in the main beat-level pipeline was retained to minimize low-frequency drift.

\subsubsection*{MLP configuration and training protocol}
For comparability across leads and tasks, we reused the \texttt{MLP} training configuration from the multi-lead SHAP experiment: ReLU activations, dropout regularization (0.2), and Adam optimization with learning rate $10^{-3}$ under a fixed random seed. Training was conducted independently per lead, so each reported score reflects a separate single-lead model fit and evaluation. This design isolates lead informativeness while keeping the classifier capacity fixed.

\subsubsection*{Consistency with SHAP lead rankings}
Across tasks, the single-lead results are broadly consistent with the lead rankings implied by the \texttt{SHAP} attribution maps (\href{fig:mlp_workflow}{Figure~\ref{fig:mlp_workflow}}) and the best-lead patterns discussed in the main text. In particular, SHAP emphasizes leads that carry infarction-relevant evidence in the ST--T and QRS regions; the same leads tend to remain informative when evaluated in isolation, providing an empirical check that the attributions correspond to predictive content in the digitized waveform rather than multi-lead interactions alone.

\subsubsection*{Best-performing leads and clinical interpretation}
\href{tab:per_lead_performance}{Table~\ref{tab:per_lead_performance}} summarizes per-lead performance across three endpoints. For digitized \texttt{PTB-XL} (\textit{MI vs.~Normal}), lead~II achieves the strongest single-lead F1 score (F1 $=0.795$), consistent with the diagnostic value of inferior leads for capturing ischemic and infarction-related ST--T deviations and Q-wave changes. Additional limb leads (I, III, aVF) and several precordial leads (V2--V6) achieve competitive scores, indicating that infarction evidence is not confined to a single channel.

For digitized \texttt{ECG Matrix} \textit{Pre- vs.~Post-Procedural MI}, aVF provides the strongest single-lead performance (F1 $=0.646$), but overall separability remains lower than \texttt{PTB-XL}, consistent with heterogeneous peri-surgical dynamics and weaker, more variable ECG signatures. For digitized \texttt{ECG Matrix} \textit{OMI vs.~non-OMI}, the best leads are more distributed across inferior and precordial channels (e.g., V2, aVF, III), aligning with an occlusion-centric labeling scheme in which ischemic territory and reciprocal patterns can vary substantially across patients.

Taken together, this per-lead study provides a simple, model-agnostic corroboration of the interpretability analysis: the leads emphasized by SHAP also tend to retain discriminative value under single-lead training, supporting the conclusion that the digitized reconstructions preserve physiologically meaningful, lead-local morphology that is usable for downstream diagnosis.

\href{tab:per_lead_performance}{Table~\ref{tab:per_lead_performance}} reports per-lead accuracy, precision, recall, specificity, and F1 score for each dataset/task. Overall, the integration of interpretable \texttt{MLP}-based feature analysis and lightweight, high-performing classifiers such as \texttt{Rocket} establishes a transparent, efficient, and clinically relevant framework for automated ECG diagnosis from digitized paper recordings.

\end{document}